%% file: sn-article.tex
\let\stdcline\cline
\PassOptionsToPackage{table, dvipsnames}{xcolor}

\documentclass[pdflatex,sn-mathphys-num, iicol]{sn-jnl}%

\let\cline\stdcline
\usepackage{graphicx}%
\usepackage{multirow}%
\usepackage{amsmath,amssymb,amsfonts}%
\usepackage{amsthm}%
\usepackage{mathrsfs}%
\usepackage[title]{appendix}%
\usepackage{xcolor}%
\usepackage{textcomp}%
\usepackage{manyfoot}%
\usepackage{booktabs}%
\usepackage{algorithm}%
\usepackage{algorithmicx}%
\usepackage{algpseudocode}%
\usepackage{listings}%
\usepackage{fixmetodonotes}
\usepackage{hyperref}
\usepackage[nameinlink]{cleveref}
\usepackage{enumitem}
\usepackage{cancel}
\usepackage{makecell}
\usepackage{tikz}
\usetikzlibrary{shapes, arrows, positioning, fit, calc, tikzmark}
\usepackage{ifthen}
\usepackage{pifont}
\usepackage{soul}
\usepackage{capt-of}%
\usepackage[OT1]{fontenc}

\def\tableautorefname{Tab.}
\newcommand{\Tableref}[1]{\begingroup\def\tableautorefname{Table}\autoref{#1}\endgroup}

\definecolor{caribbeangreen}{rgb}{0.0, 0.8, 0.6}
\definecolor{deepskyblue}{rgb}{0.0, 0.75, 1.0}
\definecolor{deepmagenta}{rgb}{0.8, 0.0, 0.8}
\definecolor{azure(colorwheel)}{rgb}{0.0, 0.5, 1.0}
\definecolor{darkcyan}{rgb}{0.0, 0.55, 0.55}

\newboolean{showcomments}
\setboolean{showcomments}{false} %

\ifthenelse{\boolean{showcomments}}{

  \newcommand{\RMc}[1]{\textcolor{deepmagenta}{\textbf{[RM: #1]}}}
  
  \newcommand{\RCc}[1]{\textcolor{azure(colorwheel)}{\textbf{[RdC: #1]}}}
  \newcommand{\ABc}[1]{\textcolor{blue}{\textbf{[AB: #1]}}}
  \newcommand{\TMc}[1]{\textcolor{red}{\bf[TM: #1]}}

  \newcommand{\RCremove}[1]{\setstcolor{azure(colorwheel)}\st{#1}}
}{

  \newcommand{\RMc}[1]{}
  
  \newcommand{\RCc}[1]{}
  \newcommand{\ABc}[1]{}
  \newcommand{\TMc}[1]{}

  \newcommand{\RCremove}[1]{}
}

\definecolor{MyGreen}{RGB}{0, 180, 0}
\definecolor{MyRed}{RGB}{180, 0, 0}
\definecolor{MyYellow}{RGB}{200, 160, 0}
\newcommand{\cmark}{{\textcolor{MyGreen}{\ding{51}}}}%
\newcommand{\nmark}{{\textcolor{MyYellow}{(\ding{55})}}}%
\newcommand{\xmark}{{\textcolor{MyRed}{\ding{55}}}}%

\newcommand\colormain{blue!15}
\newcommand\colortitle{green!15}
\newcommand\colorspecial{black!15}

\newcolumntype{H}{>{\setbox0=\hbox\bgroup}c<{\egroup}}

\theoremstyle{thmstyleone}%

\theoremstyle{thmstyletwo}%

\theoremstyle{thmstylethree}%

\begin{document}

\title[SSC-Priors]{Do we need a new name for the journal version?}
\title[SSC-Priors]{\quad SSC-Priors: Exploring Semantic and Visibility Priors \\ to Boost Lidar Semantic Scene Completion}

\author*[1,2]{\fnm{Tetiana} \sur{Martyniuk}}\email{tetyanka.martynyuk@gmail.com}

\author[1,3]{\fnm{Jonathan} \sur{Seele}}\email{jseele@ethz.ch}

\author[1,2]{\fnm{Alexandre} \sur{Boulch}}\email{alexandre.boulch@valeo.com}

\author[1,2]{\fnm{Gilles} \sur{Puy}}\email{gilles.puy@gmail.com}

\author[1,2,4]{\fnm{Renaud} \sur{Marlet}}\email{renaud.marlet@valeo.com}

\author[1]{\fnm{Raoul} \sur{de Charette}}\email{raoul.de-charette@inria.fr}

\affil[1]{\orgdiv{Inria}, \orgaddress{\city{Paris}, \country{France}}}
\affil[2]{\orgdiv{valeo.ai}, \orgaddress{\city{Paris}, \country{France}}}
\affil[3]{\orgname{ETH Z{\"u}rich}, \orgaddress{\country{Switzerland}}}
\affil[4]{\orgdiv{LIGM}, \orgname{CNRS, Univ Gustave Eiffel, ENPC, IP Paris}, \orgaddress{\country{France}}}

\abstract{

\input{sections/000_abstract}

}

\maketitle
\clearpage

\input{sections/001_introduction}

\input{sections/002_related}

\input{sections/003_recipe}

\input{sections/004_experiments_NEW}

\input{sections/006_conclusion}

\section*{Acknowledgments}
The authors thank Anh-Quan Cao for fruitful discussions.
This work was granted access to the HPC resources of IDRIS under the allocation AD011014484R3 made by GENCI. Gilles Puy contributed to this work while employed by Valeo; he accessed and processed all datasets on valeo.ai compute infrastructures; he is presently employed by Meta. 

\newpage
~
\newpage

\begin{appendices}

\input{sections/007_appendix}

\end{appendices}

\newpage
~
\newpage

\bibliography{strings,refs_short}%

\end{document}

%% file: sections/000_abstract.tex
This paper investigates easy strategies to boost the performance of existing networks for
lidar semantic scene completion (SSC) without requiring complex architectural redesigns. 
The fact is that, over the last years, SSC methods have mostly pursued architectural innovations, making the models heavier and more complex, e.g., by jointly training a
point cloud semantic segmentation branch.
In this work, we take a step back and explore two priors used as simple ingredients (possibly noisy) to improve existing approaches: %
semantic pseudo-labels and sensor visibility information. %
Concretely, we provide both kinds of information 
directly as additional inputs to a given SSC network, requiring only a minimal adaptation of the original architecture.
We first demonstrate that endowing input point clouds with semantic pseudo-labels from off-the-shelf segmenters significantly improves the performance of existing SSC models. %
In fact, by evaluating these models against an oracle, we establish that high-quality semantic priors are a primary driver of semantic %
gains (mIoU), and that the SSC model can be trained just once with ground-truth semantics and then exploited without retraining using any segmenter.
Furthermore, we equip the input lidar point cloud %
with visibility information that distinguishes between \emph{empty} \emph{spaces} (between the lidar and a scanned point) and \emph{unknown spaces} (outside of lines of sight), providing a secondary performance boost across the tested architectures.
We study the design space of data for representing visibility information and bound the remaining headroom with a ground-truth oracle on the free-space labels.
On SemanticKITTI, these %
enhancements make older models competitive with state-of-the-art systems across four architectures, in one case even outperforming them. %
On the SSCBench-nuScenes benchmark, both priors also transfer with %
the sparser 32-beam sensor.
Our code is available at https://github.com/astra-vision/SSC-Priors.

%% file: sections/001_introduction.tex
\input{figures/journal_teaser}

\section{Introduction}
\label{sec:intro}

Perceiving and understanding the 3D environment 
is a critical task for agents interacting with the physical world. For robots, such as self-driving cars, this is particularly important and requires comprehensive sensing of the scene's geometry and semantics.
In fact, in recent years, the task of semantic scene completion (SSC) 
has seen a surge of interest~\cite{semantickitti,sscbench,occ3d,roldao20223d}. %
SSC
jointly addresses the estimation of both geometry and semantics. 
The %
output is %
typically a voxel grid where each voxel is assigned a label indicating whether it is \emph{free} (i.e., empty) or occupied with a given %
\emph{semantic class}.

In the literature, SSC has been addressed using various kinds of sensor inputs such as cameras~\cite{cao2022monoscene,yu2024cgformer,yao2025depthssc}, lidar~\cite{diffssc}, or radar~\cite{ma2025licrocc}, possibly taking time into account \cite{li2024htcl}.
In this work, we address \emph{SSC from lidar}, %
using a \emph{single scan} as input.
Lidar provides accurate 3D geometry but, unlike a camera, it is sparse and lacks photometric information. This combination makes both sub-problems (geometry and semantics) hard, especially from a single sweep. %
Existing methods range from lightweight BEV completion networks~\cite{lmscnet} to computationally intensive diffusion models~\cite{diffssc}, reflecting a recent trend 
favoring increasingly large, intricate architectures.

A recurring feature %
is the use of auxiliary priors derived from the input, such as semantic or visibility cues.
In existing methods, these priors are usually entangled with the architecture, e.g., produced by a branch trained jointly with the geometric completion head~\cite{js3cnet,sscrs,ssasc}, or introduced at the loss level by leveraging information originating from several frames~\cite{talos}. %
Because each method exploits priors %
differently, the contribution of a particular %
prior %
is not isolated from the specific benefit of the corresponding %
architecture. %

We take a different stance: we supply both kinds of priors (semantic and visibility cues) %
to an existing SSC model %
as additional input data, %
minimally adapting the architecture to accommodate them.
This model is then trained under the same conditions as the original one. 
This approach gives complete freedom to the network to possibly exploit the prior information.
Notably, neither prior comes from a jointly trained branch. 
This lets us measure the impact of each prior on its own.
Moreover, %
keeping %
the source of priors %
decoupled from the completion network %
makes prior cues freely replaceable, e.g., when a better segmenter becomes available. %

In our experiments, both priors are consumed by the completion network at the voxel level. 
The semantic cues are \emph{semantic class} labels predicted per point by a frozen, off-the-shelf point cloud segmenter~\cite{waffleiron,minkunet} trained for the same target classes as the SSC model, then aggregated by majority vote to each voxel containing at least one scanned point. 
A semantically labeled voxel is deemed occupied, while the status of other voxels, which do not contain any points, is \emph{unknown}.
As for the visibility cues, they consist of another label indicating which voxels are expected to be \emph{empty}. %
These labels are %
obtained by casting rays between scanned points and the sensor. 
The status of unobserved voxels remains \emph{unknown}. %

In our study, we also %
bound each prior with a ground-truth oracle, which is not deployable but reveals %
how much of the residual error is geometric versus semantic. %

Equipping four established completion networks with these input priors makes older, lightweight models competitive with current state-of-the-art systems, even surpassing them in one case (\autoref{fig:teaser_journal_3pics}, left), and visibly sharpens their predictions (\autoref{fig:teaser_journal_3pics}, right). 
What is more, this performance is achieved at no architectural cost beyond widening the input layer.

\smallskip
\noindent Our contributions are as follows:
\begin{itemize}[itemsep=3pt,topsep=3pt]
    \item We introduce a simple setting to study the impact of adding semantic and visibility priors into existing SSC networks, in order to improve their performance.
    \item We study the impact of the quality of pseudo-labels as semantic cues, both at train and test time, and show it is enough to train the SSC network once with ground-truth labels, then use any semantic segmenter at inference.
    \item 
    We explore the design space of visibility cues and show the gain persists across lidar densities.%
    
    \item Our experiments, spanning two datasets and four existing architectures, %
    show that the semantic and geometric performance of SSC models can be systematically boosted with our priors. In this setting, older SSC methods can even outperform current state-of-the-art models, shedding new light on the performance bottlenecks of SSC approaches.
\end{itemize}

\textcolor{purple}{
This article is an extended version of our conference paper \emph{``Exploring Easy Boosts for Lidar Semantic Scene Completion''} \cite{martyniuk2026exploring}. Additions to the conference paper are described %
in \autoref{app:conf_paper_extension}. 
}

The remainder of the paper is organized as follows. 
\Cref{sec:related} reviews related work on lidar and camera SSC, prior injection, and ray-based visibility reasoning. 
\Cref{sec:method} serves as an ablation study and analyzes %
each prior in isolation on SemanticKITTI. 
\Cref{sec:experiments} evaluates the combination of the best semantic and visibility cues across four different architectures and two datasets (SemanticKITTI and SSCBench-nuScenes), establishes segmenter-agnostic deployment, and analyzes where the gains concentrate and how much headroom the oracle priors leave.
\Cref{sec:conclusion} concludes the study. 
\hyperref[app:conf_paper_extension]{The Appendices} report visibility prior precompute profiling, and detail additional results, e.g., classwise segmentation tables. %

%% file: figures/journal_teaser.tex
\begin{strip}
    \centering
    \small
    \begin{minipage}[t]{0.40\linewidth}
        \vspace{0pt}%
        \centering
        \includegraphics[width=\linewidth]{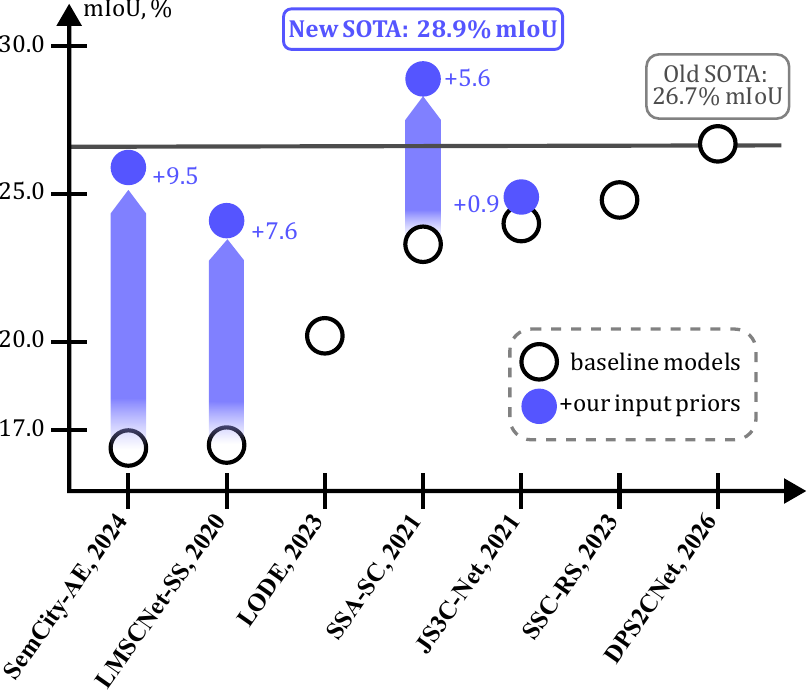}
    \end{minipage}%
    \hfill
    \begin{minipage}[t]{0.59\linewidth}
        \vspace{0pt}%
        \centering
        \newcommand{\teaserimg}[1]{\includegraphics[width=2.6cm,trim={291.1bp 27.1bp 291.7bp 28.9bp},clip]{#1}}
        \newcommand{\teaserimgcrop}[2]{\includegraphics[width=2.6cm,trim={#1},clip]{#2}}
        \newcommand{\teasercell}[2]{#1}
        \definecolor{teaserblue}{HTML}{5555FF}%
        \resizebox{\linewidth}{!}{%
        \begin{tikzpicture}[
            cell/.style={draw=black!50, minimum width=2.6cm, minimum height=2.6cm, inner sep=0pt, align=center, font=\footnotesize},
            boost/.style={->, >=latex, line width=1.6pt, draw=teaserblue, shorten <=2pt, shorten >=2pt},
            pcell/.style={cell, draw=teaserblue, line width=1pt},%
            colhead/.style={above=2pt, font=\normalsize},
            rowhead/.style={rotate=90, anchor=south, align=center, font=\normalsize},%
            group/.style={inner sep=4pt},%
        ]
            \node[cell]  (sc-b)  at (0,0)   {\teaserimg{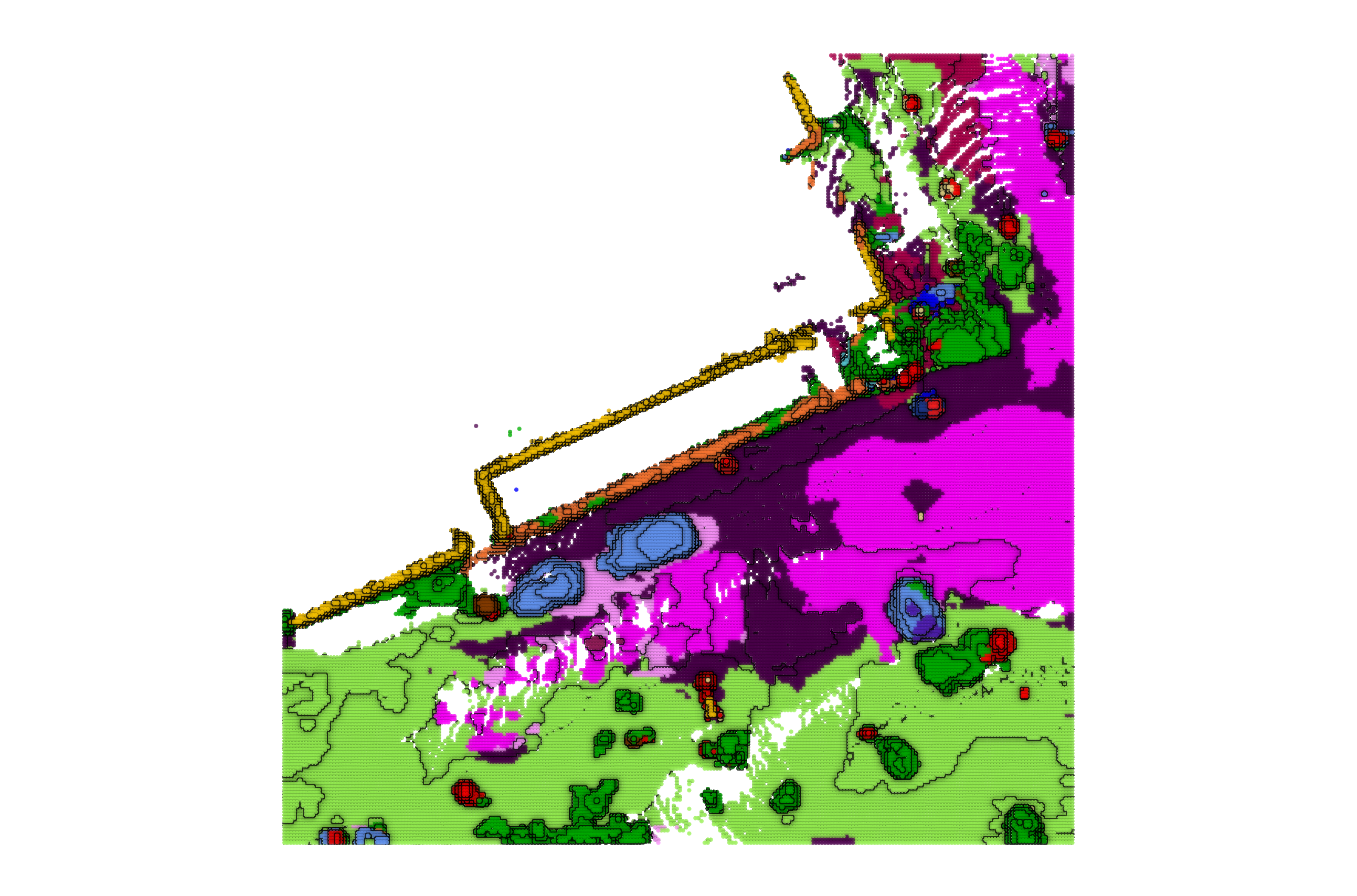}};
            \node[pcell] (sc-p)  at (3.5,0) {\teaserimg{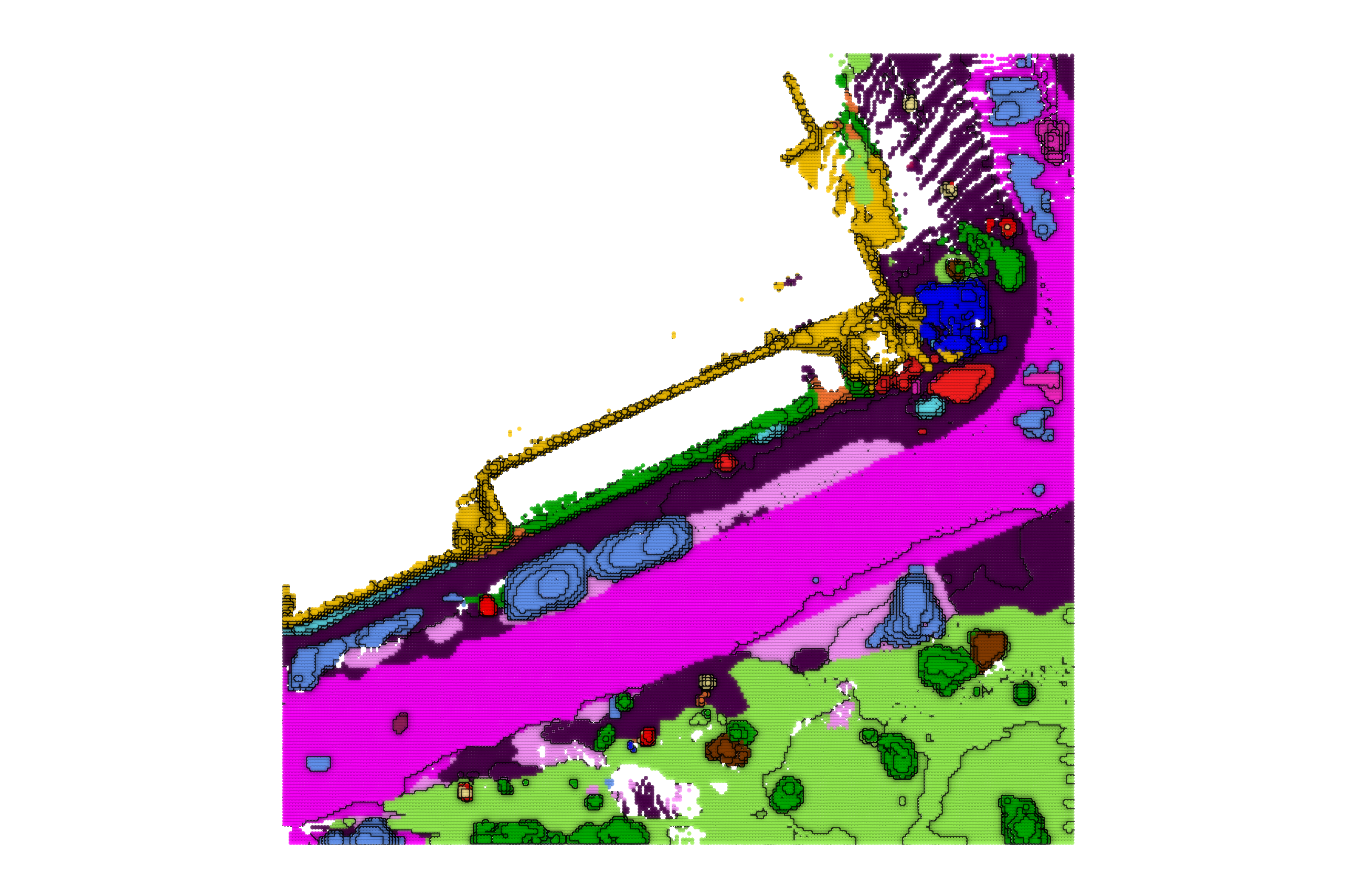}};
            \node[cell]  (sc-gt) at (6.6,0) {\includegraphics[width=2.6cm,trim={161.2bp 28.9bp 161.8bp 28.9bp},clip]{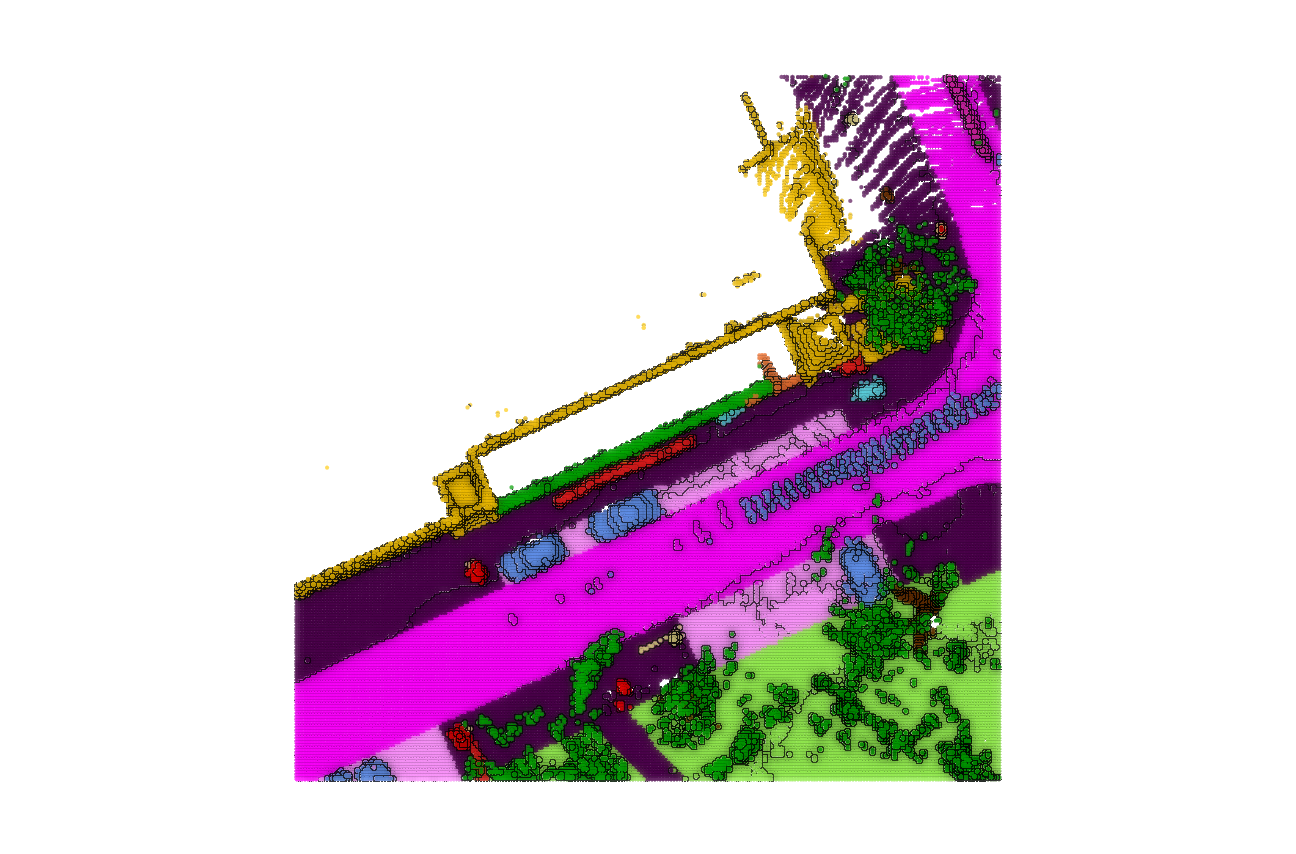}};
            \node[cell]  (ssa-b)  at (0,-2.9)   {\includegraphics[width=2.43cm]{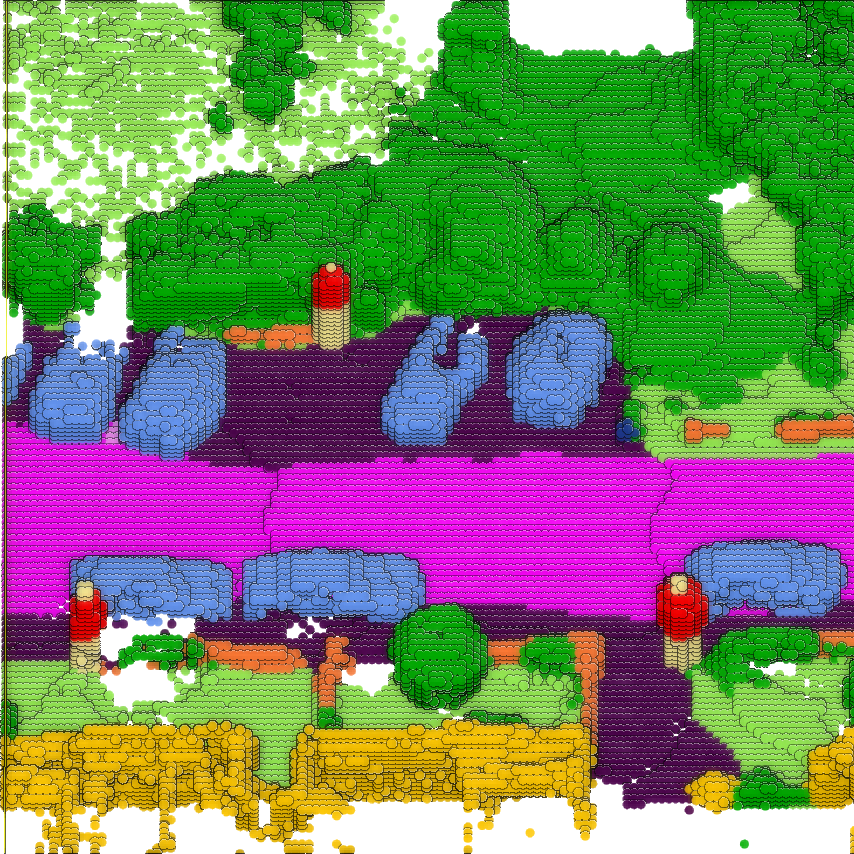}};
            \node[pcell] (ssa-p)  at (3.5,-2.9) {\includegraphics[width=2.43cm]{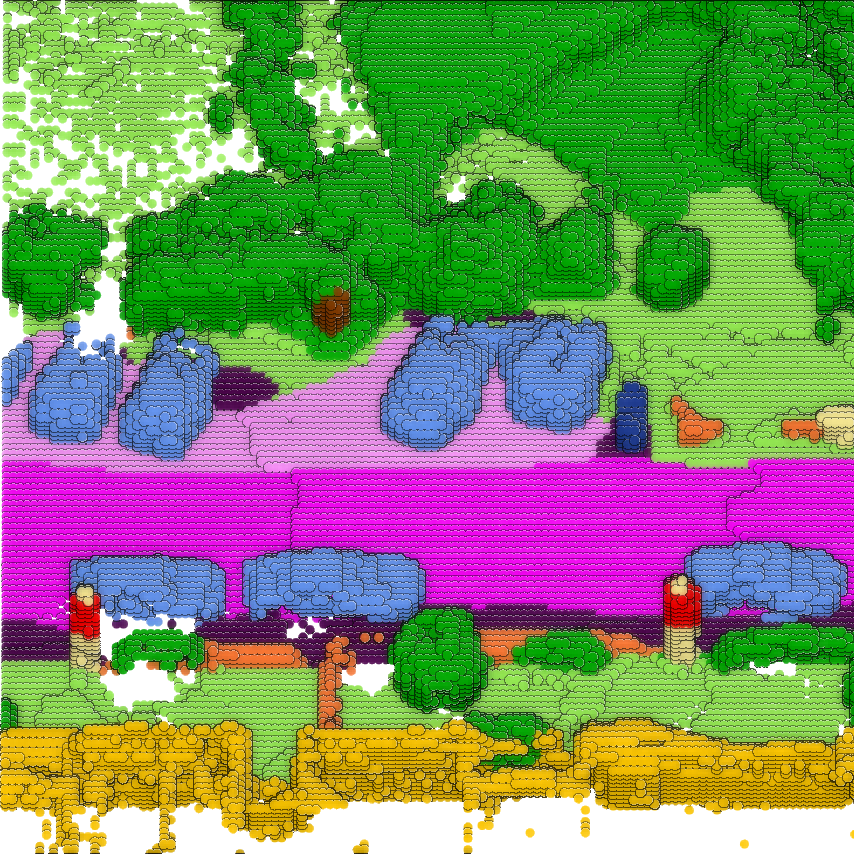}};
            \node[cell]  (ssa-gt) at (6.6,-2.9) {\includegraphics[width=2.43cm]{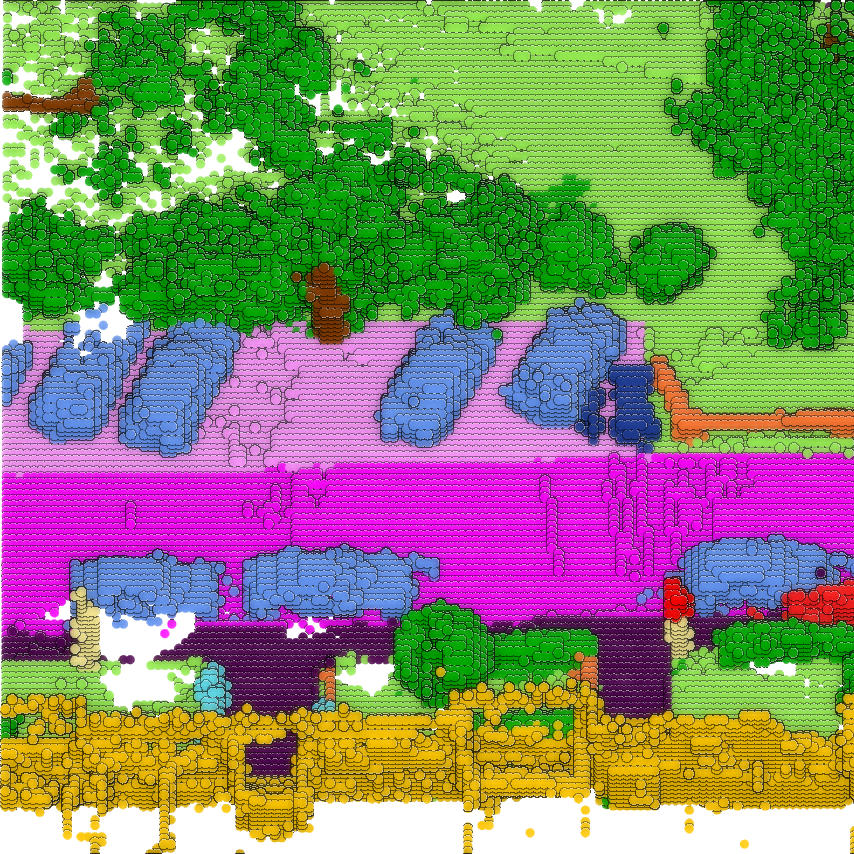}};
            \foreach \mdl in {sc, ssa} \draw[boost] (\mdl-b.east) -- (\mdl-p.west);
            \node[colhead]        at (sc-b.north)  {baseline};
            \node[colhead]        at (sc-p.north)  {+ our priors};
            \node[colhead] (h-gt) at (sc-gt.north) {GT};
            \node[rowhead] (l-sc)  at ($(sc-b.west)  + (-4pt,0)$) {\textbf{SemCity-AE}\\(biggest gain)};
            \node[rowhead] (l-ssa) at ($(ssa-b.west) + (-4pt,0)$) {\textbf{SSA-SC}\\(best final\\performance)};
            \node[group, fit=(l-sc)(l-ssa)(h-gt)(ssa-gt)] (g) {};
            \draw[black!35, dashed, line width=0.8pt] ($(g.north west) + (-0.3cm,0)$) -- ($(g.south west) + (-0.3cm,0)$);
        \end{tikzpicture}%
        }
    \end{minipage}

    \captionof{figure}{
    \textbf{Impact of priors on
    lidar semantic scene completion (SSC) on SemanticKITTI.} 
    Our systematic study reveals that semantics and visibility are readily-available priors that can significantly boost the performance of single-frame SSC baselines. 
    While conventional lidar SSC relies solely on point locations as input, we augment the latter with semantics, from off-the-shelf 3D segmenters, and visibility information obtained through ray-based sampling to infer empty space. 
    \textbf{Left:} Carefully combining both priors transforms early methods into competing ones and pushes the best-performing method to a new state of the art
    among fully-reproducible single-frame methods.
    \textbf{Right:} Qualitative results on SemanticKITTI for SemCity-AE (top) and SSA-SC (bottom): the baseline, the same model with our input priors, and the ground truth.
    }
    \label{fig:teaser_journal_3pics}
\end{strip}

%% file: sections/002_related.tex
\section{Related work}
\label{sec:related}

\subsection{Semantic Scene Completion}

SSC was originally formulated as a densification task from a single depth image~\cite{sscnet}.
The dual task aims to predict both the geometry (occupancy of occluded regions beyond sensor view) and the semantics of the scene.
The problem has then been studied for various input sensors.

\smallskip
\noindent\textbf{Lidar SSC.}
Enabled by large-scale automotive lidar benchmarks~\cite{semantickitti,sscbench,occ3d}, early approaches like TS3D~\cite{ts3d} were adapted to lidar inputs by processing voxelized TSDF point clouds with 3D CNNs~\cite{semantickitti}. LMSCNet~\cite{lmscnet} introduced dedicated lidar SSC by projecting the point cloud into a 2D Bird's-Eye-View (BEV) representation, treating the height axis as channels to enable lightweight 2D CNN completion. Most subsequent methods adopt discriminative 3D or sparse convolutional backbones~\cite{s3cnet,sscrs,ssasc,js3cnet}, incorporating BEV features~\cite{s3cnet,sscrs} or multi-task auxiliary semantic streams~\cite{completeandlabel,js3cnet,ssasc}.
Moving beyond voxels, Local-DIF~\cite{localdif} encodes the lidar %
with a point network~\cite{pointnet} to learn SSC as an ensemble of local implicit functions, enabling arbitrary-resolution inference.
LODE~\cite{lode} similarly completes the scene implicitly, conditioning an Eikonal signed-distance formulation on local shape priors to handle non-watertight, large-scale lidar scenes.
Recently, diffusion methods~\cite{diffssc,semcity,lee2023diffusion,occtreediff} have framed SSC as conditional generation. 
Lee et al.~\cite{lee2023diffusion} and SemCity~\cite{semcity} keep dense volumetric diffusion tractable using a discrete latent space and triplane representations, respectively. 
DiffSSC~\cite{diffssc} instead diffuses point positions and semantic classes directly in point space, conditioned on a semantically segmented input scan.\footnote{Its public code, however, covers only the geometric part.} 
Octree latent diffusion~\cite{occtreediff} extends the compressed-representation line with a sparse octree structure. 
SemCity further treats SSC as a downstream refinement task conditioned on predictions from existing networks, rather than end-to-end generation.
We do not adopt this refinement: as detailed in \autoref{subsec:exp_setup_section_exp}, its official implementation leaks ground-truth occupancy.

\smallskip
\noindent\textbf{Camera SSC.}
Although we target lidar SSC, parallel progress in camera-based SSC offers highly relevant design patterns. 
MonoScene~\cite{cao2022monoscene} established monocular SSC by lifting 2D features into a 3D voxel grid.
VoxFormer~\cite{li2023voxformer} then introduced a two-stage transformer that uses a depth-derived \textit{visibility/occupancy prior} to seed voxel queries, mirroring our own use of visibility priors on the lidar side.
OccFormer~\cite{zhang2023occformer}, TPVFormer~\cite{huang2023tri}, and SurroundOcc~\cite{wei2023surroundocc} extend camera-based occupancy prediction to multi-view and tri-perspective settings, and Symphonies~\cite{jiang2024symphonize} further explores occlusion-aware design.

\smallskip
\noindent\textbf{Multimodal SSC}
has also been investigated.
Here lidar, camera and radar may be used jointly, possibly using several successive frames~\cite{ma2025licrocc, Cao2024SLCFNetSLA, Wang2025L2COccLCA, Lu2024LiDARCameraCFA, Wang2024LearnableFSA}.
While not directly comparable to single-frame lidar SSC, these methods confirm a broader trend: explicit modeling of what is observed, versus what remains unobserved, is increasingly recognized as a primary lever for occupancy prediction, regardless of input modality.

\subsection{Priors in SSC}
\label{subsec:related-priors}

To ease joint geometric and semantic prediction, several methods leverage auxiliary priors.

\smallskip
\noindent\textbf{Semantic priors} are most commonly introduced through explicit multi-task supervision. 
SSA-SC~\cite{ssasc} and JS3C-Net~\cite{js3cnet} couple the completion network with a learned point-cloud segmentation branch, whereas S3CNet~\cite{s3cnet} fuses a separately-trained 3D completion network and 2D BEV variant post hoc.
JS3C-Net additionally uses the completion output as a contextual shape prior fed back into segmentation.
This kind of architecture tightly couples completion performance to the specific architecture and training of the joint segmenter: replacing or upgrading the segmenter requires retraining the completion network.
SSC-RS~\cite{sscrs} disentangles the two tasks, separating semantic and geometric representations before fusing them in BEV.
Taking an alternative approach to auxiliary guidance, SCPNet~\cite{scpnet} leverages dense-to-sparse knowledge distillation, transferring semantic and structural relations from a multi-frame teacher to regularize the single-frame student.

\underline{\textit{Our study}} focuses on the systematic injection of an off-the-shelf, frozen segmenter directly into the input tensor to quantify its exact performance headroom. 
While sequential or conditional methods like TS3D~\cite{ts3d}, DiffSSC~\cite{diffssc}, and LODE's~\cite{lode} semantic extension have previously leveraged decoupled semantic predictions, we contrast explicitly with multi-task methods that lock completion performance to a jointly trained internal branch. 
This decoupling not only allows us to establish clear oracle bounds, but also proves segmenter-agnostic transferability (see \autoref{sec:agnostic}).

\smallskip
\noindent\textbf{Visibility priors} in lidar SSC are seldom used.
TALoS~\cite{talos} performs ray casting from multiple lidar frames and uses the resulting visibility map to drive test-time adaptation of the completion head; it requires temporal sequences (including future frames) and per-sequence optimization. %

For camera-based SSC,
visibility has been more actively studied: VoxFormer~\cite{li2023voxformer} 
estimates camera depth to seed sparse voxel queries,
and recent works further decouple the prediction of visible and occluded regions either at the architectural~\cite{lu2025vishall3d} or supervision~\cite{han2025voic} level.

\underline{\textit{Our study}} operates on single frames, with no test-time adaptation, and trains the network only once, before inference, to exploit visibility information, as opposed to \cite{talos}.
Also, we exploit visibility at \textit{input} level, as a precomputed signal consumed by the SSC network, requiring only a minor adaptation of the first input layers.
To our knowledge, this is the first study of explicit empty-space visibility priors decoupled from the completion architecture in single-frame lidar SSC.

\subsection{Lidar semantic pseudo-labels}

Three-dimensional semantic segmentation has matured rapidly through point-based~\cite{pointnet}, sparse-convolutional~\cite{minkunet,zhou2020cylinder3d}, BEV-projection~\cite{waffleiron}, and transformer-based~\cite{wu2024point, lai2023spherical} architectures.
State-of-the-art segmenters now exceed $70\%$ mIoU on SemanticKITTI-lidarseg~\cite{wu2024point, waffleiron}, well above the SSC mIoU of current SSC methods. 
We quantify this gap directly in \autoref{tab:lidar_pseudo_labels} via oracle experiments.
Said gap motivates our use of a strong off-the-shelf segmenter as a frozen teacher whose predictions are fed as part of the input of the SSC model, 
contrary to the internally coupled multi-task methods discussed above (\autoref{subsec:related-priors}).

Outside SSC, the pattern of consuming frozen-teacher predictions as input features for a downstream 3D task is established by PointPainting~\cite{vora2020pointpainting}, which decorates lidar points with frozen 2D-segmenter scores for 3D object detection. Related ideas appear in pseudo-label self-training for cross-domain 3D detection~\cite{yang2021st3d}.

\underline{\textit{Our study}} 
raises an open question: how does the quality of the pseudo-label teacher at inference interact with the quality of the labels seen at training time?
In fact, we show that an SSC network trained with ground-truth semantic input generalizes well to inference-time pseudo-labels, %
eliminating the need to retrain the completion network whenever a stronger semantic teacher becomes available.

\subsection{Ray-based visibility reasoning}
\label{subsec:relworkvisib}

Lidar point clouds are commonly processed as unordered point sets, discarding the sensor geometry.
However, the line of sight between the %
sensor and %
each return carries information about traversed free space and occluded regions.

Classical occupancy-mapping methods %
exploit this directly via ray casting, %
e.g., in OctoMap framework~\cite{hornung2013octomap}.
Rold{\~a}o et al.~\cite{roldao} refine the standard binary per-scan update by modeling the distance each ray traverses within a cell and down-weighting observations by a range-dependent density term. It helps reduce %
the discretization artifacts, in particular %
spurious free space from grazing or distant rays %
that arise when a cell's state is set from a single coarse traversal.
In learned 3D perception, Hu et al.~\cite{hu2020you} showed that visibility features substantially boost lidar object detection.

Ray-based reasoning has been successfully leveraged for map cleaning and dynamic-object removal~\cite{removeRevert, lim2021erasor} or as a pretraining objective for lidar backbones~\cite{boulch2023also}.
More recently, explicit ray-based supervision has also been applied to volumetric occupancy prediction in the camera setting, such as in RenderOcc~\cite{pan2024renderocc} and UniOcc~\cite{pan2023uniocc}, where rendered rays supervise the voxel field.
Closer to our setting, Sulzer et al.~\cite{sulzer2022deep} augmented point clouds with sensor-visibility information for surface reconstruction, demonstrating that input-level visibility can be consumed by downstream networks with minimal architectural adaptation.
While early methods like TS3D~\cite{ts3d} and S3CNet~\cite{s3cnet} incorporated visibility inherently through (flipped) TSDF inputs, explicit ray-based free-space priors have otherwise received little attention in recent single-frame lidar SSC architectures (cf.\ \autoref{subsec:related-priors}).

\underline{\textit{Our study}} explores various efficient visibility representations and evaluates their impact when provided at input level of the SSC network.

%% file: sections/003_recipe.tex
\section{Study of individual priors}
\label{sec:method}
\input{figures/method}

In this section, after defining our simple way to integrate raw priors into existing lidar SSC architectures (\autoref{subsec:integrpriors}) and presenting our experimental setup (\autoref{subsec:setup}), we study two input-level priors that boost SSC performance and can be applied to any existing completion network with only minor input-layer adaptations: a \emph{semanticized} input point cloud (%
\autoref{subsec:semantic_prior}) and \emph{visibility} information 
representing the emptiness of the lines of sight between the lidar and the scanned surfaces
(%
\autoref{subsec:vis}).

We analyze each prior in isolation, contrasting realistic off-the-shelf prior estimators against ground-truth \emph{oracles}.
The oracles are not deployable, but they bound the performance achievable with a perfect prior and thus reveal the remaining room for improvement with respect to the prior. %
We combine the two priors and evaluate their association %
in \autoref{sec:experiments}.

\subsection{Integration of raw priors in a lidar SSC architecture}
\label{subsec:integrpriors}

Formally, lidar SSC takes a sparse lidar point cloud $\mathbf{P}$ and outputs a completed volume in which every voxel is assigned one of the $C+1$ labels: $C$ %
semantic classes, and the \emph{empty} (or \emph{free}) class.

Most SSC methods first voxelize the point cloud of the input scan into a grid on which the backbone operates. 
This raw data %
supports only two states per voxel: a voxel is \emph{occupied} if it contains at least one lidar point %
and \emph{unknown} otherwise. 
In the second case, the sensor provides no evidence that an empty-looking voxel is truly free rather than merely unobserved. We write this input as a per-voxel categorical grid
\[\mathbf{V}\in\mathcal{S}^{X\times Y\times Z},\qquad \mathcal{S}=\{\text{occupied},\ \text{unknown}\},\]
\textit{i.e.}, $|\mathcal{S}|=2$ states, over a 3D voxel grid of dimensions $X\times Y\times Z$.

Our ``recipe'' to add priors enriches this state set at the \emph{input} only, adding no new module and training no auxiliary network. Each prior we study enlarges $\mathcal{S}$ along one axis: the semantic prior (\autoref{subsec:semantic_prior}) refines the bare \emph{occupied} state into $C$ semantic classes, and the visibility prior (\autoref{subsec:vis}) splits \emph{unknown} into (observably) \emph{empty} and \emph{unknown}. Their combination (\autoref{sec:experiments}) yields a %
$C+2$ state set. Feeding this categorical grid to a backbone requires only a backbone-specific input adaptation: widening the first layer to a one-hot encoding of $\mathcal{S}$ for grid-convolutional networks, a learned per-state embedding, or a concatenation to the per-point features that a point backbone already ingests. No other architectural change is needed.
For a backbone that does not consume a single grid input (\textit{e.g.}\ JS3C-Net~\cite{js3cnet}), the two components of $\mathbf{V}$ (the semantic state on occupied voxels and the empty/unknown partition) need not enter as one tensor: each is routed to the pathway the architecture already exposes.  (Per-architecture details are in \autoref{subsec:setup_4_arch}).

Without loss of generality, we present below the priors based on this voxel-grid input representation. It however directly extends to a point-based backbone: the semantic prior then adds semantic classes to input points rather than voxels, and the visibility prior adds free-space information as extra points marked \emph{empty} in the point cloud. %

\subsection{Experimental setup}
\label{subsec:setup}

\subparagraph{Dataset}

We conduct our prior %
studies on the SemanticKITTI validation set (sequence~08)~\cite{semantickitti}, using the official SemanticKITTI ground-truth voxel grid of $256\times256\times32$ at $0.2$\,m resolution with $C=19$ semantic classes.

\subparagraph{Metric}

We report two complementary metrics. 
\emph{Geometric completion} is measured with the binary occupied-vs-empty voxel intersection-over-union (IoU). 
\emph{Semantic completion quality} is measured with the mean intersection-over-union (mIoU) over the $C$ semantic classes (thus excluding free space), where the IoU for each individual semantic class is computed over the voxels that are not empty in at least the ground truth (GT) or the prediction. 
(Unobserved areas, as defined in the GT, are excluded from the computation of both metrics).
Beyond the dual
nature of the SSC task, reporting the two metrics separately is central to our analysis, since each prior is expected to act primarily on one of the two subtasks.

\subparagraph{Augmented SSC methods}

We run the study on individual priors with two lightweight networks: LMSCNet-SS~\cite{lmscnet} and the SemCity autoencoder~\cite{semcity} repurposed for SSC (SemCity-AE).
Both are chosen deliberately for being efficient, yet not state-of-the-art (SOTA) regarding both the semantic and geometric metrics (%
5 to 7 IoU pts and about 10 mIoU pts below the best reproducible method, DPS2CNet, cf.~\autoref{tab:semkitti_sota}),
so that the effect of an input prior is clearly observable rather than partially masked by an already strong backbone.
We however later demonstrate that our findings translate to stronger backbones (\autoref{sec:experiments}). %

For a controlled comparison, 
all variants of LMSCNet-SS are %
trained for 80 epochs with its original scheduler, and all variants of SemCity-AE are trained for 50 epochs with a cosine-annealing scheduler. %

\subsection{Semantic prior}
\label{subsec:semantic_prior}

Some %
SSC methods exploit the ground-truth (GT) semantics of the input point cloud (or its voxelized representation) %
by training a separate, explicit segmentation branch jointly with the completion network, either as a first stage in the pipeline~\cite{js3cnet} or as a parallel branch interacting with completion~\cite{ssasc}.
Lidar semantic segmentation is, however, a mature field with strong networks available off-the-shelf.
We therefore isolate semantics as an individual prior: a frozen,
independently pretrained segmenter supplies per-point labels, which we voxelize directly into the input tensor~$\mathbf{V}$. %

\paragraph*{Point labels as semantic prior}

Concretely, the single \emph{occupied} state is refined into $C$ semantic classes, and the new input voxel grid $\mathbf{V}_{\text{w/sem}}$ is built as follows.
All voxels are first set to the \emph{unknown} state.
Every voxel containing at least one lidar point is then considered as %
\emph{occupied} and assigned a single semantic class by majority vote over the points it contains.
Every non-occupied voxel remains \emph{unknown}. This enlarges the state set to $\mathcal{S}_{\text{w/sem}}=\{1,\dots,C, %
\text{unknown}\}$, \textit{i.e.}, $C+1$ states. %
Therefore, the only work to create the variant based on the semantic prior is to increase from 2 to $C\,{+}\,1$ the number of channels representing the state, %
with no change to the downstream architecture.

Please note that semantic information is very sparse in the voxel grid: it only concerns the semantics of the points in the input frame, not the completed volume in the GT constructed from accumulated frames.

In the following, unless otherwise stated, the adapted network is both trained and tested with the same semantics-augmented input.

\input{tables/lmscnet_pseudo_labels}

\paragraph*{Impact of pseudo-labels on SSC semantics}

To create our semantic prior, %
the input scan is first segmented by an off-the-shelf semantic segmentation network, %
before being fed to the input-widened SSC network.
\autoref{tab:lidar_pseudo_labels} reports results with (1)~vanilla SSC models (\textit{i.e.}, with no semantic labels at input) and using four segmenters or variants to create pseudo-labels: (2)~SalsaNext~\cite{salsanext}, %
(3)~WaffleIron~\cite{waffleiron} (WI), %
(4)~MinkUNet~\cite{minkunet}, %
and (5)~WaffleIron with test-time augmentation (WI-TTA), %
with input label quality ranging from $55.8\%$ to $70.3\%$.
\input{figures/monotonic-segmenters}

We observe that the pseudo-labels provided on the input frame lead to a systematic semantic completion %
gain: $+4.8$ to $+5.8$ mIoU pts for LMSCNet-SS, and $+6.6$ to $+9.6$ pts for SemCity-AE. This is a very significant gain considering also that it relies only on an extremely simple extension of the input to an SSC model, and retraining.

The benefit is moreover robust to the choice of segmenter, as different as they may be (\textit{e.g.}, 2D dense convolution on a BEV projection for WaffleIron vs 3D sparse convolution for MinkUNet). 
Completion mIoU broadly increases with the mIoU of the semanticized input frame (see \autoref{fig:monotonic-segmenters}, left), although segmenters of comparable input quality are not separated: MinkUNet and WaffleIron+TTA differ by 0.3 input mIoU pt and rank differently on the two networks.

\input{tables/lmscnet_visibility_ablation}

\paragraph*{Oracle headroom}
To upper-bound the gain attainable from such an input semantics, we also consider constructing $\mathbf{V}_{\text{w/sem}}$ from the GT-annotated input points. %

Table~\ref{tab:lidar_pseudo_labels} (rows~(6)) shows that the effect is almost entirely on the semantic axis: mIoU rises by $+13.9$ pts for LMSCNet-SS (to $30.4$\% mIoU) and $+16.3$ pts for SemCity-AE (to $32.7$\% mIoU). %

We observe that the pseudo-labels used as augmented input recover a large fraction of the semantic oracle gain. Yet, a substantial share of the semantic completion headroom remains unrealized: +8.1 mIoU pts for LMSCNet-SS (after already gaining $+5.8$ mIoU pts with pseudo-labels), and $+6.7$ mIoU pts for SemCity-AE (after gaining +9.6 pts). These SSC gaps with the oracle are attributable to the remaining ${\sim}30$-pt gap in input mIoU between the best off-the-shelf segmenter used here (70.3\% input mIoU) and the perfect input segmentation. Thanks to the modular nature of our pipeline, part of this gap can however be recovered ``for free" when external segmenters improve, with no change to the completion network.

Experiments in \autoref{sec:experiments} also show that, for simplicity, instead of retraining with the pseudo-labels of a new segmenter, it is even enough to train just once with the semantic oracle and then to infer with pseudo-labels from any segmenter.

\paragraph*{Impact of the semantic prior on geometry}

While we may expect some kind of coupling between semantics and geometry (the historical motivation of SSC~\cite{sscnet}), the effect of semantic pseudo-labels on geometric completion is small and even occasionally slightly negative: between $-0.9$ and $+0.5$ IoU pt. %

With oracle semantic labels, the effect on completion is positive but remains small: the IoU gains $+0.7$ IoU pt on both SSC networks. 
In fact, the IoU of these models remains at $56.3$ and $54.6$\%, which is still some $4.5$ to $6$ pts below DPS2CNet~\cite{liu5333789dual}, the strongest reproducible method in geometry ($60.8$\% IoU, cf.\ \autoref{tab:semkitti_sota}).

\paragraph{Conclusion}

This study of LMSCNet-SS and SemCity-AE networks shows that, thanks to an extremely simple semantic extension of the input to an SSC model, and retraining, it is possible to obtain significant SSC gains regarding semantics, up to +9.6 mIoU pts. 
While input semantics, if not too noisy, naturally is a dominant driver of the quality of SSC semantics, it however has little impact on the quality of SSC geometry.

Geometric completion thus seems to be chiefly insensitive to semantic quality and is the subject of the visibility prior, discussed below (Sec.~\ref{subsec:vis}).

\subsection{Visibility prior}
\label{subsec:vis}

The 3D points of a single lidar scan only provide very sparse information to reconstruct a complete scene, which makes SSC a particularly difficult task. 
Yet, lidar point clouds provide more information when the sensor location is known, which is generally the case. %
Rather than uniformly consider all space besides observed points as \emph{unknown}, it is then possible to distinguish observed free space (between the sensor and scanned points) from unobserved free space (behind scanned points, between scan rays).
Early lidar SSC methods already encode this distinction implicitly through (flipped) TSDF inputs~\cite{ts3d,s3cnet}, and TALoS~\cite{talos} exploits ray-cast visibility at test time over multiple frames.
We now study an explicit visibility prior that recovers this distinction %
from the lidar lines of sight.

Given a single lidar scan, we cast one ray for each lidar point, between %
the sensor origin and the %
measured 3D point, and place along each ray a number of special 3D points as \emph{free-space markers}.
Markers and lidar points are voxelized into the $256\times256\times32$ grid $\mathbf{V}_{\text{w/vis}}$ by majority vote, with lidar points %
taking precedence over free-space markers within any shared voxel.
The result is a ternary grid: a voxel is (i)~\emph{empty} %
if it contains a free-space marker (hence is traversed by at least one ray) and %
no %
lidar point, (ii)~\emph{occupied} if it received a lidar point, %
or (iii)~\emph{unknown} %
otherwise. %
This enlarges the input states to $\mathcal{S}=\{\text{occupied},\ \text{empty},\ \text{unknown}\}$. %
The visibility prior thus keeps the \emph{occupied} voxel state untouched, but splits the baseline \emph{unknown} voxel state into ``observably %
\emph{empty}'' %
and ``remaining \emph{unknown}''. %

Note that, as for semantic information, this visibility prior can be imperfect. Beyond possible noise in lidar scans (reflections, missing returns, ego vehicle speed compensation, etc.), there can also be lidar registration issues that impact the accumulation of scans when building an SSC ground truth and create inconsistencies with free spaces of single scans, which are rarely taken into account when constructing such a GT.

As in \autoref{subsec:semantic_prior}, we study here the visibility prior in isolation: occupied voxels only carry bare occupancy, not semantic classes. 
Combining semantic and visibility priors is addressed in \autoref{sec:experiments}.

\paragraph{Impact of visibility marker distribution}
The construction above leaves open how densely and where 
free-space markers are placed along each ray, offering several variants. %
Different variants trade precompute cost against the fidelity of the recovered free space (see discussion below). %

In \autoref{tab:lidar_vis_ablation}, we compare a variety of marker distributions: (1)~none, \textit{i.e.}, vanilla model with no prior; a single marker placed (2)~just before the scanned point, %
or (3)~at a random position along the ray; (4-7)~$n$ markers at random positions with $n$ respectively in $\{10, 25, 50, 100\}$; and (8)~a dense uniform %
marker sampling that places a marker every voxel size ($0.2$\,m in our experiments) on the ray, %
(9)~optionally followed by a safety margin of $5\times5\times5$ voxels around each lidar point, that demotes near-surface traversed voxels back to the default unknown state (``dense with dilation''\footnote{named ``our visibility prior'' in our previous work~\cite{martyniuk2026exploring}.} in \autoref{tab:lidar_vis_ablation}), creating a kind of safety volume to minimize discretization artifacts caused by grazing rays slightly penetrating occupied voxels. %
(We tried a few more heuristics but every variant we tested merely slides along a precision/recall frontier). 
The dense uniform variant, which is our default setting, 
has about 60 points per ray on average.

All variants improve completion over the visibility-free baseline. 
From 10 markers per ray on, they lie within 0.7 IoU pt of each other on both networks: having enough markers matters more than where they are placed.
Unlike the random placements, the dense uniform variant is deterministic, and its empty voxels agree best with their ground-truth-denoised version (\autoref{fig:monotonic-segmenters}, right).
It reaches 57.5\% IoU on LMSCNet-SS and 56.9\% on SemCity-AE, \textit{i.e.}, $+1.9$ and $+3.0$ IoU pts over the prior-free models.

\paragraph*{Oracle headroom}

To bound how much accuracy %
single-scan visibility leaves unrealized, we evaluate a %
\emph{denoising oracle} that runs the same dense uniform ray casting, but uses GT only to purge false-empty markers, relabeling them as \emph{unknown}. It typically corresponds to cases of %
grazing rays that clip a surface. %

Using this oracle as visibility prior provides a significant gain on geometric completion, with $+8.0$ and $+6.7$ IoU pts on LMSCNet-SS and SemCity-AE, respectively, compared to the vanilla models. This leaves plenty of room for improvement over the simple dense uniform sampling, with $+6.1$ and $+3.7$ IoU pt margin, respectively.

\paragraph{Impact of the visibility prior on semantics}

We observe a mild but consistent improvement on semantics when using a visibility prior: up to $+1.6$ and $+1.1$ mIoU pts using dense cues on LMSCNet-SS and SemCity-AE, respectively. The effect is a bit higher than the slight occupancy improvement observed when, conversely, exploiting the semantic prior (\autoref{subsec:semantic_prior}): occupancy indirectly helps more the semantics than the other way around.

Denoising the dense visibility prior with an oracle further benefits semantics, achieving an extra $+1.5$ and $+2.5$ mIoU pts on LMSCNet-SS and SemCity-AE, respectively, to $+3.1$ and $+3.6$ mIoU pts with respect to the vanilla models. The virtual gain on semantics of this oracle geometric information is not negligible.

\paragraph*{Conclusion}

These experiments show that the information of lidar lines of sight, exploited purely at the input to recover the empty/unknown distinction, %
yields a consistent completion boost (up to $+3.0$ IoU pts), alongside a mild semantic gain, and at no architectural cost beyond an extra input channel. %

While the visibility prior has slightly more impact on semantics than the semantic prior has on geometric completion, the coupling of semantics and occupancy remains weak and the two input priors are chiefly complementary. %
We thus expect the combination of the semantic and visibility priors to compound rather than overlap.

In the following section (\autoref{sec:experiments}), we therefore combine both priors into a single augmented input and evaluate the resulting recipe, which involves a full $C+2$ state set that carries, on every voxel, both the semantic class on occupied voxels and the empty/unknown visibility partition on the other voxels.

%% file: figures/method.tex
\begin{figure*}[t]
    \centering
    \includegraphics[width=\linewidth]{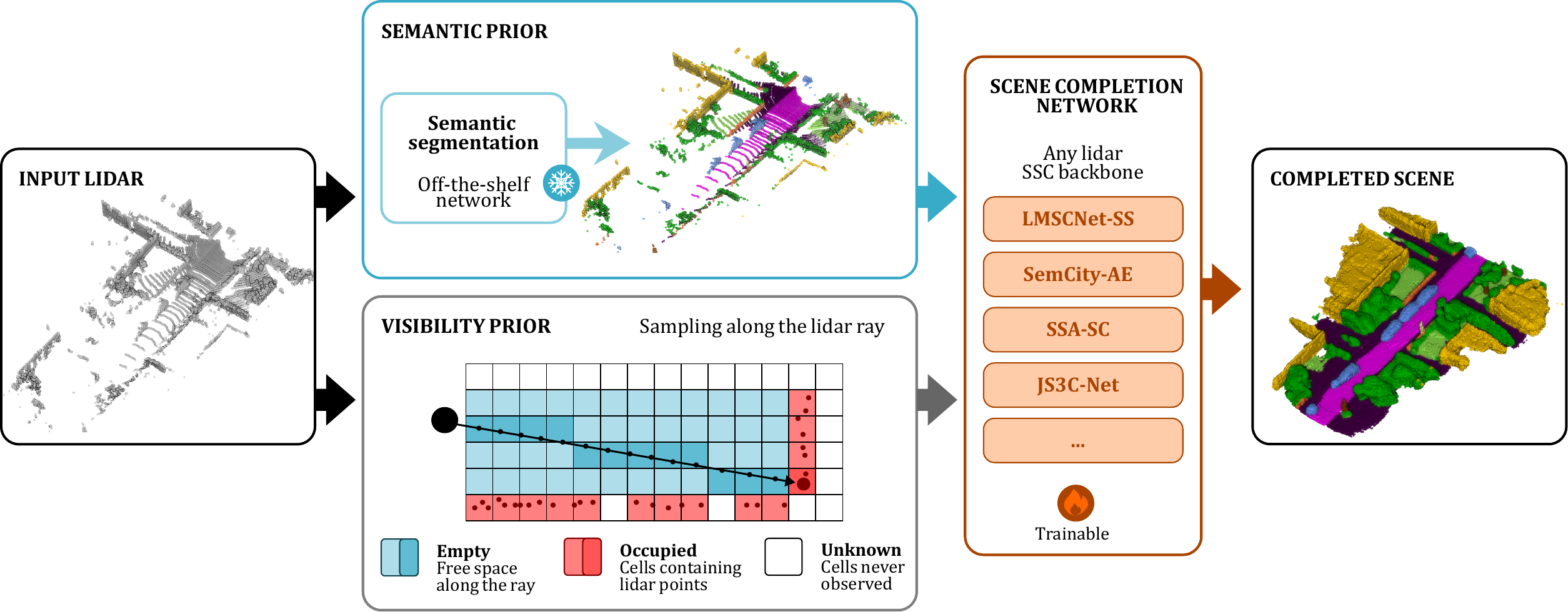}
    \caption{\textbf{Method overview}. 
    Given a single lidar scan, we build an augmented input grid $\mathbf{V}\in\mathcal{S}^{X\times Y\times Z}$ over the full $C+2$ state set $\mathcal{S}=\{1,\dots,C,\text{empty},\text{unknown}\}$, as detailed in \autoref{subsec:integrpriors}.
    \emph{Semantics} comes from a frozen, off-the-shelf point cloud segmenter; \emph{visibility} separates empty from unknown voxels via single-scan ray casting. 
    Both enter the completion network only at the input.
    }
    \label{fig:method}
\end{figure*}

%% file: tables/lmscnet_pseudo_labels.tex
\begin{table}[t]
    \caption{\textbf{Semantic prior} effect on LMSCNet-SS and SemCity-AE %
    for SemanticKITTI, with various sources of semantic labels. Labels are used both at train and test time; TTA only at inference. $\Delta$~measures the difference with the vanilla model (no semantic labels).
    $^\dagger$baseline is retrained.
    }
    \label{tab:lidar_pseudo_labels}

    \setlength{\tabcolsep}{2pt}
    \centering
    \setlength{\aboverulesep}{0pt}
    \setlength{\belowrulesep}{0pt}
    
    \begin{tabular}{c|H@{}l@{}r|c@{~}c|c@{}c}
    \toprule
    \rule{0pt}{1.1em} & & \multicolumn{2}{c|}{1-frame input} &  \multicolumn{4}{c}{SSC output~~} \\[2pt]
    & & Semantics & &  \multicolumn{2}{c|}{geometry}  & \multicolumn{2}{c}{\,semantics}  \\
    & & of lidar labels & \llap{mIoU\,\%} & IoU\,\rlap\% & ~~~$\Delta$ & mIoU\,\rlap\% & ~~~$\Delta$ \\
    \midrule
    
    \rowcolor{\colortitle}
    \multicolumn{8}{@{}l}{\textbf{LMSCNet-SS$\vphantom{X^{X^X}}$}~\cite{lmscnet}}\\
    (1) & \xmark & none (vanilla model)$^\dagger$ & --~~ & 55.6 & -- & 16.5 & -- \\
    (2) & \cmark & SalsaNext~\cite{salsanext} & 55.8 & 54.7 & \textcolor{MyRed}{$-$0.9} & 21.3 & \textcolor{MyGreen}{+4.8} \\
    (3) & \cmark & WaffleIron~\cite{waffleiron} & 68.0 & 55.6 & \textcolor{MyGreen}{+0.0} & 21.6 & \textcolor{MyGreen}{+5.1} \\
    (4) & \cmark & MinkUNet~\cite{minkunet} & 70.0 & 55.1 & \textcolor{MyRed}{$-$0.5} & 22.3 & \textcolor{MyGreen}{+5.8}\\
    \rowcolor{\colormain}
    (5) & \cmark & WaffleIron\,+\,TTA~\cite{waffleiron}\, %
    & 70.3 & 55.6 & \textcolor{MyGreen}{+0.0} & 21.9 & \textcolor{MyGreen}{+5.4} \\
    \rowcolor{\colorspecial}
    (6) & \cmark & GT (oracle) & 100.0 & 56.3 & \textcolor{MyGreen}{+0.7} & 30.4 & \textcolor{MyGreen}{+13.9}\\
    \midrule
    
    \rowcolor{\colortitle}
    \multicolumn{8}{@{}l}{\textbf{SemCity-AE$\vphantom{X^{X^X}}$}~\cite{semcity}}\\
    (1) & \xmark & none (vanilla model)$^\dagger$ & --~~ & 53.9 & -- & 16.4 & -- \\
    (2) & \cmark & SalsaNext~\cite{salsanext} & 55.8 & 54.1 & \textcolor{MyGreen}{+0.2} & 23.0 & \textcolor{MyGreen}{+6.6} \\
    (3) & \cmark & WaffleIron~\cite{waffleiron} & 68.0 & 54.4 & \textcolor{MyGreen}{+0.5} & 25.5 & \textcolor{MyGreen}{+9.1} \\
    (4) & \cmark & MinkUNet~\cite{minkunet} & 70.0 & 54.2 & \textcolor{MyGreen}{+0.3}& 25.5 & \textcolor{MyGreen}{+9.1}\\
    \rowcolor{\colormain}
    (5) & \cmark & WaffleIron\,+\,TTA~\cite{waffleiron}%
    & 70.3 & 54.4 & \textcolor{MyGreen}{+0.5} & 26.0 & \textcolor{MyGreen}{+9.6} \\
    \rowcolor{\colorspecial}
    (6) & \cmark & GT (oracle) & 100.0 & 54.6 & \textcolor{MyGreen}{+0.7} & 32.7 & \textcolor{MyGreen}{+16.3}\\
    \bottomrule
    \end{tabular}
    
\end{table}

%% file: figures/monotonic-segmenters.tex
\begin{figure*}[t]
    \centering
    \begin{minipage}[t]{0.48\textwidth}
        \centering
        \includegraphics[width=\linewidth]{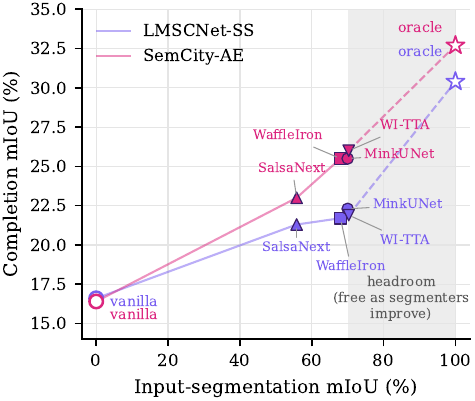}
    \end{minipage}%
    \hfill
    \begin{minipage}[t]{0.48\textwidth}
        \centering
        \includegraphics[width=\linewidth]{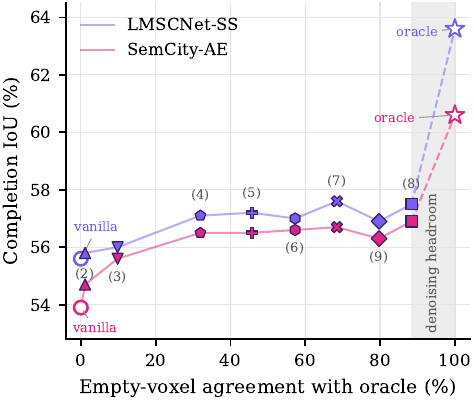}
    \end{minipage}
    \caption{\textbf{Completion quality vs input prior quality} %
    on SemanticKITTI for LMSCNet-SS and SemCity-AE.
    \textbf{Left:} Semantic prior: completion mIoU vs input-segmentation mIoU. 
    Each point is an off-the-shelf segmenter %
    from \autoref{tab:lidar_pseudo_labels} or the GT oracle; the no-semantics (vanilla) baseline sits at $x=0$.
    The shaded region marks headroom recoverable as segmenters improve.
    \textbf{Right:} Visibility prior: completion IoU vs agreement (IoU) between the empty voxels of the input prior and those of the oracle, i.e., dense markers denoised with the ground truth, which sits at $x=100$. Each point is a free-space marker placement, labeled with its row number in \autoref{tab:lidar_vis_ablation}; the no-marker (vanilla) baseline sits at $x=0$.
    The shaded region marks the headroom recoverable by denoising the dense markers.
    }
    \label{fig:monotonic-segmenters}
\end{figure*}

%% file: tables/lmscnet_visibility_ablation.tex
\begin{table}[t]
    \setlength{\tabcolsep}{2pt}
    \centering

    \caption{\textbf{Visibility prior} effect on LMSCNet-SS and SemCity-AE for SemanticKITTI, with variants for placing free-space markers. Markers are used both at train and test time. $\Delta$ measures the difference with the vanilla model (no markers).    
    $^\dagger$baseline is retrained.
    }
    \label{tab:lidar_vis_ablation}

    {
    \begin{tabular}{@{}c|H@{}l|cc|cc@{}}
    \toprule
   & & \multicolumn{1}{c|}{1-frame input} &  \multicolumn{4}{c}{SSC output~~} \\[2pt]
    & & Ray-sampled &  \multicolumn{2}{c|}{geometry}  & \multicolumn{2}{c}{\,semantics}  \\
    & \multicolumn{2}{l|}{free-space markers}   & IoU\,\% & $\Delta$ & mIoU\,\rlap{\%}\, & ~~$\Delta$ \\
    \midrule
    \rowcolor{\colortitle}
    \multicolumn{7}{@{}l}{\textbf{LMSCNet-SS}~\cite{lmscnet}} \\
    (1)& \xmark & none (vanilla model)$^\dagger$ & 55.6  & -    & 16.5  & - \\
    (2)& \cmark & 1 pt before the hit & 55.8 & \textcolor{MyGreen}{+0.2} & 16.9 & \textcolor{MyGreen}{+0.4} \\
    (3)& \cmark & 1 pt at random & 56.0 & \textcolor{MyGreen}{+0.4} & 16.8 & \textcolor{MyGreen}{+0.3} \\
    (4)& \cmark & 10 pts at random & 57.1 & \textcolor{MyGreen}{+1.5} & 17.6 & \textcolor{MyGreen}{+1.1} \\
    (5)& \cmark & 25 pts at random & 57.2 & \textcolor{MyGreen}{+1.6} & 17.8 & \textcolor{MyGreen}{+1.3} \\
    (6)& \cmark & 50 pts at random & 57.0 & \textcolor{MyGreen}{+1.4} & 17.7 & \textcolor{MyGreen}{+1.2} \\
    (7)& \cmark & 100 pts at random & 57.6 & \textcolor{MyGreen}{+2.0} & 17.8 & \textcolor{MyGreen}{+1.3} \\
    \rowcolor{\colormain}
    (8)& \cmark & dense ($\delta\,{=}$\,voxel size) %
    & 57.5 & \textcolor{MyGreen}{+1.9} & 18.1 & \textcolor{MyGreen}{+1.6} \\
    (9)& \cmark & dense with dilation & 56.9 & \textcolor{MyGreen}{+1.3} & 17.3 & \textcolor{MyGreen}{+0.8} \\
    \rowcolor{\colorspecial}
    (10)& \cmark & oracle (dense denoised) & 63.6 & \textcolor{MyGreen}{+8.0} & 19.6 & \textcolor{MyGreen}{+3.1} \\
    \midrule
    \rowcolor{\colortitle}
    \multicolumn{7}{@{}l}{\textbf{SemCity-AE}~\cite{semcity}} \\
    (1)& \xmark & none (vanilla model)$^\dagger$ & 53.9  & -    & 16.4  & - \\
    (2)& \cmark & 1 pt before the hit & 54.7 & \textcolor{MyGreen}{+0.8} & 16.8 & \textcolor{MyGreen}{+0.4} \\
    (3)& \cmark & 1 pt at random & 55.6 & \textcolor{MyGreen}{+1.7} & 17.5 & \textcolor{MyGreen}{+1.1} \\
    (4)& \cmark & 10 pts at random & 56.5 & \textcolor{MyGreen}{+2.6} & 17.5 & \textcolor{MyGreen}{+1.1} \\
    (5)& \cmark & 25 pts at random & 56.5 & \textcolor{MyGreen}{+2.6} & 17.7 & \textcolor{MyGreen}{+1.3} \\
    (6)& \cmark & 50 pts at random & 56.6 & \textcolor{MyGreen}{+2.7} & 17.9 & \textcolor{MyGreen}{+1.5} \\
    (7)& \cmark & 100 pts at random & 56.7 & \textcolor{MyGreen}{+2.8} & 17.9 & \textcolor{MyGreen}{+1.5} \\
    \rowcolor{\colormain}
    (8)& \cmark & dense ($\delta\,{=}$\,voxel size) %
    & 56.9 & \textcolor{MyGreen}{+3.0} & 17.5 & \textcolor{MyGreen}{+1.1} \\
    (9)& \cmark & dense with dilation & 56.3 & \textcolor{MyGreen}{+2.4} & 17.4 & \textcolor{MyGreen}{+1.0} \\
    \rowcolor{\colorspecial}
    (10)& \cmark & oracle (dense denoised) & 60.6 & \textcolor{MyGreen}{+6.7} & 20.0 & \textcolor{MyGreen}{+3.6} \\
    \bottomrule
    \end{tabular}
    }

\end{table}

%% file: sections/004_experiments_NEW.tex
\section{Prior combination}
\label{sec:experiments}

We now evaluate %
whether the complementarity hypothesized %
in \autoref{sec:method} holds and applies beyond the two tested SSC methods and the single dataset used in experiments. %
Concretely, we evaluate our combination of priors %
on four architectures (still the lightweight LMSCNet-SS~\cite{lmscnet} and SemCity-AE~\cite{semcity}, but also the stronger SSA-SC~\cite{ssasc} and JS3C-Net~\cite{js3cnet}) and two datasets (SemanticKITTI~\cite{semantickitti} and SSCBench-nuScenes~\cite{sscbench}, which features a sparser lidar sensor). %

\subsection{Merging priors in SSC networks}
\label{subsec:setup_4_arch}

We combine the semantic and visibility prior information into a single input tensor $\mathbf{V}_{\text{w/sem+vis}}$
that carries, on every voxel, both the semantic class (applying to occupied voxels) and the empty/unknown visibility partition (applying to voxels without lidar points).
This yields the following full state set, of size $C+2$:
\[\mathcal{S}_{\text{w/sem+vis}}=\{1,\dots,C, \text{empty}, \text{unknown}\}.\]

Our recipe to integrate the priors %
touches only the input pathway that each backbone already exposes. It adds no auxiliary network and introduces no new jointly trained branch. It merely routes %
the priors through %
entry points that the architecture already provides or, where present, a pre-existing jointly trained branch. The entry point is nonetheless backbone-specific; we detail them below for reproducibility.

\medskip

\subparagraph{SemCity-AE}
Originally designed to process dense, fully annotated point clouds, we adapt SemCity's autoencoder \cite{semcity} for SSC by supplying the sparse augmented tensor $\mathbf{V}_{\text{w/sem+vis}}$ and adjusting its input embedding layer to fit. 
Its prior-free baseline counterpart instead encodes a binary per-voxel occupancy state, capturing only whether at least one lidar point is present in the voxel.

\subparagraph{LMSCNet-SS}
The LMSCNet method \cite{lmscnet} projects 3D spatial volumes into 2D feature maps by collapsing the height axis, converting an $X \times Y \times Z$ grid into an $X \times Y$ map with $Z = 32$ channels. 
We preserve this layout but expand each height bin to carry the complete one-hot vector, resulting in $X \times Y$ maps with $(C + 2) \cdot Z$ channels. 
Only the initial convolutional layer is widened to accept this expanded feature depth.

\subparagraph{SSA-SC}
In SSA-SC \cite{ssasc}, the prior information is integrated exclusively into the BEV completion U-Net. 
Replacing binary occupancy with our $(C + 2)$-dimensional one-hot encoding (covering semantics, empty space, and unknown states) expands the U-Net input width from $32 + 32$ channels (pillar occupancy plus point features) to $(C + 2) \cdot 32 + 32$.
The PointNet module and 3D segmentation branches remain identical to the baseline network.
In our experiments we use the official code, where the 3D semantic segmentation branch gets supervision only from the 2D completion branch.

\subparagraph{JS3C-Net}
Finally, JS3C-Net \cite{js3cnet} incorporates the two priors at distinct entry points. 
The visibility prior enters the completion module as an additional channel flagging observably %
empty voxels, mirroring the setup of other backbones. 
Conversely, the semantic prior lacks a direct pathway to the completion input; it is instead injected into the jointly trained per-point segmentation branch by concatenating one-hot pseudo-labels to the point-wise feature vectors. 
We discuss below the impact of this indirect routing scheme. %

\subsection{Experimental setup}
\label{subsec:exp_setup_section_exp}

\subparagraph{Datasets}

In addition to SemanticKITTI, already presented in \autoref{subsec:setup}, we also experiment here on SSCBench-nuScenes~\cite{sscbench}. This dataset features a sparser lidar sensor, which has 32 beams, against 64 in SemanticKITTI. A single scan is thus considerably sparser vertically.

We adopt the 12-class scheme of SSCBench (\emph{truck} and \emph{other-vehicle} kept separate, \emph{bus} folded into \emph{other-vehicle}), matching Table~II of the SSCBench paper~\cite{sscbench}. 
This taxonomy applies on both sides of the network: it is both the supervised output space and the space the input semantic prior lives in. WaffleIron's 16-class pseudo-labels are therefore remapped into these $C=12$ classes before voxelization, keeping input prior semantics and supervision targets on a single class set. In contrast, SemanticKITTI uses $C=19$ classes.

SSCBench-nuScenes uses 500 training sequences and 199 validation sequences\footnote{The SSCBench paper reports 200 validation scenes, but the publicly released \texttt{trainval} split contains only 699 sequences; the validation portion (sequences 500--698) therefore comprises 199. 
The missing scene issue was raised to the authors but never resolved: \url{https://github.com/ai4ce/SSCBench/issues/3}.
The authors' released checkpoint is evaluated on the same 199-sequence validation split.}
on the same voxel grid as SemanticKITTI, i.e., $256 \times 256 \times 32$ at $20$~cm resolution.

Our retrained LMSCNet-SS on SSCBench-nuScenes is substantially stronger than the LMSCNet baseline reported by the SSCBench paper~\cite{sscbench} (37.6 vs. 21.1 IoU, see \autoref{tab:nuscenes_sota}). 
Since the authors' released checkpoint follows a different class scheme than their reported numbers\footnote{The released checkpoint is LMSCNet-SS trained under the 10-class scheme, whereas Table II of \cite{sscbench} reports LMSCNet under 12 classes; the reported 12-class results are therefore not recoverable from the released weights.
}, we retrain the %
baselines under the paper's 12-class scheme, isolating the effect of the priors.

\subparagraph{Priors}
Unless stated otherwise, we use WaffleIron with test-time augmentations (WI-TTA)~\cite{waffleiron}, the most accurate segmenter in
\autoref{tab:lidar_pseudo_labels}, and plain WaffleIron (WI) without TTA.
We keep this single segmenter family so that \autoref{tab:semkitti_impact} compares one semantic prior with and without TTA; \autoref{tab:lidar_pseudo_labels} already shows that the semantic gain holds for all segmenters, and MinkUNet is used again in \autoref{tab:segmenter_agnostic_deployment}.

As for the geometry, we use the best visibility prior reported in \autoref{subsec:vis}, i.e., the dense uniform sampling of free-space markers along lidar lines of sight, with spacing $\delta = \text{voxel size} = 20\,\text{cm}$. %

We also report results using oracle semantics and/or oracle visibility.

\subparagraph{Refinement}

We did not use the generic SSC refinement proposed in SemCity~\cite{semcity} as the process %
leaks ground-truth occupancy\footnote{
In the official SemCity code, the locations of non-annotated unobserved voxels in the ground truth
are used during refinement, leaking the surface locations that should be reconstructed.}.

\subsection{Quantitative results}
\label{subsec:quantitative}

\subsubsection{Results on SemanticKITTI}
\label{subsubsec:quantitative_SK}

\input{tables/semkitti_impact}

\autoref{tab:semkitti_impact} reports results of prior augmentations on SemanticKITTI. %
Endowing the models with both priors yields consistent gains across all four evaluated architectures.

\paragraph{Semantics}

The semantic gain is uneven, concentrated where semantics is the bottleneck: the lightweight models that predict semantics worst on their own gain the most (+9.5 mIoU pts for SemCity-AE, +7.6 for LMSCNet-SS), while the architectures with a strong internal semantic pathway gain less (+5.6 mIoU pts for SSA-SC, +0.9 for JS3C-Net).

JS3C-Net is the clearest case. 
Like SSA-SC, it trains a point cloud segmentation branch jointly with completion, but our semantic pseudo-labels reach its completion head only through that branch, whereas SSA-SC receives them directly at the completion input. 
The branch is trained to predict the same per-point semantics as the pseudo-labels, so pseudo-labels of similar quality add little ($+0.9$ mIoU pts for both priors combined). 
The routing itself is not the limit: ground-truth labels, injected in the same way, still add $+8.1$ mIoU pts (dense $|$ oracle vs.\ dense $|$ WI-TTA in \autoref{tab:semkitti_impact}).

This is consistent with our observation above: the priors help most where the backbone is weakest, and a model that has already paid for a jointly trained segmenter has little headroom left for an input prior to recover, at the price of being locked to that segmenter. 
The per-class breakdown (\autoref{tab:perclass_semkitti}) confirms that the gains concentrate on the classes that the baselines predict worst unaided. 
For instance, \emph{truck} jumps from 3.7\% IoU to 44.4\% with LMSCNet-SS, and \emph{bicycle} goes from 0.3\% IoU to 12.7\% with SemCity-AE.

\paragraph{Geometric completion}

This remark applies to a lesser extent to geometric completion, where the gains are smaller but substantial for methods that are weaker on completion (+3.1\% IoU for SemCity-AE and +2.3\% for LMSCNet-SS) while they are only incremental for methods that are better at reconstructing geometry (+0.9\% IoU for SSA-SC and +0.4\% for JS3C-Net).

\paragraph{Prior complementarity}

It can also be observed that the priors are mostly complementary. 
Indeed, adding semantic cues to a model equipped with visibility information barely improves geometry (at most +0.4 IoU pt with LMSCNet-SS). 
Conversely, providing a visibility prior on top of a model already augmented with semantic cues has little effect on semantics (+2.2 mIoU pts with LMSCNet-SS, but %
practically no effect with SemCity-AE).

\paragraph*{Oracle headroom} %
The oracle rows of \autoref{tab:semkitti_impact} with dense visibility prior and GT semantic labels confirm the substantial unrealized semantic headroom in the evaluated networks, with gains from +6.2 mIoU pts (LMSCNet-SS) to +11.0 mIoU pts (SSA-SC). The substantial improvement on completion when using not only the GT semantics but also the visibility oracle, from +4.1 IoU pts (SemCity-AE) to +6.3 IoU pts (JS3C-Net), further boosts the semantics, from +2.9 mIoU pts (SemCity-AE) to +8.4 mIoU pts. It shows that resolving free space continues to help the semantic metric even when the semantic prior is already perfect.
This mirrors the \autoref{subsec:vis} finding that visibility lifts mIoU as well as IoU, and confirms the two priors remain complementary at their joint upper bound rather than saturating against each other.

\input{tables/semkitti_sota}

\paragraph*{Comparison to reproducible SSC baselines}
With SSA-SC, the combined prior recipe reaches 28.9\% mIoU, the best semantic score among fully-reproducible single-frame methods, ahead of the strongest prior method in this setting, DPS2CNet~\cite{liu5333789dual} (26.7\%). 
Other scores reported in the literature come from methods that are not directly comparable: DiffSSC~\cite{diffssc} (26.7\% mIoU) reports a panoramic ($360^{\circ}$) validation protocol and its public code has no semantic component, S3CNet~\cite{s3cnet} (33.1\% mIoU) is closed-source, SCPNet~\cite{scpnet} (37.2\% mIoU) is not reproducible from its release (cf.\ \cite[Sec.\,5.2.1]{liu5333789dual} and \cite[Supp., \S 6]{pasco}), and TALoS~\cite{talos} (39.3\% mIoU) relies on multi-frame test-time optimization. 
Within this reproducible single-frame setting, our input-only recipe leads on semantics while remaining plug-and-play.

As for geometric completion, DPS2CNet performs best (60.8\% IoU), with our prior-augmented version of SSA-SC ranking second (58.9\%). The oracle headroom however reveals that the older methods that we evaluated with prior augmentation have the potential to outperform it (\autoref{tab:semkitti_impact}), provided more accurate priors are fed to the networks.

\paragraph*{Inference time}

SSC models have a wide range of run times, from a bit more than 30\,ms/frame on a Nvidia A100 GPU (SSA-SC) to almost 700\,ms/frame (JS3C-Net).
The use of prior augmentations on top of these SSC models incurs an extra latency (details in \autoref{app:profiling}).

It turns out that most of the run time of SSC when adding semantic and visibility priors to a model is spent in the semantic segmenter to infer the pseudo-labels of the semantic cues (one to several hundreds of ms/frame), unless a fast but poor-performing segmenter is used. The effect of the increased number of channels at the first layer of the SSC network mostly remains negligible ($<5$\,ms/frame). Last, the sampling of free-space markers ($\sim$\,6\,ms/frame) and the voxelization that follows ($\sim$\,9\,ms/frame) also are negligible, unless using very fast SSC models such as LMSCNet-SS and SSA-SC, that run in less than 50\,ms/frame.

In practice, the visibility prior (for any variant) and the semantic prior can be precomputed once for a given dataset and cached. %

\input{tables/nuscenes_impact}

\subsubsection{Results on SSCBench-nuScenes}
\label{sec:nuscenes}

We evaluated the combined prior recipe on SSCBench-nuScenes with the two lightweight methods LMSCNet-SS and SemCity-AE (\autoref{tab:nuscenes12}).

\subparagraph{Performance of prior augmentation}

The semantic prior alone carries over cleanly: WaffleIron pseudo-labels, without visibility cues, raise mIoU by $+11.1$ pts on LMSCNet-SS and $+11.7$ pts on SemCity-AE; it is a similar %
effect as seen on SemanticKITTI, and it is preserved once visibility is added, confirming it as a sensor-agnostic driver of the mIoU gain.

The dense sampling of free-space markers also helps here: the visibility prior alone adds $+0.7$ IoU pt on LMSCNet-SS and $+1.2$ IoU pt on SemCity-AE, consistently positive but slightly smaller than the $+1.9$ to $3.0$ IoU pts that it yields on the denser SemanticKITTI sensor (\autoref{tab:semkitti_impact}). 
This is the expected consequence of the sparser lidar sensor, as a single scan certifies less free space per frame. 
The free-space markers are also noisier: the visibility oracle, which uses the same markers but removes those that fall on occupied voxels, gains $+11.6$ to $+13.7$ IoU pts here, and the dense prior obtains less than 10\% of this gain, against 24 to 45\% on SemanticKITTI.

The combination of the semantic and visibility priors  provides a little more gain than the best of both individual priors: +0.8 to +0.9 IoU pt for geometric completion when adding semantics, but no extra gain on semantics when adding the visibility prior (difference within 0.1 mIoU pt). As already noted in \autoref{subsec:vis}, it shows and confirms that the two priors are mostly complementary, with a small ripple effect.

\input{tables/nuscenes_sota}
\paragraph*{SSC baseline comparison}

There are few baselines on SSCBench-nuScenes. When equipped with both the semantic and visibility priors, the two methods we evaluated, which were already ranking first and second, get a large boost in semantics (+11.1 to 12.2 mIoU pts) and a modest gain in geometry (+1.5 to 2.0 IoU pts) and keep their ranking.

\input{figures/qualitative}

\subsection{Qualitative results}
\label{subsec:qualitative}

\autoref{fig:qualitative} compares the SSC outputs of the two main networks we studied (SemCity-AE and LMSCNet-SS) against their prior-free baselines on both datasets: SemanticKITTI (\autoref{fig:qualitative}A) and the sparser SSCBench-nuScenes (\autoref{fig:qualitative}B), on two validation scenes each.

Operating only on the raw sparse input point cloud (\emph{Input}), the baselines (\emph{Baseline}) generate wrong semantics and noisy, fragmented class boundaries for SemanticKITTI.
In contrast, when endowed with our input priors (\emph{+\,priors}), both models produce cleaner and more structured outputs, closer to the ground truth (\emph{GT}): continuous surfaces such as roads (purple) become fully connected, vegetation aligns better with the GT, and thin structures such as poles and traffic signs are more reliably restored.

The same improvement holds on the 32-beam SSCBench-nuScenes scans (\autoref{fig:qualitative}B), confirming that the qualitative benefit transfers across sensors.
This is also confirmed quantitatively by the per-class breakdown in \autoref{app:perclass} (\autoref{tab:perclass_semkitti} and \autoref{tab:perclass_nuscenes12}).

\subsection{Segmenter-agnostic training}
\label{sec:agnostic}

A practical concern with an input semantic prior is model coupling: if the SSC network is trained on the pseudo-labels of one segmenter, must it be retrained whenever that segmenter is replaced or upgraded?
We show it need not be.

\input{tables/segmenter-agnostic-deployment}

\autoref{tab:segmenter_agnostic_deployment} quantifies it for the two lightweight architectures we use in our study, based on the full prior recipe (both semantic cues and dense visibility markers). %
For each off-the-shelf segmenter, we compare the matched setting (the completion network trained on that segmenter's pseudo-labels) against a single network trained once on GT semantics and switched to the segmenter labels only at inference.
The reported $\Delta$ is the cost of not retraining; a small value indicates that a single GT-trained network can be used as an alternative to any network trained with a specific segmenter. %

As can be seen from the table, the cost (or benefit) of not retraining for a particular segmenter is negligible: between $-$0.5 and $+$0.2 IoU pt for occupancy, and between $-$0.2 and $+$1.0 mIoU pt for semantics.
Prior sequential pipelines (\autoref{subsec:related-priors}) process semantic predictions from separate networks, but do not establish how robust the completion head is to which segmenter supplies them.
Specifically, the ability to train the completion head just once with ground-truth semantics and subsequently swap in arbitrary off-the-shelf segmenters at inference without retraining is a distinct advantage we systematically exploit here, contrasting with the rigid coupling of multi-task methods like SSA-SC~\cite{ssasc} and JS3C-Net~\cite{js3cnet}.

We hypothesize that part of the explanation of this result is related to overfitting.
Lidar segmenters tend indeed to overfit their training split. The pseudo-labels they produce on the SSC training sequences thus sit close to the ground-truth annotations.
A completion network trained on GT input semantics therefore sees a training-time input distribution that closely matches what any reasonable off-the-shelf segmenter produces at inference.
The completion network can thus be trained once, on GT semantics, and paired at inference with an arbitrary segmenter (including a future, stronger one) at no retraining cost.

This decoupling is specific to the semantic prior: performance drops when training on the visibility oracle and testing on the dense sampling of free-space markers (\autoref{app:vis_mismatch}).

%% file: tables/semkitti_impact.tex
\begin{table}[t]

    \caption{\textbf{SSC prior combination on SemanticKITTI} (validation set).
    We report geometric completion (IoU) and semantics (mIoU). $\Delta$ measures the difference with the model that does not use the corresponding prior.
    ``-'': no prior provided. $^\dagger$baseline is retrained.
    }
    \label{tab:semkitti_impact}

    \setlength{\tabcolsep}{5pt}
    \centering
    {%
    \begin{tabular}{c|c|cc|cc}
\toprule
    \multicolumn{2}{c|}{1-frame input} &  \multicolumn{4}{c}{SSC output~~} \\[1pt]

     visibility & semantics &  \multicolumn{2}{c|}{geometry} & \multicolumn{2}{c}{semantics}\\

     prior & prior & IoU\,\% & $\Delta$ & mIoU\rlap{\,\%} & $\Delta$ \\
\midrule
    \rowcolor{\colortitle}
    \multicolumn{6}{l}{\textbf{SemCity-AE}~\cite{semcity}$^\dagger$ }\\

    - & -  & 53.9  & -    & 16.4 & - \\

    dense & - & 56.9 & \textcolor{MyGreen}{+3.0} & 17.5 & \textcolor{MyGreen}{+1.1} \\

\cline{1-2}
&&&&&\\[-1.9ex]

    - & WI & 54.4 & \textcolor{MyGreen}{+0.5} & 25.5 & \textcolor{MyGreen}{+9.1} \\

    dense & WI & \textbf{57.0} & \textcolor{MyGreen}{+3.1} & 25.5 & \textcolor{MyGreen}{+9.1} \\

\cline{1-2}
&&&&&\\[-1.9ex]

    - & WI-TTA & 54.4 & \textcolor{MyGreen}{+0.5} & \textbf{26.0} & \textcolor{MyGreen}{+9.6} \\

    \rowcolor{\colormain}
    dense & WI-TTA & \textbf{57.0} & \textcolor{MyGreen}{+3.1} & 25.9 & \textcolor{MyGreen}{+9.5} \\

\cline{1-2}
&&&&&\\[-1.9ex]

    \rowcolor{\colorspecial}
    dense & oracle  & 57.1 & \textcolor{MyGreen}{+3.2} & 34.4 & \textcolor{MyGreen}{+18.0} \\

    \rowcolor{\colorspecial}
    oracle &  oracle   & 61.2 & \textcolor{MyGreen}{+7.3} & 37.3 & \textcolor{MyGreen}{+20.9} \\

\midrule
    \rowcolor{\colortitle}
    \multicolumn{6}{l}{\textbf{LMSCNet-SS}~\cite{lmscnet}$^\dagger$ }\\
    
    - & -     & 55.6 & -    & 16.5 & - \\

    dense  & -       & 57.5 & \textcolor{MyGreen}{+1.9} &  18.1 & \textcolor{MyGreen}{+1.6} \\

\cline{1-2}
&&&&&\\[-1.9ex]

    -  & WI       & 55.6 & \textcolor{MyGreen}{+0.0} &  21.6 & \textcolor{MyGreen}{+5.1} \\

    dense  & WI       & \textbf{57.9} & \textcolor{MyGreen}{+2.3} &  23.8 & \textcolor{MyGreen}{+7.3} \\

\cline{1-2}
&&&&&\\[-1.9ex]

    -  & WI-TTA       & 55.6 & \textcolor{MyGreen}{+0.0} &  21.9 & \textcolor{MyGreen}{+5.4} \\

    \rowcolor{\colormain}
    dense  & WI-TTA       & \textbf{57.9} & \textcolor{MyGreen}{+2.3} &  \textbf{24.1} & \textcolor{MyGreen}{+7.6} \\

\cline{1-2}
&&&&&\\[-1.9ex]

    \rowcolor{\colorspecial}
    dense & oracle    & 58.2 & \textcolor{MyGreen}{+2.6} &  30.3 & \textcolor{MyGreen}{+13.8} \\

    \rowcolor{\colorspecial}
    oracle & oracle    & 63.2 & \textcolor{MyGreen}{+7.6} &  34.8 & \textcolor{MyGreen}{+18.3} \\

\midrule
    \rowcolor{\colortitle}
    \multicolumn{6}{l}{\textbf{JS3C-Net}~\cite{js3cnet}}\\

    -  & -   & 57.0 & -    & 24.0 & - \\
    
    dense & WI          & \textbf{57.5} & \textcolor{MyGreen}{+0.5} & 24.2 & \textcolor{MyGreen}{+0.2} \\
    
    \rowcolor{\colormain}
    dense & WI-TTA          & 57.4 & \textcolor{MyGreen}{+0.4} & \textbf{24.9} & \textcolor{MyGreen}{+0.9} \\

\cline{1-2}
&&&&&\\[-1.9ex]
    
    \rowcolor{\colorspecial}
    dense & oracle            & 57.7 & \textcolor{MyGreen}{+0.7} & 33.0 & \textcolor{MyGreen}{+9.0} \\
    
    \rowcolor{\colorspecial}
    oracle & oracle    & 64.0 & \textcolor{MyGreen}{+7.0} & 37.8 & \textcolor{MyGreen}{+13.8} \\

\midrule
    \rowcolor{\colortitle}
    \multicolumn{6}{l}{\textbf{SSA-SC}~\cite{ssasc}$^\dagger$}\\

    - & -       & 58.0 & - & 23.3 &- \tikzmark{ssaA} \\

    dense & WI         &  \textbf{58.9} & \textcolor{MyGreen}{+0.9} & 28.6 & \textcolor{MyGreen}{+5.3} \\

    \rowcolor{\colormain}
    dense  & WI-TTA        &  \textbf{58.9} &\textcolor{MyGreen}{+0.9} & \textbf{28.9} & \textcolor{MyGreen}{+5.6}\tikzmark{ssaB} \\

&&&&&\\[-1.9ex]

    \rowcolor{\colorspecial}
    dense  & oracle              & 60.0 & \textcolor{MyGreen}{+2.0} & 39.9 & \textcolor{MyGreen}{+16.6}\tikzmark{ssaC} \\

    \rowcolor{\colorspecial}
    oracle & oracle   & 65.9 & \textcolor{MyGreen}{+7.9} & 48.3 & \textcolor{MyGreen}{+25.0}\tikzmark{ssaD} \\

\bottomrule
    \end{tabular}
    }
\end{table}

%% file: tables/semkitti_sota.tex
\begin{table}[t]
    \caption{\textbf{SSC comparison with baselines on SemanticKITTI} (validation set, 19 classes). We use the best priors, i.e., WI-TTA semantic pseudo-labels and dense uniform sampling of free-space markers.
    $^\dagger$retrained; other baselines taken from the respective papers.
    $^a$as reported in the paper.
    $^b$public code has no semantic part; values from the paper use the panoramic ($360^{\circ}$) protocol on the validation set.
    $^c$not reproducible, cf.\ \cite[Sec.\,5.2.1]{liu5333789dual} and \cite[Supp., \S 6]{pasco}.
    $^d$multi-frame test-time optimization.
    }
    \label{tab:semkitti_sota}

    \setlength{\tabcolsep}{5pt}
    \centering
    {%
    \begin{tabular}{@{~~}c|l@{}r|c|c@{~~}}
    \toprule
    &&& geom. & sem.\\
    \smash{\rotatebox[origin=bl]{90}{\!\raisebox{3.5pt}{code}}} & Model && IoU\,\% & mIoU\,\% \\
    \midrule

    \cmark &
    SemCity-AE~\cite{semcity}$^\dagger$ && 53.9  & 16.4\tikzmark{semcityaeS}\\
    
    \cmark &
    LMSCNet-SS~\cite{lmscnet}$^\dagger$ && 55.6 & 16.5\tikzmark{lmscnetssS}\\
    
    \cmark &
    LODE~\cite{lode} && 51.2 & 20.2 \\
    
    \cmark &
    SSA-SC~\cite{ssasc}$^\dagger$ && 58.0 & 23.3\tikzmark{ssascS} \\
    
    \cmark &
    JS3C-Net~\cite{js3cnet} && 57.0 & 24.0\tikzmark{js3cnetS} \\
    
    \rowcolor{\colormain}
    \cmark &
    LMSCNet-SS~\cite{lmscnet} & +priors & 57.9 & 24.1\tikzmark{lmscnetssT}\\
    
    \cmark &
    SSA-SC~\cite{ssasc}$^a$ && 58.3 & 24.5 \\

    \cmark &
    SSC-RS~\cite{sscrs} && 58.6 & 24.8 \\

    \rowcolor{\colormain}
    \cmark &
    JS3C-Net~\cite{js3cnet} & +priors & 57.4 & 24.9\tikzmark{js3cnetT} \\

    \rowcolor{\colormain}
    \cmark &
    SemCity-AE~\cite{semcity} &+priors & 57.0 & 25.9\tikzmark{semcityaeT} \\
    
    \cmark &
    DPS2CNet~\cite{liu5333789dual} && \bf 60.8 & 26.7 \\

    \rowcolor{\colormain}
    \cmark &
    SSA-SC~\cite{ssasc} &+priors & 58.9 & \bf 28.9\tikzmark{ssascT} \\

    \midrule

    \rowcolor{\colorspecial}
    \multicolumn{5}{l}{\emph{Other methods from the literature}}\\

    \rowcolor{\colorspecial}

    \rowcolor{\colorspecial}
        \nmark &
    DiffSSC~\cite{diffssc}$^b$ && 60.3 & 26.7 \\
        \rowcolor{\colorspecial}

        \xmark &
    S3CNet~\cite{s3cnet} && 57.2 & 33.1 \\
        \rowcolor{\colorspecial}

    \nmark &
    SCPNet~\cite{scpnet}$^c$ && 49.9 & 37.2 \\
    
    \rowcolor{\colorspecial}
    \cmark &
    TALoS~\cite{talos}$^d$ && 56.1 & 39.3 \\
    \bottomrule
    \end{tabular}
    \begin{tikzpicture}[overlay, remember picture, shorten >=1pt, shorten <=1pt]

    \draw [->, black] ([xshift=3pt,yshift=0.7ex]pic cs:semcityaeS) to [bend left=50]([xshift=3pt,yshift=0.7ex]pic cs:semcityaeT);
    \draw [->, black] ([xshift=3pt,yshift=0.7ex]pic cs:lmscnetssS) to [bend left=50]([xshift=3pt,yshift=0.7ex]pic cs:lmscnetssT);
    \draw [->, black] ([xshift=3pt,yshift=0.7ex]pic cs:js3cnetS) to [bend left=50]([xshift=3pt,yshift=0.7ex]pic cs:js3cnetT);
    \draw [->, black] ([xshift=3pt,yshift=0.7ex]pic cs:ssascS) to [bend left=50]([xshift=3pt,yshift=0.7ex]pic cs:ssascT);
\end{tikzpicture}
    }

\end{table}

%% file: tables/nuscenes_impact.tex
\begin{table}[t]

    \caption{\textbf{Prior combination on SSCBench-nuScenes} (validation set, 12-class scheme).
    We report geometric completion (IoU) and semantics (mIoU). $\Delta$ measures the difference with the model that does not use the corresponding prior.
    $^\dagger$baseline is retrained.
    }
    \label{tab:nuscenes12}

    \setlength{\tabcolsep}{5pt}
    \centering
    {%
    \begin{tabular}{c|c|cc|cc}
\toprule
    \multicolumn{2}{c|}{1-frame input} &  \multicolumn{4}{c}{SSC output~~} \\[1pt]

     visibility & semantics &  \multicolumn{2}{c|}{geometry} & \multicolumn{2}{c}{semantics}\\

    prior & prior & IoU\,\% & $\Delta$ & mIoU\rlap{\,\%} & $\Delta$ \\
\midrule
    \rowcolor{\colortitle}
    \multicolumn{6}{l}{\textbf{SemCity-AE}~\cite{semcity}$^\dagger$ $\vphantom{X^{X^X}}$}\\

    - & -  & 37.6  & -    & 15.6 & - \\

    dense & - & 38.8 & \textcolor{MyGreen}{+1.2} & 16.1 & \textcolor{MyGreen}{+0.5} \\

\cline{1-2}
&&&&&\\[-1.9ex]

    - & WI & 38.6 & \textcolor{MyGreen}{+1.0} & 27.3 & \textcolor{MyGreen}{+11.7} \\

    dense & WI & \textbf{39.6} & \textcolor{MyGreen}{+2.0} & 27.4 & \textcolor{MyGreen}{+11.8} \\

\cline{1-2}
&&&&&\\[-1.9ex]

    - & WI-TTA & 38.6 & \textcolor{MyGreen}{+1.0} & \textbf{27.8} & \textcolor{MyGreen}{+12.2} \\

    \rowcolor{\colormain}
    dense & WI-TTA & \textbf{39.6} & \textcolor{MyGreen}{+2.0} & \textbf{27.8} & \textcolor{MyGreen}{+12.2} \\

\cline{1-2}
&&&&&\\[-1.9ex]

    \rowcolor{\colorspecial}
    oracle & -  & 49.2 & \textcolor{MyGreen}{+11.6} & 19.1 & \textcolor{MyGreen}{+3.5} \\

    \rowcolor{\colorspecial}
    oracle & WI  & 50.0 & \textcolor{MyGreen}{+12.4} & 33.4 & \textcolor{MyGreen}{+17.8} \\

    \rowcolor{\colorspecial}
    oracle & WI-TTA  & 50.0 & \textcolor{MyGreen}{+12.4} & 33.7 & \textcolor{MyGreen}{+18.1} \\

\midrule
    \rowcolor{\colortitle}
    \multicolumn{6}{l}{\textbf{LMSCNet-SS}~\cite{lmscnet}$^\dagger$ $\vphantom{X^{X^X}}$}\\

    - & -     & 37.6 & -    & 15.0 & - \\

    dense  & -       & 38.3 & \textcolor{MyGreen}{+0.7} &  16.6 & \textcolor{MyGreen}{+1.6} \\

\cline{1-2}
&&&&&\\[-1.9ex]

    -  & WI       & 38.6 & \textcolor{MyGreen}{+1.0} &  26.1 & \textcolor{MyGreen}{+11.1} \\

    dense  & WI       & \textbf{39.2} & \textcolor{MyGreen}{+1.6} &  26.0 & \textcolor{MyGreen}{+11.0} \\

\cline{1-2}
&&&&&\\[-1.9ex]

    -  & WI-TTA       & 38.6 & \textcolor{MyGreen}{+1.0} &  \textbf{26.2} & \textcolor{MyGreen}{+11.2} \\

    \rowcolor{\colormain}
    dense  & WI-TTA       & 39.1 & \textcolor{MyGreen}{+1.5} &  26.1 & \textcolor{MyGreen}{+11.1} \\

\cline{1-2}
&&&&&\\[-1.9ex]

    \rowcolor{\colorspecial}
    oracle & -    & 51.3 & \textcolor{MyGreen}{+13.7} &  21.7 & \textcolor{MyGreen}{+6.7} \\

    \rowcolor{\colorspecial}
    oracle & WI    & 51.6 & \textcolor{MyGreen}{+14.0} &  31.7 & \textcolor{MyGreen}{+16.7} \\

    \rowcolor{\colorspecial}
    oracle & WI-TTA    & 51.6 & \textcolor{MyGreen}{+14.0} &  31.8 & \textcolor{MyGreen}{+16.8} \\

\bottomrule
    \end{tabular}
    }
\end{table}

%% file: tables/nuscenes_sota.tex
\begin{table}[t]
    \caption{\textbf{SSC comparison with baselines on SSCBench-nuScenes} (validation set, 12-class scheme). We use the best priors: WI-TTA semantic pseudo-labels and dense uniform sampling of free-space markers.
    $^\dagger$retrained; $^\ddagger$scores reported in~\cite{sscbench}.
    }
    \label{tab:nuscenes_sota}

    \setlength{\tabcolsep}{5pt}
    \centering
    {%
    \begin{tabular}{@{~~}c|l@{~}r|c|c@{~~}}
    \toprule
    &&& geom. & sem.\\
    \smash{\rotatebox[origin=bl]{90}{\!\raisebox{3pt}{code}}} & Model && IoU\,\% & mIoU\,\% \\
    \midrule

    \cmark &
    LMSCNet~\cite{lmscnet}$^\ddagger$ && 21.1 & 8.4 \\

    \cmark &
    SSCNet~\cite{sscnet}$^\ddagger$ && 27.6 & 11.8 \\

    \cmark &
    LMSCNet-SS~\cite{lmscnet}$^\dagger$ && 37.6 & 15.0\tikzmark{nuLmscnetssS}\\

    \cmark &
    SemCity-AE~\cite{semcity}$^\dagger$ && 37.6 & 15.6\tikzmark{nuSemcityaeS}\\

    \rowcolor{\colormain}
    \cmark &
    LMSCNet-SS*~\cite{lmscnet} & +priors & 39.1 & 26.1\tikzmark{nuLmscnetssT}\\

    \rowcolor{\colormain}
    \cmark &
    SemCity-AE*~\cite{semcity} & +priors & \bf 39.6 & \bf 27.8\tikzmark{nuSemcityaeT}\\
    \bottomrule
    \end{tabular}
    \begin{tikzpicture}[overlay, remember picture, shorten >=1pt, shorten <=1pt]
    \draw [->, black] ([xshift=3pt,yshift=0.7ex]pic cs:nuLmscnetssS) to [bend left=40]([xshift=3pt,yshift=0.7ex]pic cs:nuLmscnetssT);
    \draw [->, black] ([xshift=3pt,yshift=0.7ex]pic cs:nuSemcityaeS) to [bend left=40]([xshift=3pt,yshift=0.7ex]pic cs:nuSemcityaeT);
    \end{tikzpicture}
    }

\end{table}

%% file: figures/qualitative.tex
\newcommand{\qualim}[1]{\includegraphics[width=0.18\linewidth, trim={2cm 4cm 2cm 3cm},clip]{#1}}
\newcommand{\qualph}{\includegraphics[width=0.18\linewidth]{example-image-a}}%
\newcommand{\qualsqA}[1]{\includegraphics[width=0.18\linewidth, trim={130.2bp 2.4bp 108.6bp 2.4bp},clip]{#1}}
\newcommand{\qualsqB}[1]{\includegraphics[width=0.18\linewidth, trim={109.8bp 0bp 129.0bp 4.8bp},clip]{#1}}
\newcommand{\qualsqBell}[1]{%
\begin{tikzpicture}%
\node[anchor=south west, inner sep=0pt] (img) at (0,0) {\qualsqB{#1}};%
\begin{scope}[x={(img.south east)}, y={(img.north west)}]%
\draw[red, line width=1.1pt] (0.5,0.4) ellipse (0.44 and 0.2);%
\end{scope}%
\end{tikzpicture}}
\newcommand{\qualAellOne}{\draw[red, line width=1.1pt] (0.7,0.65) ellipse (0.15 and 0.15);}%
\newcommand{\qualAellTwo}{\draw[red, line width=0.71pt] (0.3,0.4) ellipse (0.07 and 0.07);}%
\newcommand{\qualsqAmark}[2]{%
\begin{tikzpicture}%
\node[anchor=south west, inner sep=0pt] (img) at (0,0) {\qualsqA{#2}};%
\begin{scope}[x={(img.south east)}, y={(img.north west)}]%
#1%
\end{scope}%
\end{tikzpicture}}
\newcommand{\qualsqAell}[1]{\qualsqAmark{\qualAellOne}{#1}}
\newcommand{\qualsqC}[1]{\includegraphics[width=0.18\linewidth, trim={184.8bp -95.1bp 200.4bp -50.1bp},clip]{#1}}
\newcommand{\qualCellOne}{\draw[red, line width=1.1pt, rotate around={45:(0.5,0.5)}] (0.65,0.45) ellipse (0.27 and 0.08);}%
\newcommand{\qualsqCmark}[2]{%
\begin{tikzpicture}%
\node[anchor=south west, inner sep=0pt] (img) at (0,0) {\qualsqC{#2}};%
\begin{scope}[x={(img.south east)}, y={(img.north west)}]%
#1%
\end{scope}%
\end{tikzpicture}}
\newcommand{\qualsqD}[1]{\includegraphics[width=0.18\linewidth, trim={273.0bp 4.8bp 262.2bp 0bp},clip]{#1}}
\newcommand{\qualDellOne}{\draw[red, line width=0.71pt, rotate around={10:(0.33,0.5)}] (0.85,0.4) ellipse (0.046 and 0.138);}%
\newcommand{\qualDellTwo}{\draw[red, line width=0.71pt, rotate around={10:(0.67,0.5)}] (0.65,0.5) ellipse (0.046 and 0.138);}%
\newcommand{\qualsqDmark}[2]{%
\begin{tikzpicture}%
\node[anchor=south west, inner sep=0pt] (img) at (0,0) {\qualsqD{#2}};%
\begin{scope}[x={(img.south east)}, y={(img.north west)}]%
#1%
\end{scope}%
\end{tikzpicture}}

\newcommand{\rowlabel}[1]{\raisebox{0.09\linewidth}{\raisebox{-.5\height}{\rotatebox{90}{#1}}}}
\newcommand{\rowlabelcenter}[1]{\raisebox{-.5\height}{\rotatebox{90}{#1}}}
\newcommand{\dashedhrule}{\leavevmode\leaders\hbox to 4mm{\hfil\textcolor{black!40}{\rule{2mm}{0.5pt}}\hfil}\hfill\kern0pt}

\begin{figure*}[!tp]
\centering
\setlength{\tabcolsep}{2pt}
\renewcommand{\arraystretch}{1.0}
\small
\begin{tabular}{@{}cc@{\hspace{5mm}}cccc@{}}
 & & \multicolumn{2}{c}{\textbf{(A) SemanticKITTI} (64-beam)} & \multicolumn{2}{c}{\textbf{(B) SSCBench-nuScenes} (32-beam)} \\
\cmidrule(lr){3-4} \cmidrule(lr){5-6}
\multicolumn{2}{c}{\rowlabel{input}} &
\qualsqB{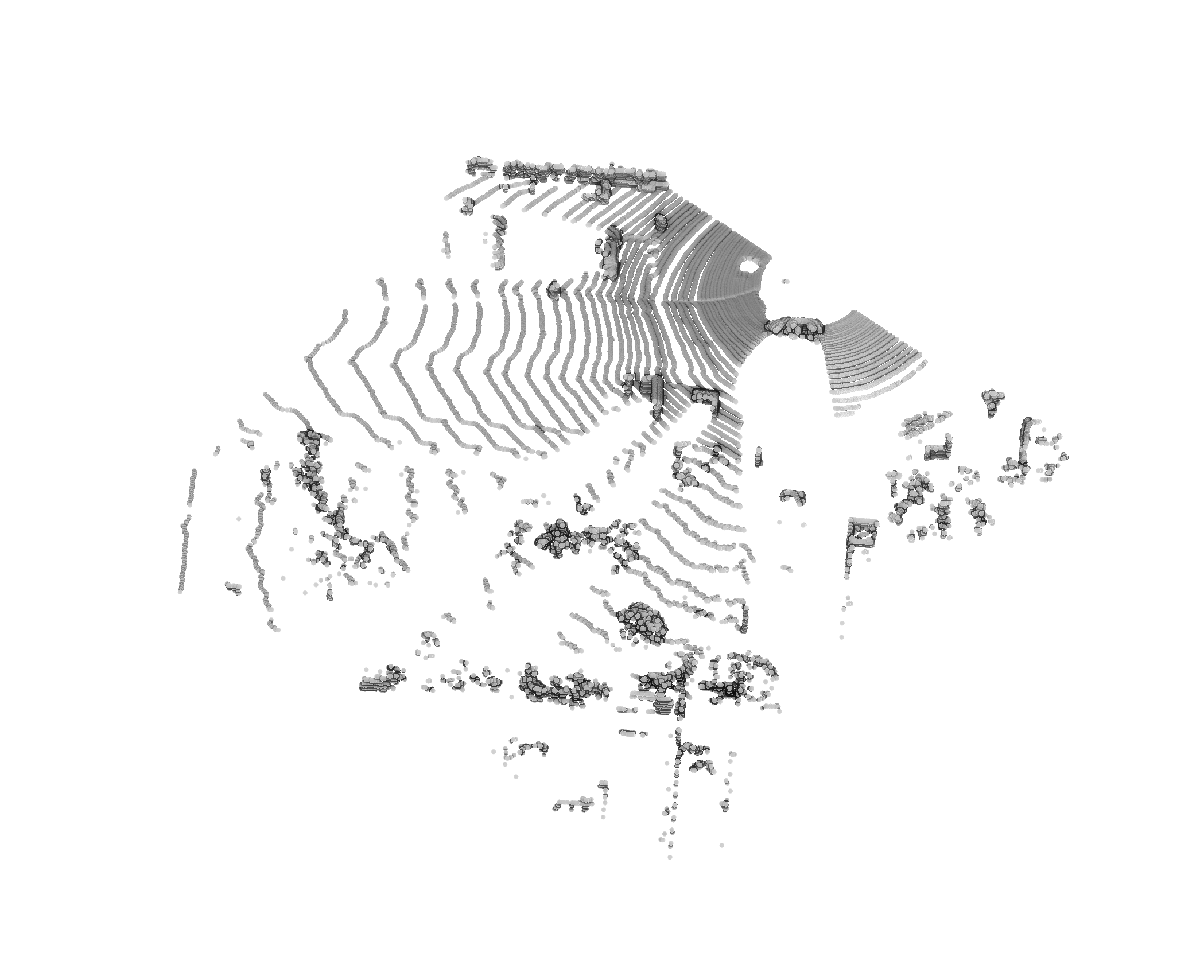} &
\qualsqAmark{\qualAellTwo}{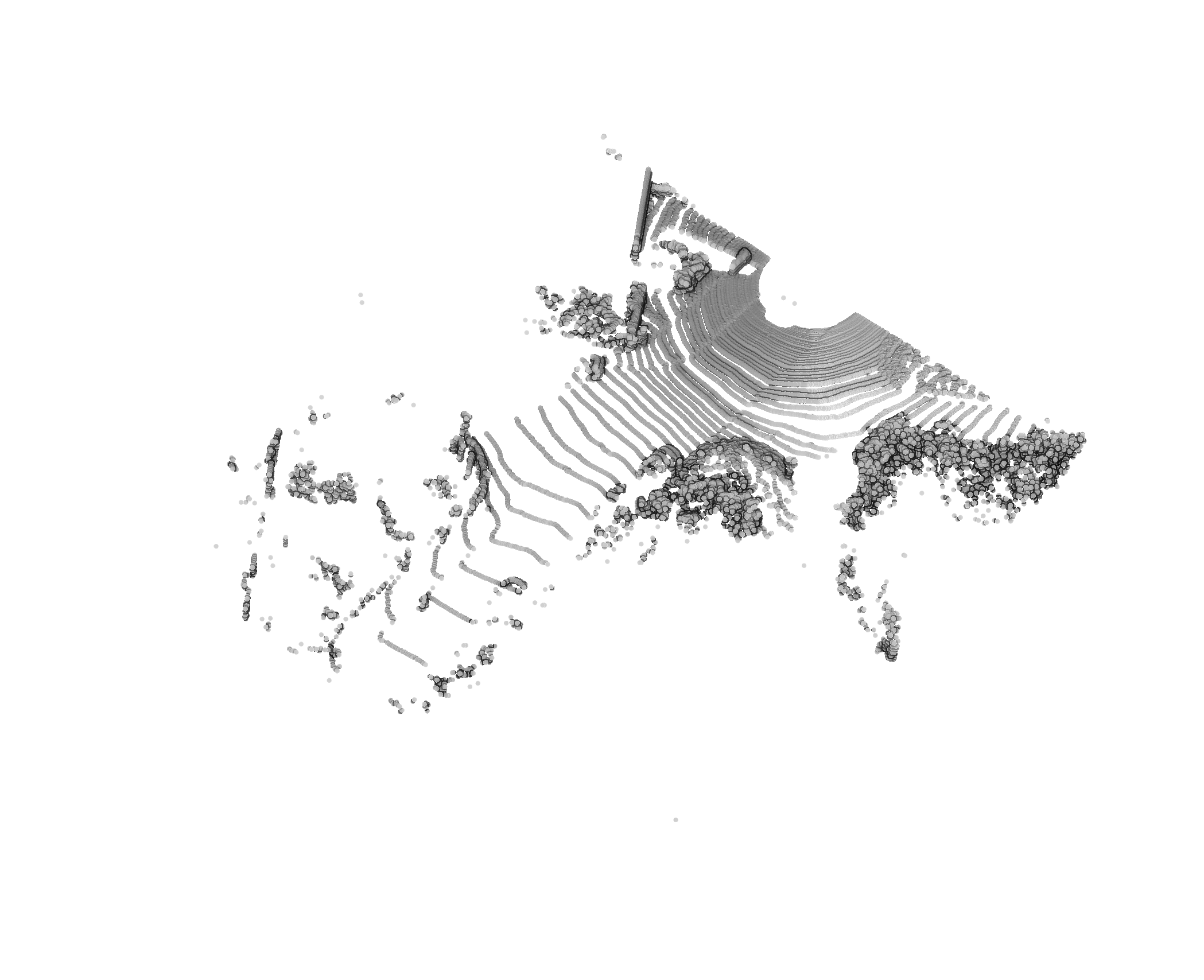} &
\qualsqC{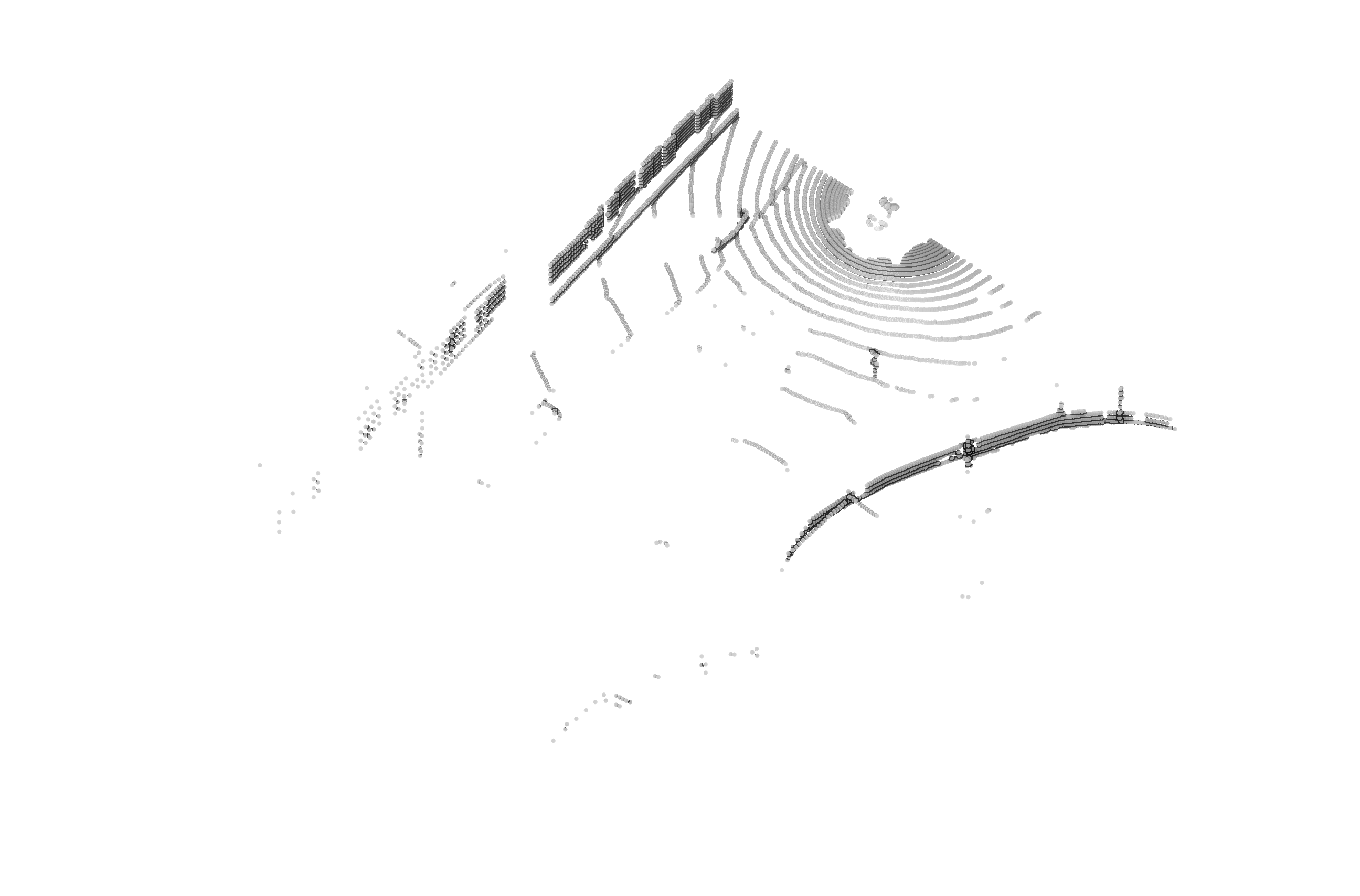} &
\qualsqD{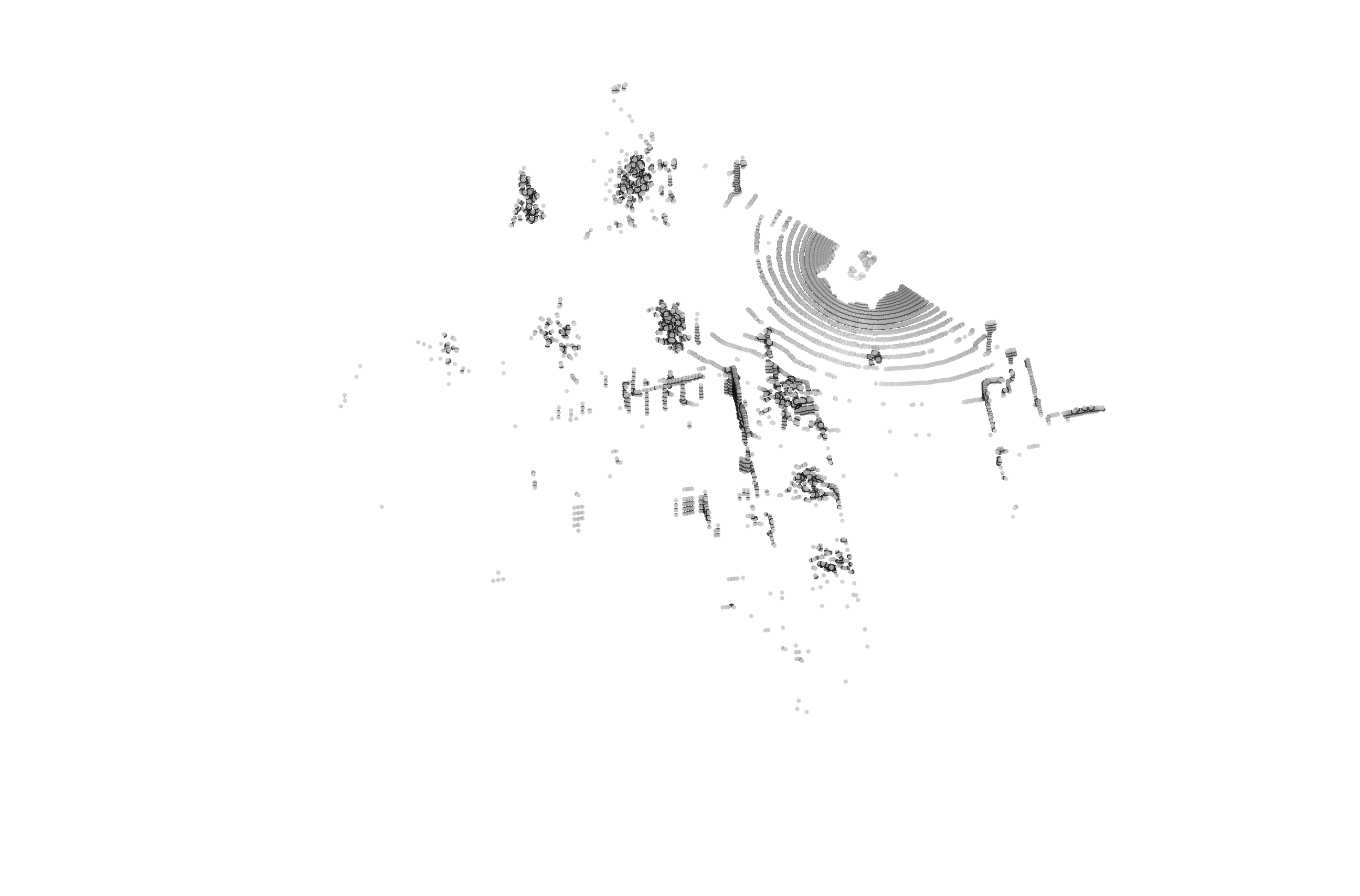} \\
\midrule
\multirow{2}{*}{\rowlabelcenter{SemCity-AE}} & \rowlabel{baseline} &
\qualsqBell{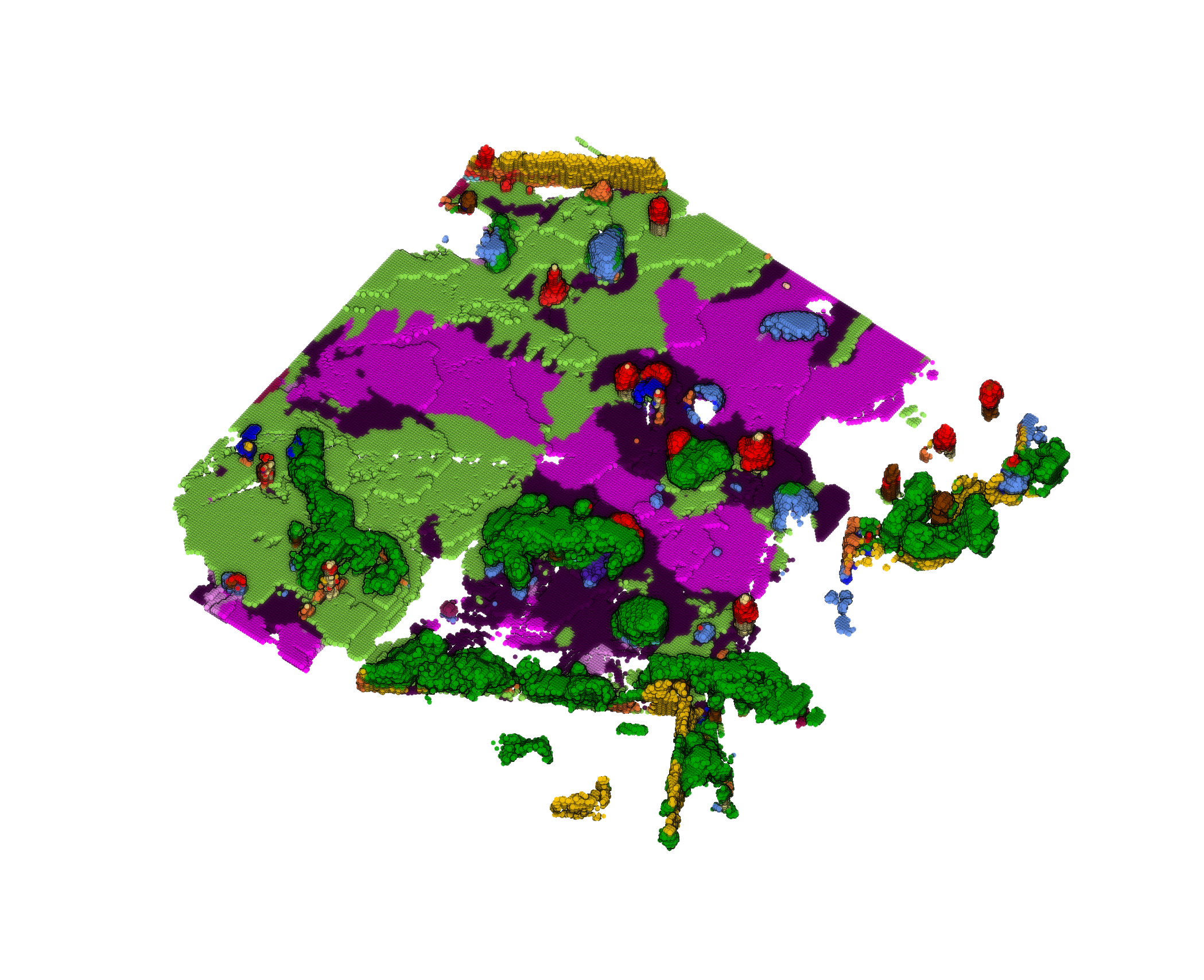} &
\qualsqAmark{\qualAellOne\qualAellTwo}{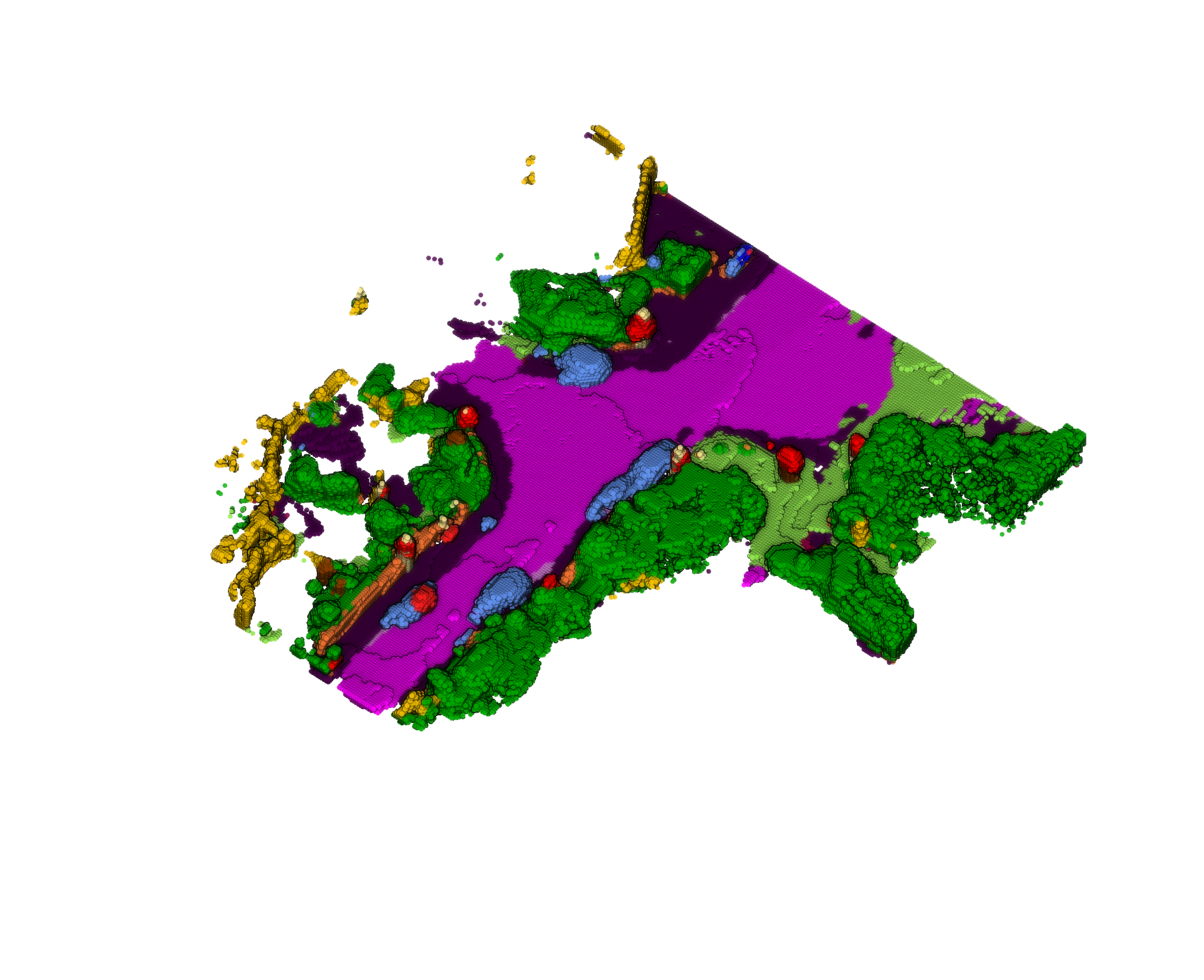} &
\qualsqCmark{\qualCellOne}{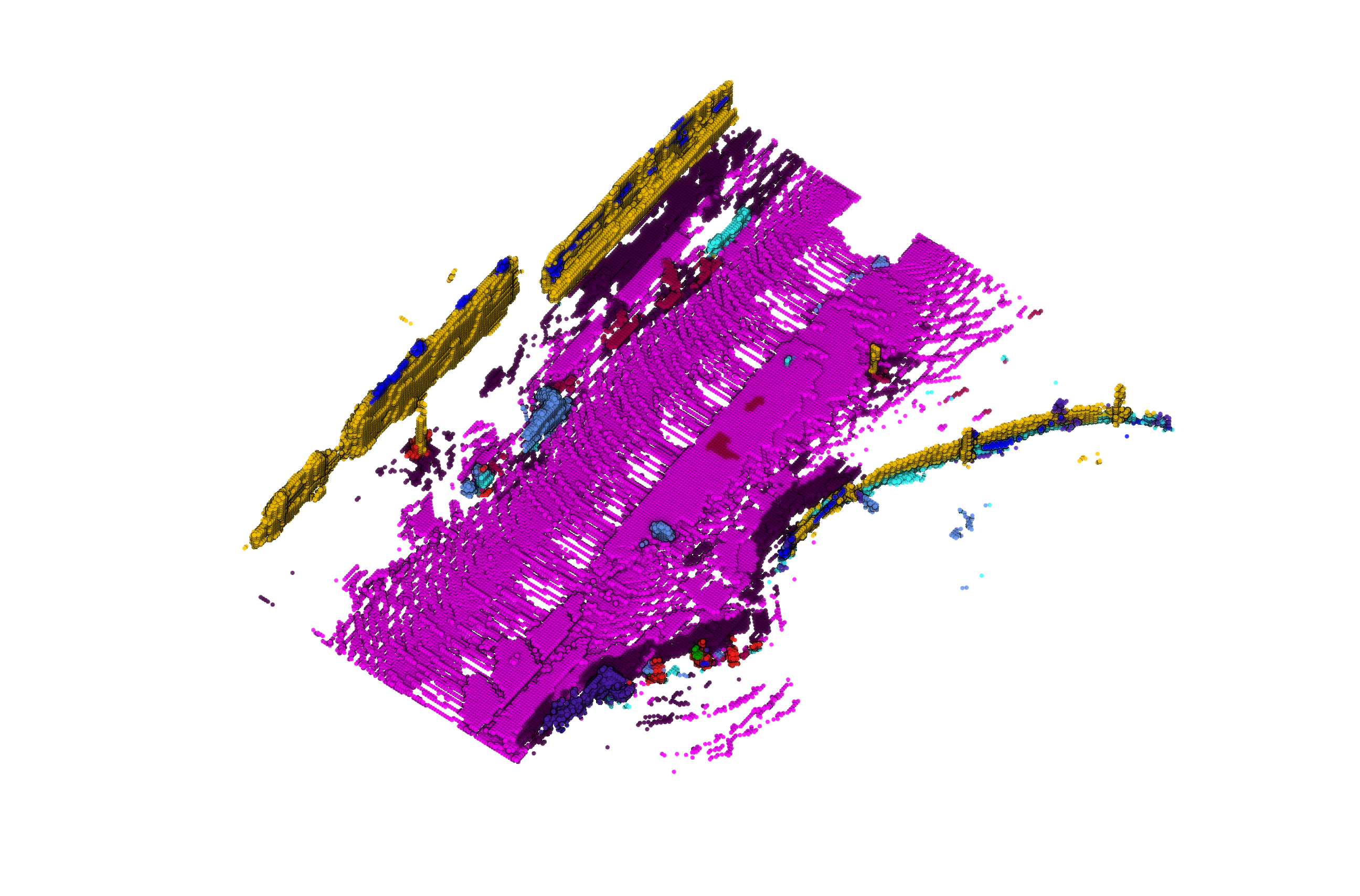} &
\qualsqDmark{\qualDellOne\qualDellTwo}{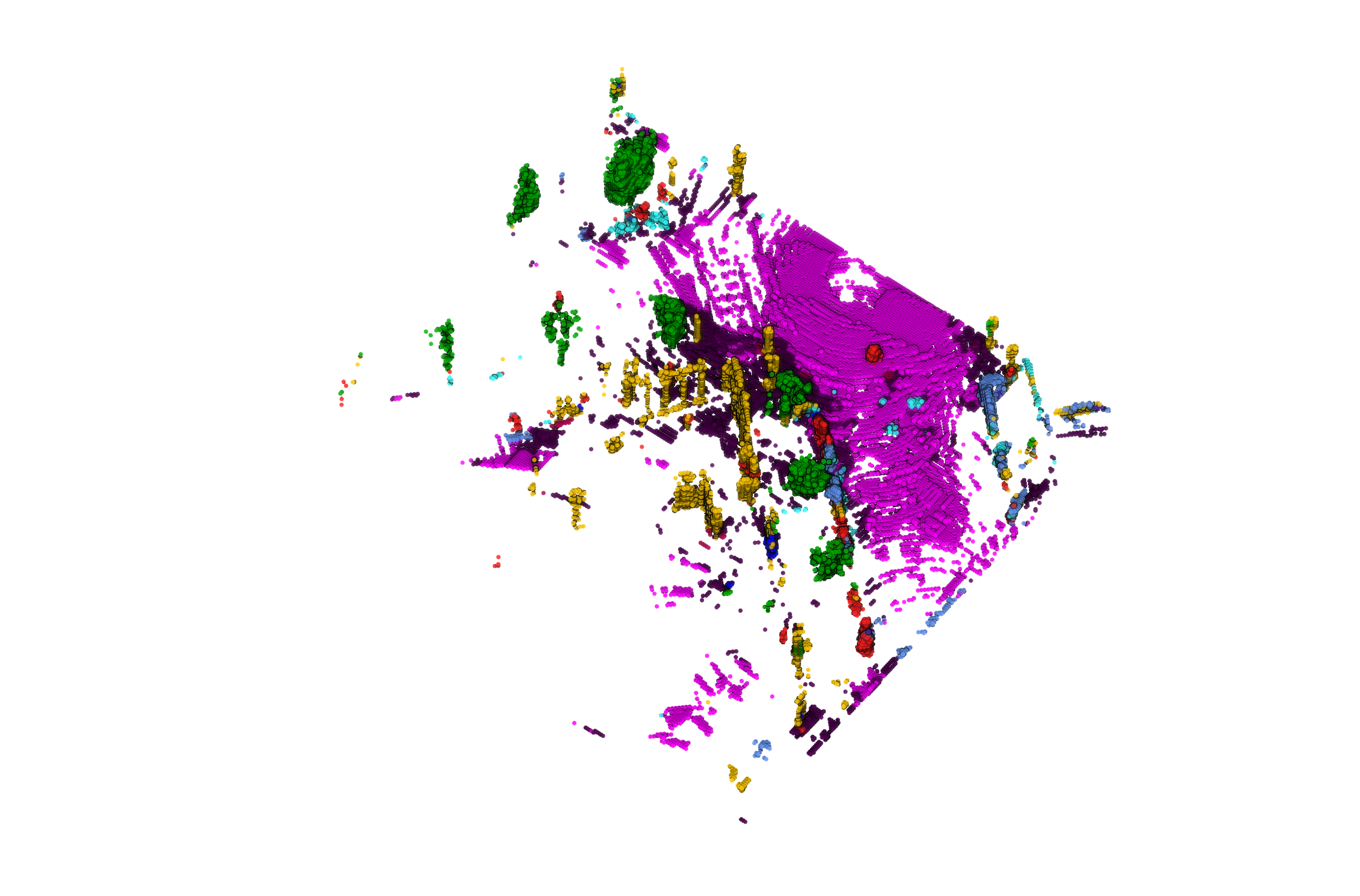} \\
 & \rowlabel{+ our priors} &
\qualsqBell{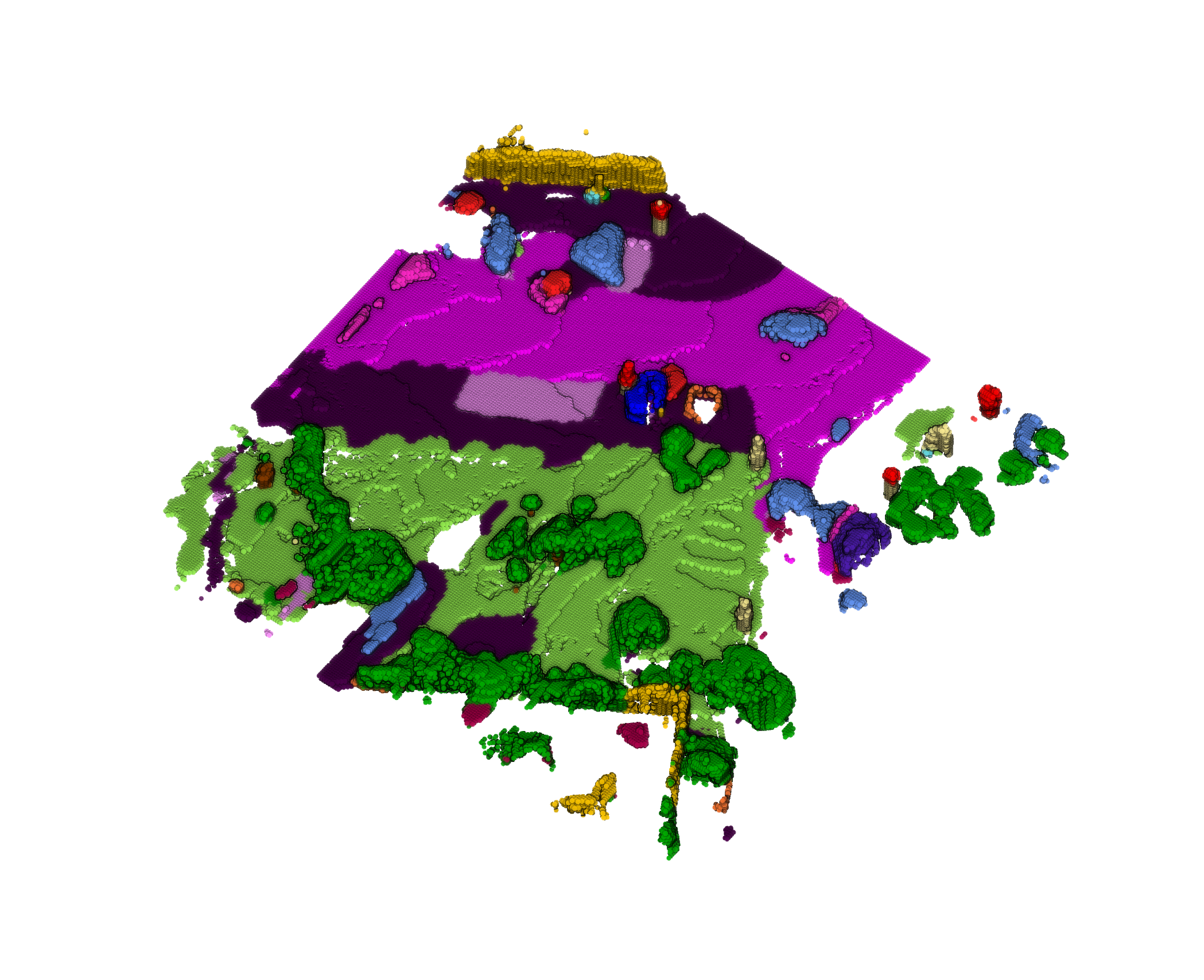} &
\qualsqAmark{\qualAellOne\qualAellTwo}{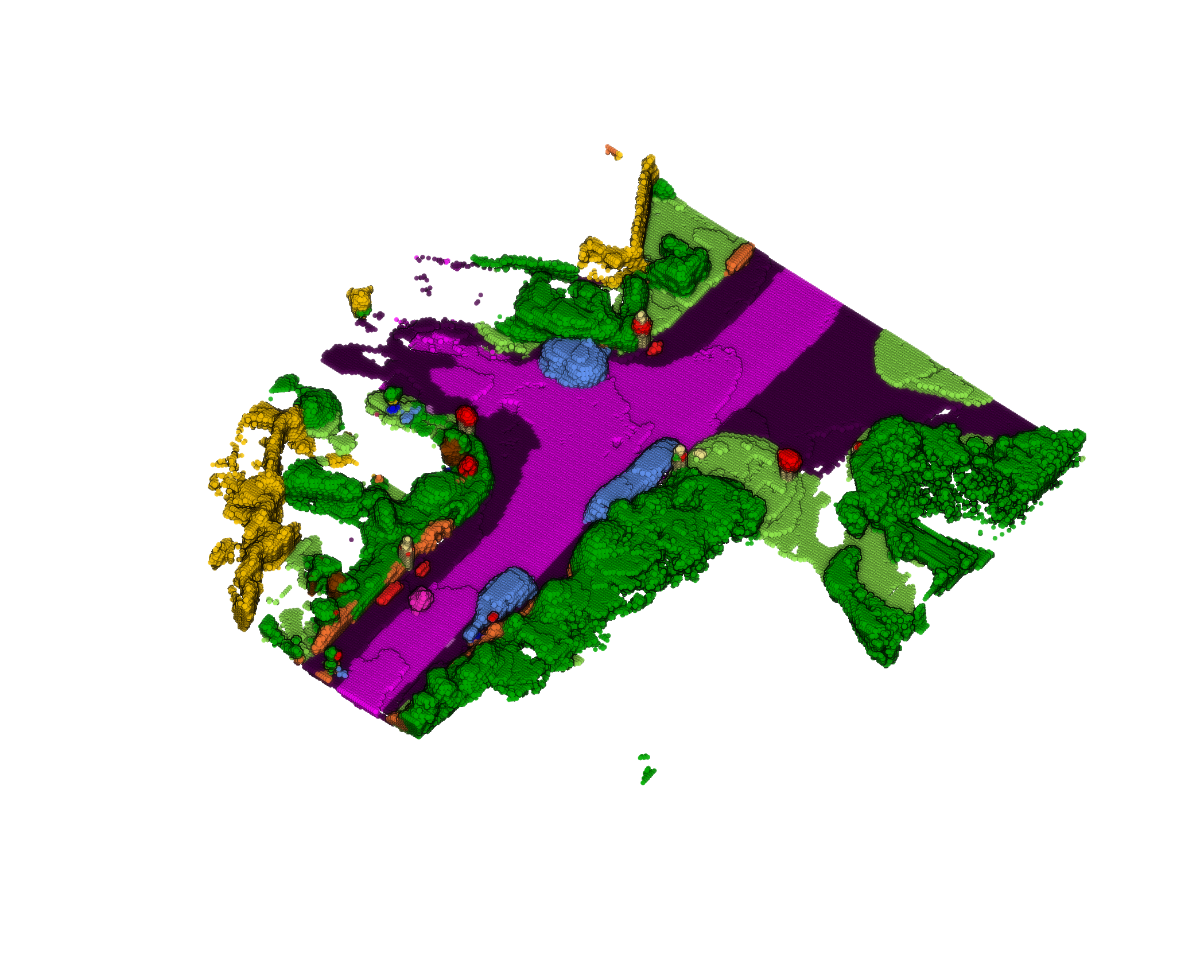} &
\qualsqCmark{\qualCellOne}{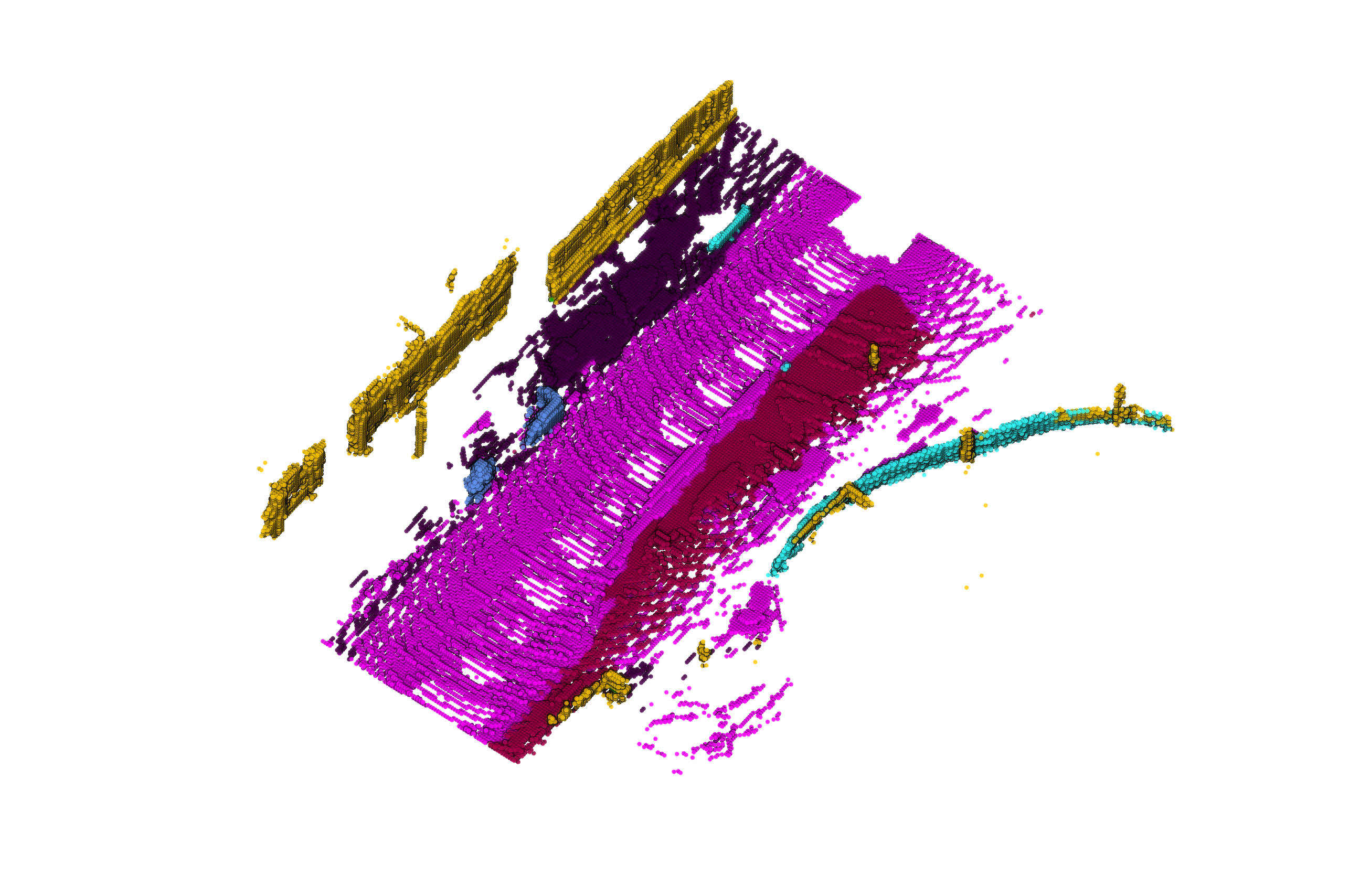} &
\qualsqDmark{\qualDellOne\qualDellTwo}{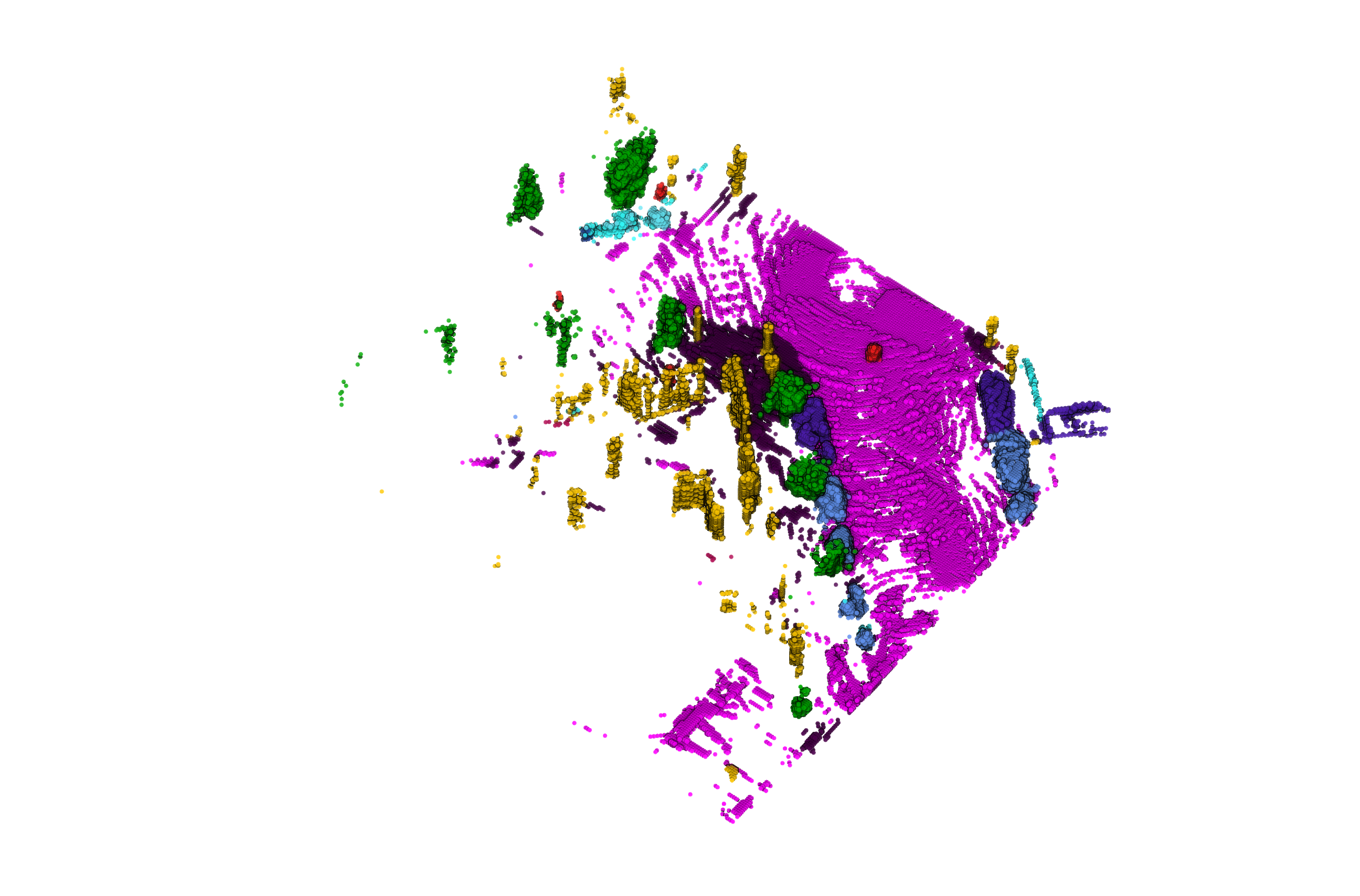} \\[1mm]
\multicolumn{6}{c}{\dashedhrule} \\[1mm]
\multirow{2}{*}{\rowlabelcenter{LMSCNet-SS}} & \rowlabel{baseline} &
\qualsqBell{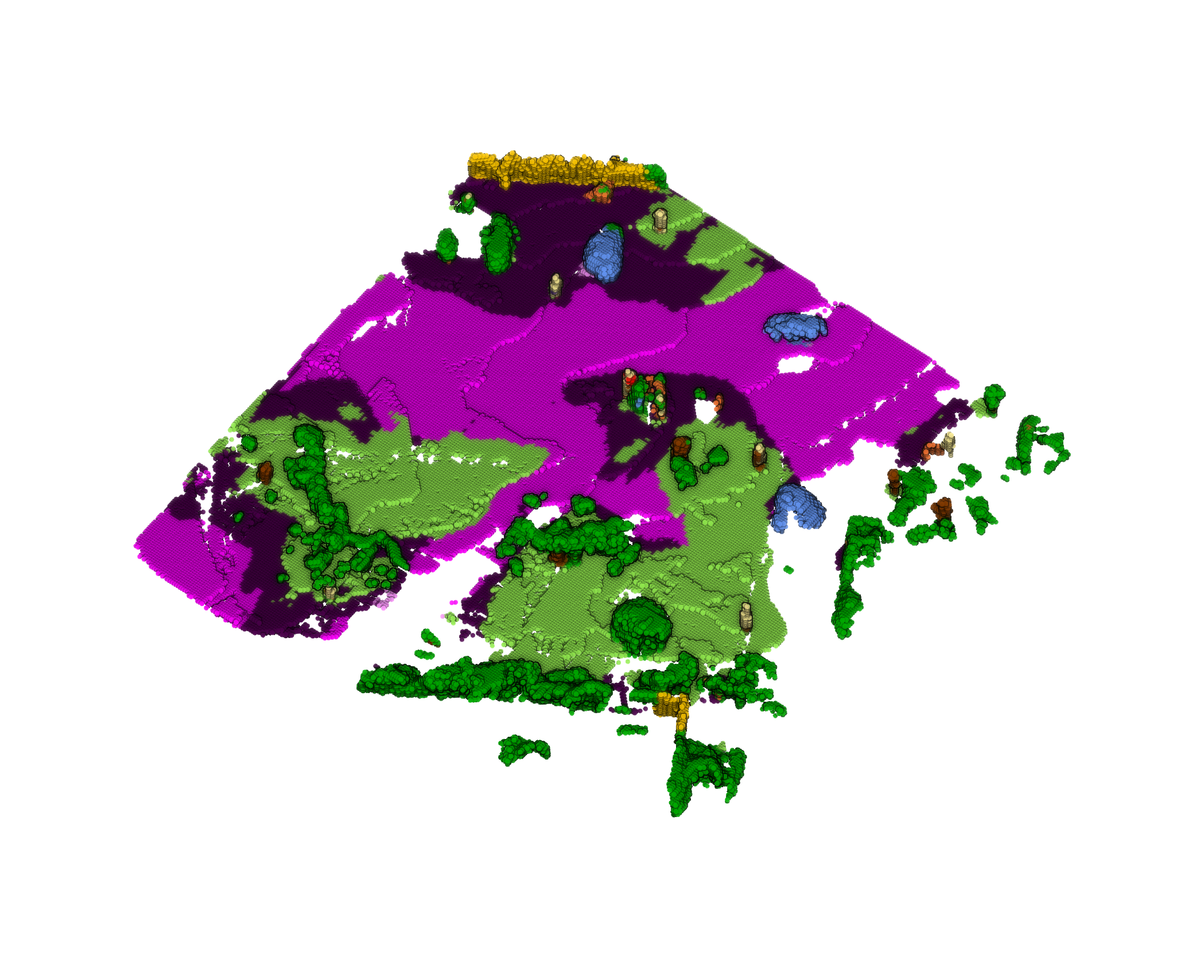} &
\qualsqAell{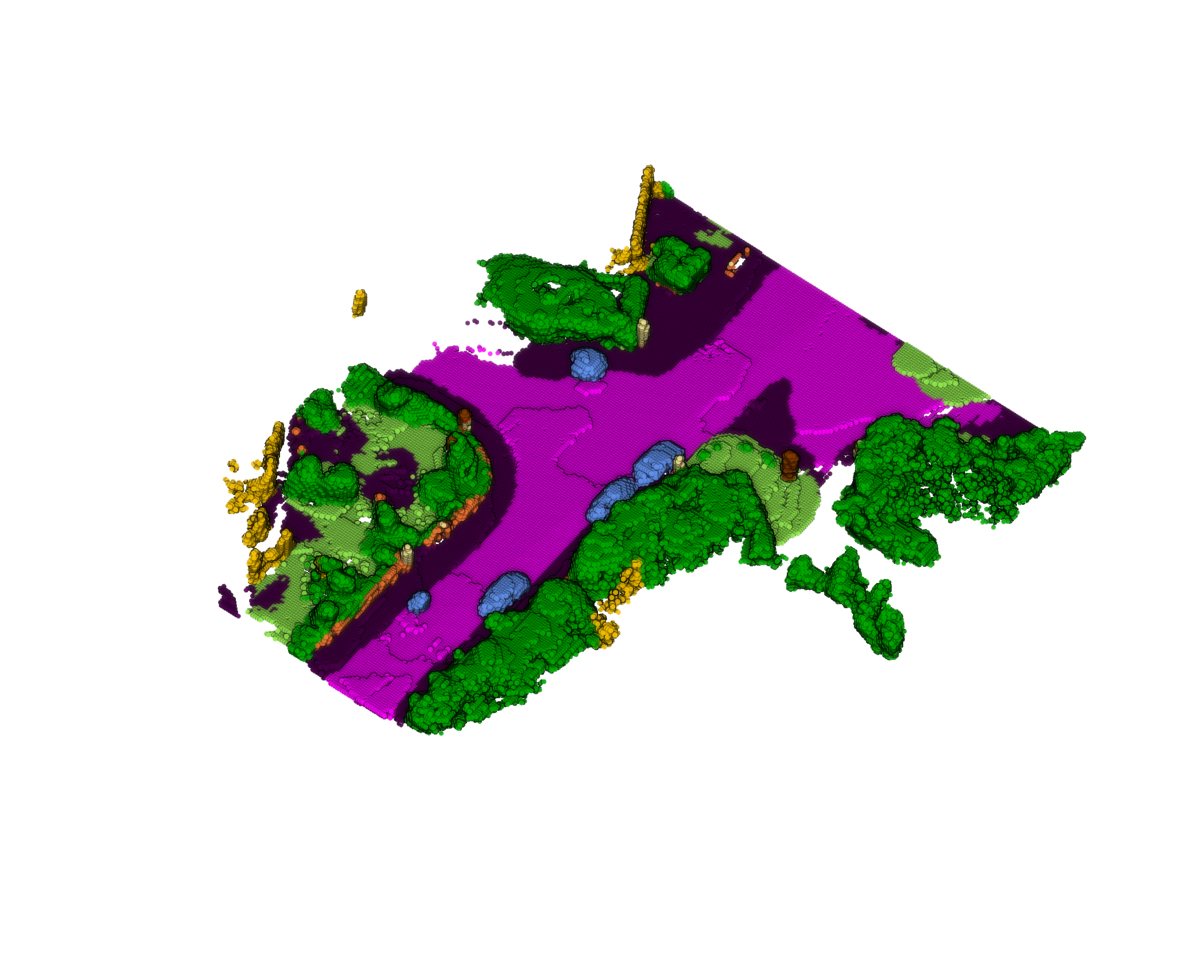} &
\qualsqCmark{\qualCellOne}{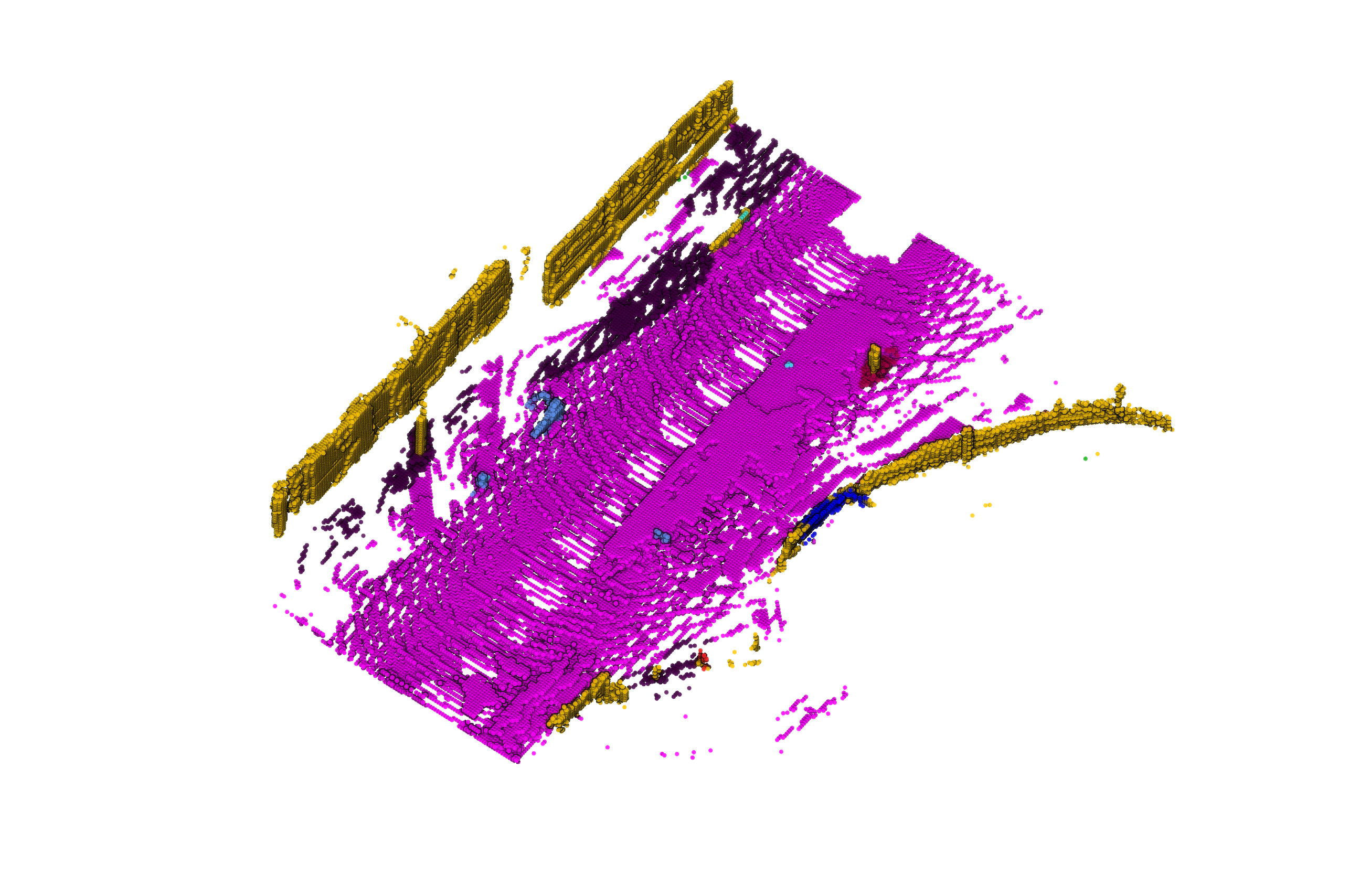} &
\qualsqDmark{\qualDellOne\qualDellTwo}{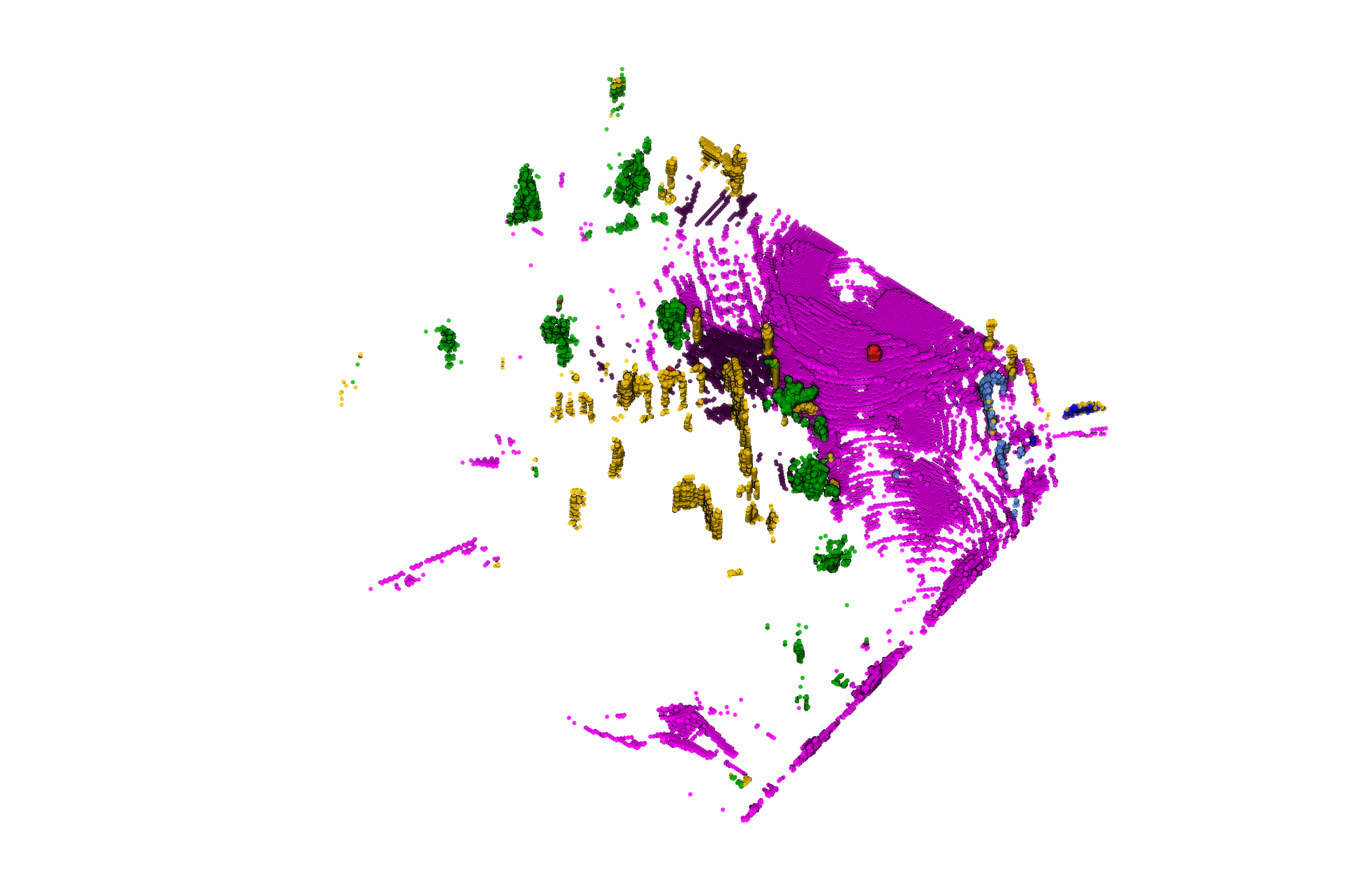} \\
 & \rowlabel{+ our priors} &
\qualsqBell{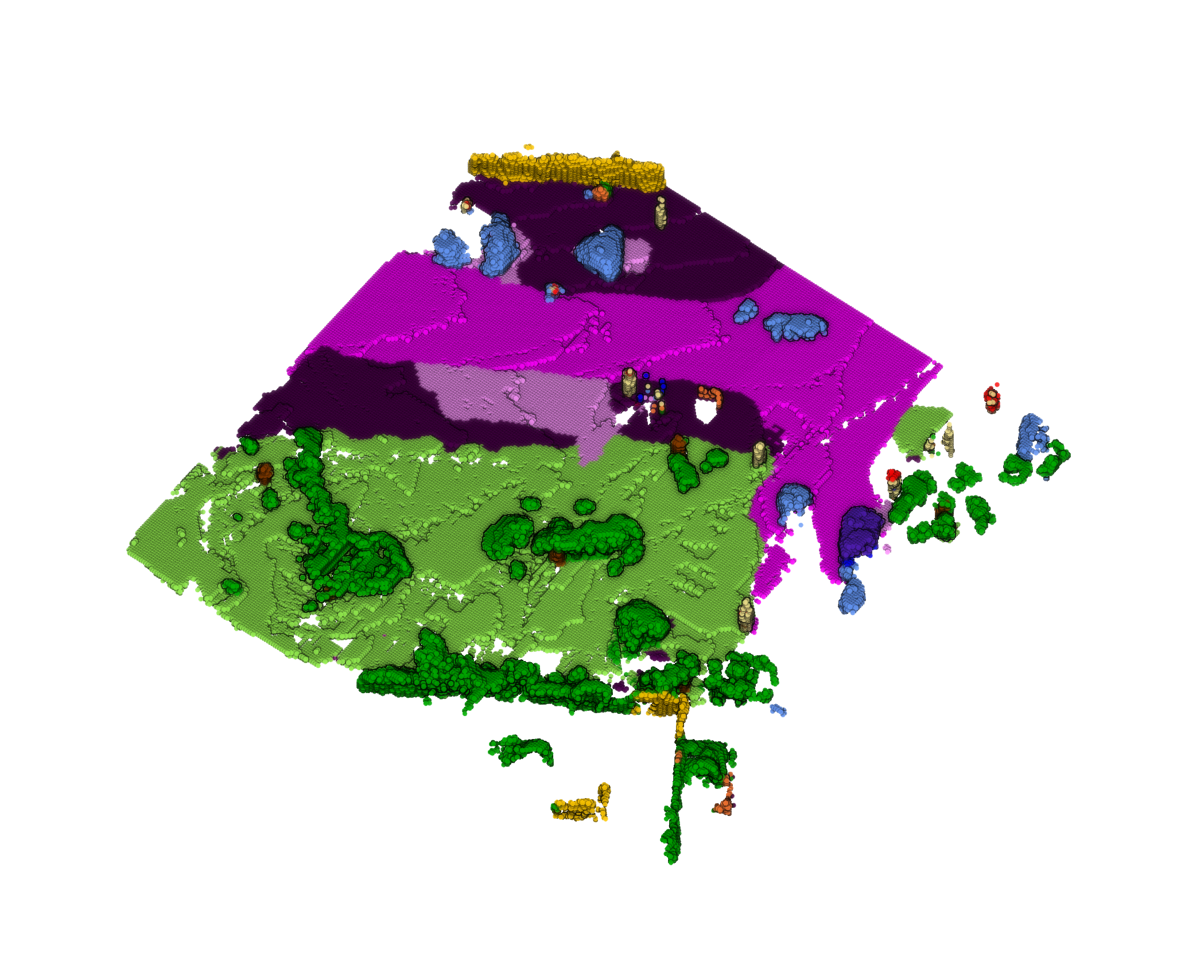} &
\qualsqAell{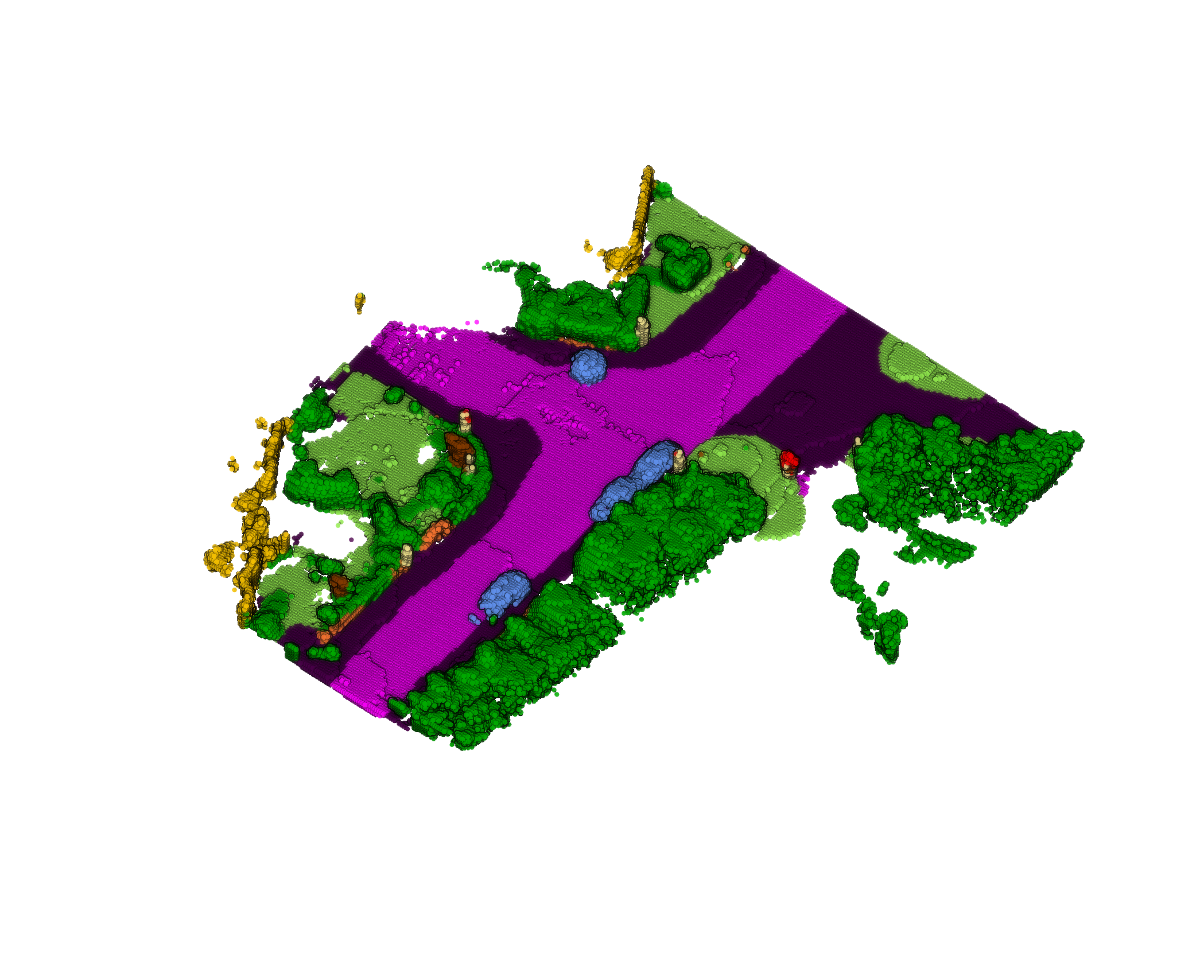} &
\qualsqCmark{\qualCellOne}{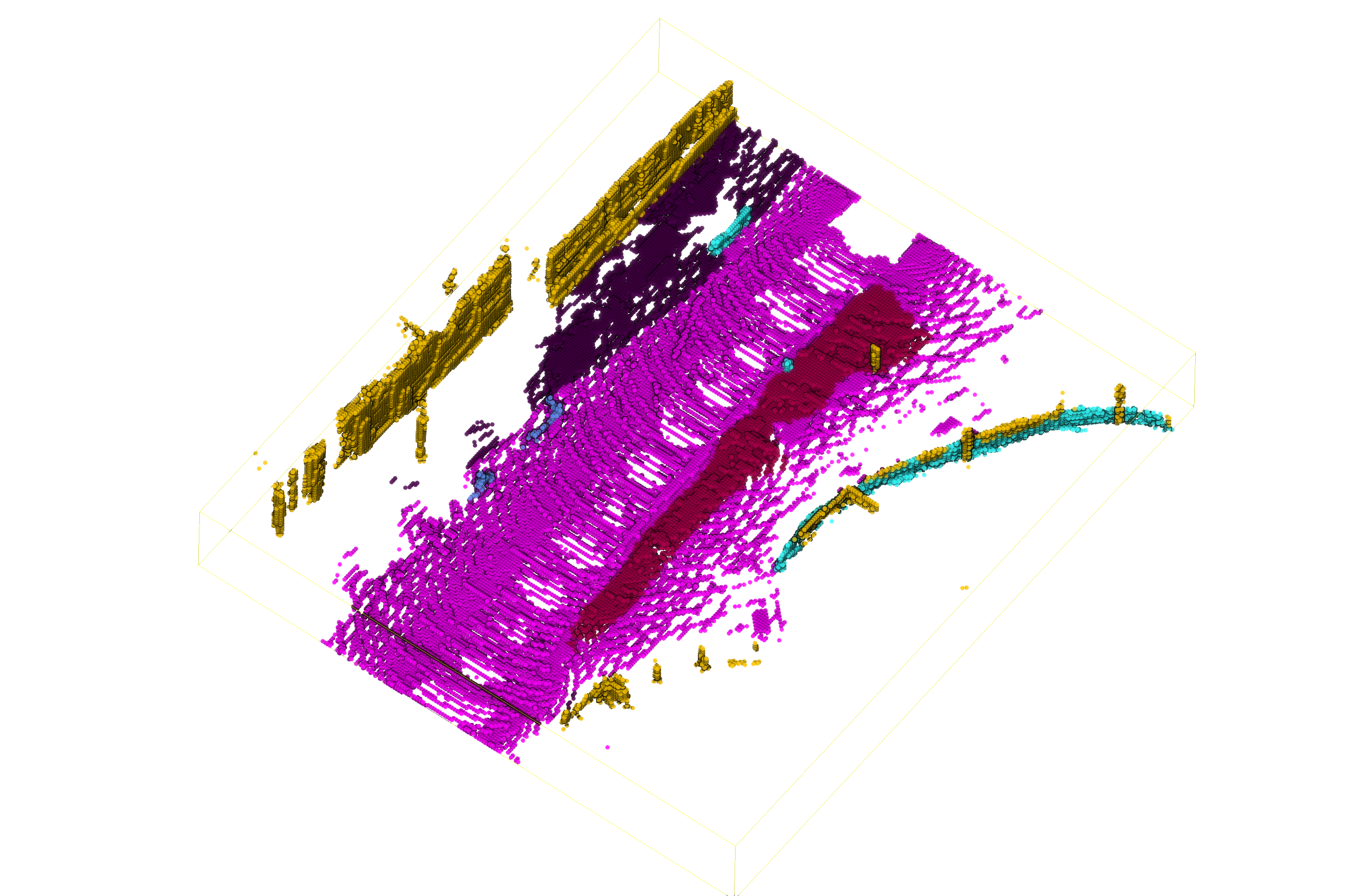} &
\qualsqDmark{\qualDellOne\qualDellTwo}{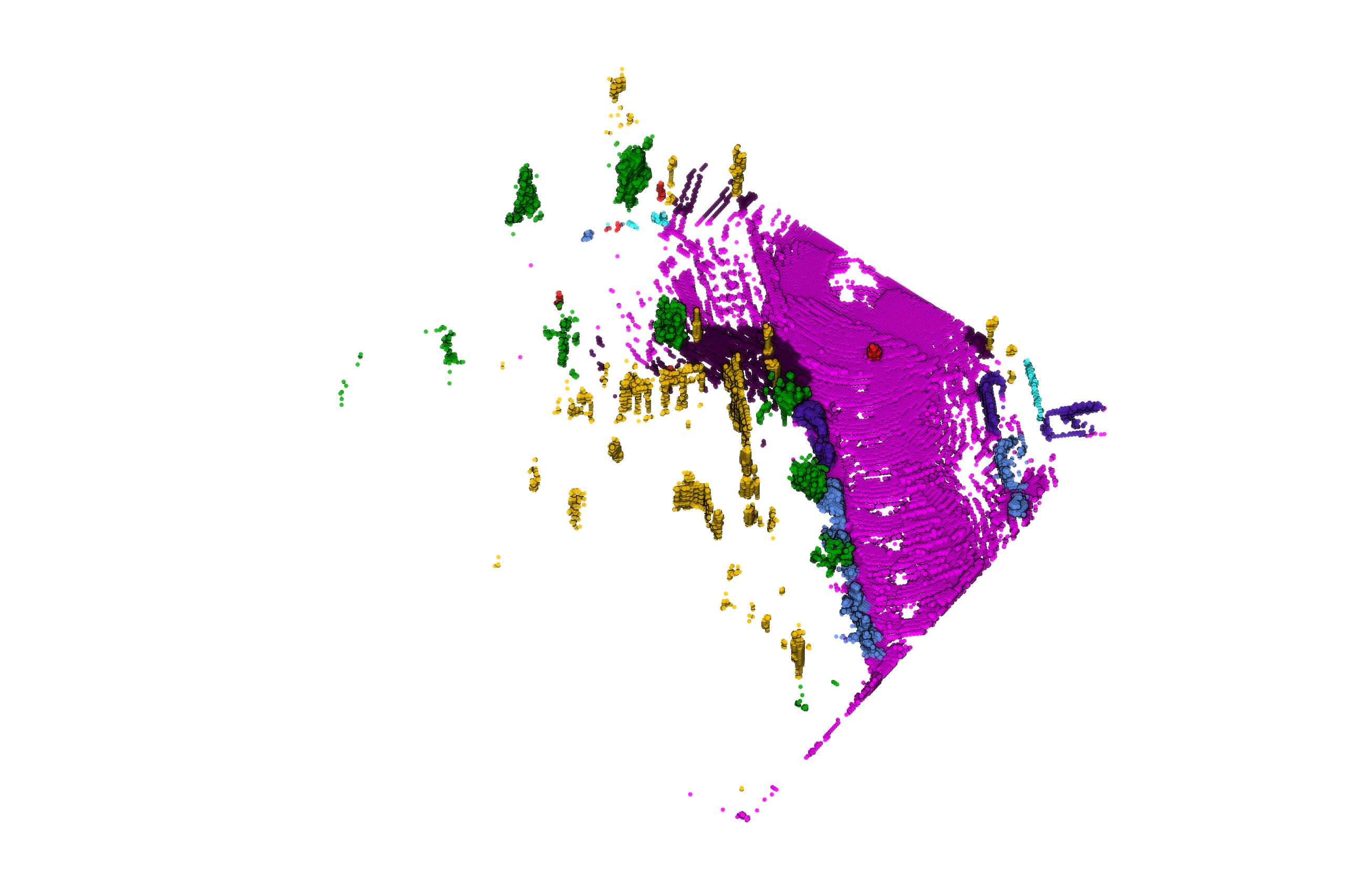} \\
\midrule
\multicolumn{2}{c}{\rowlabel{GT}} &
\qualsqBell{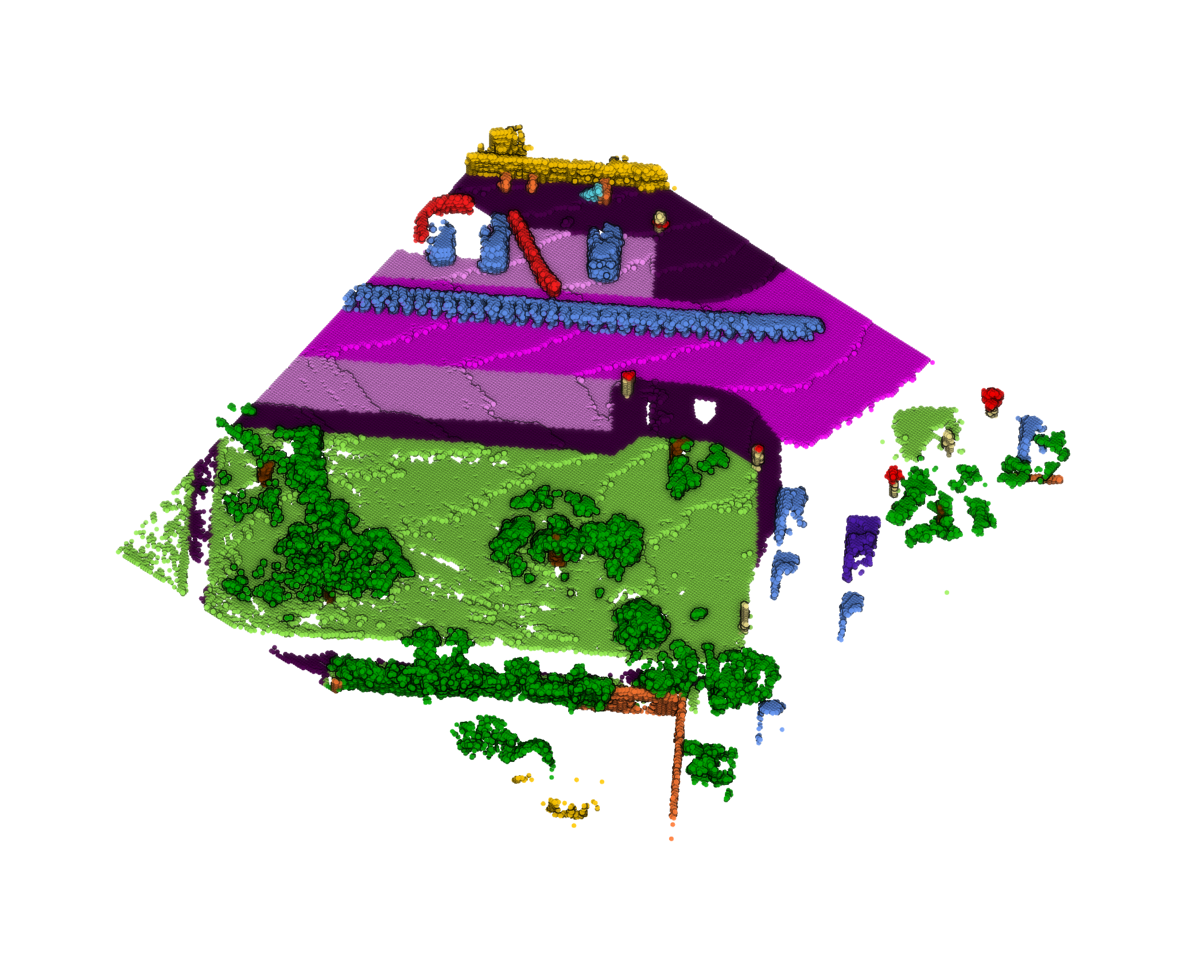} &
\qualsqAmark{\qualAellOne\qualAellTwo}{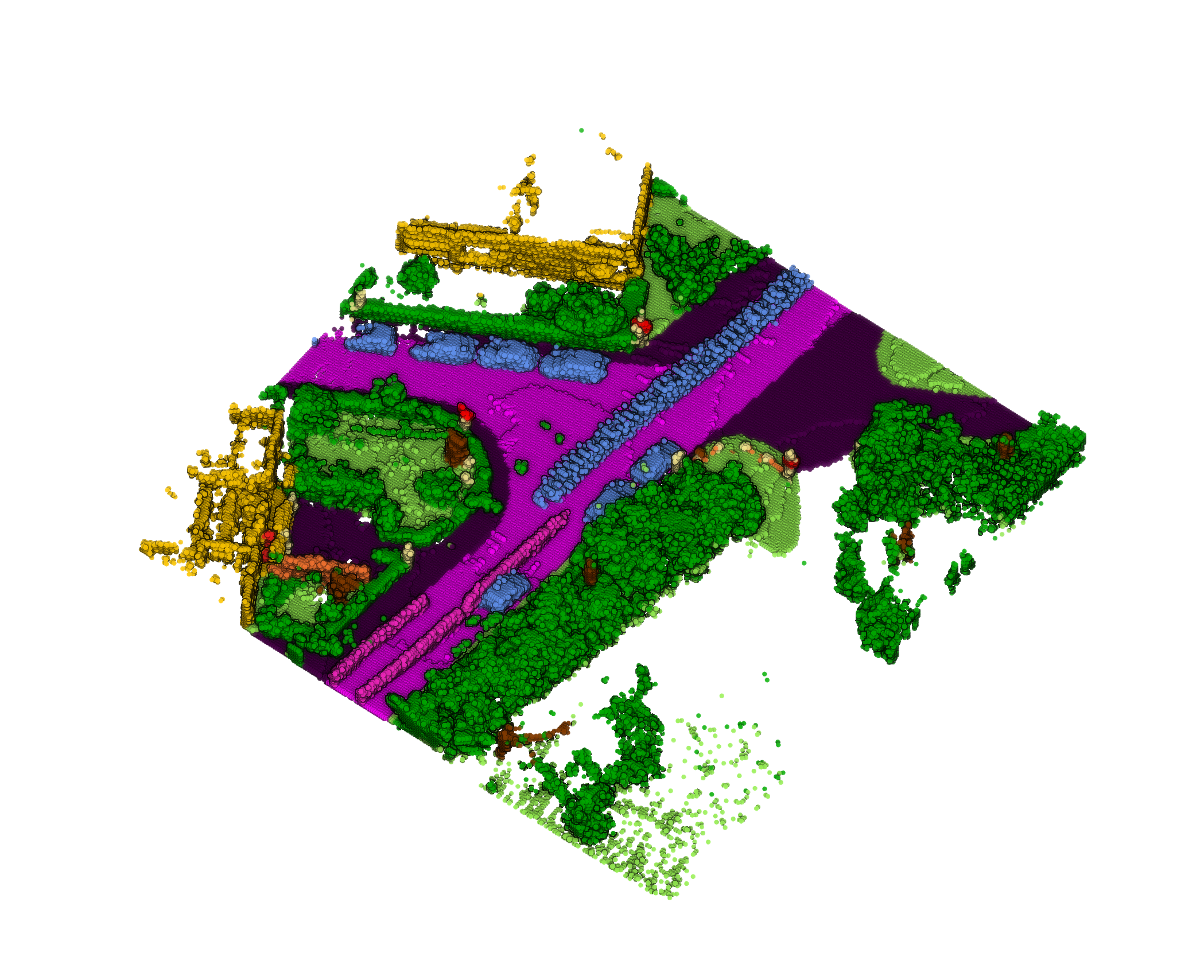} &
\qualsqCmark{\qualCellOne}{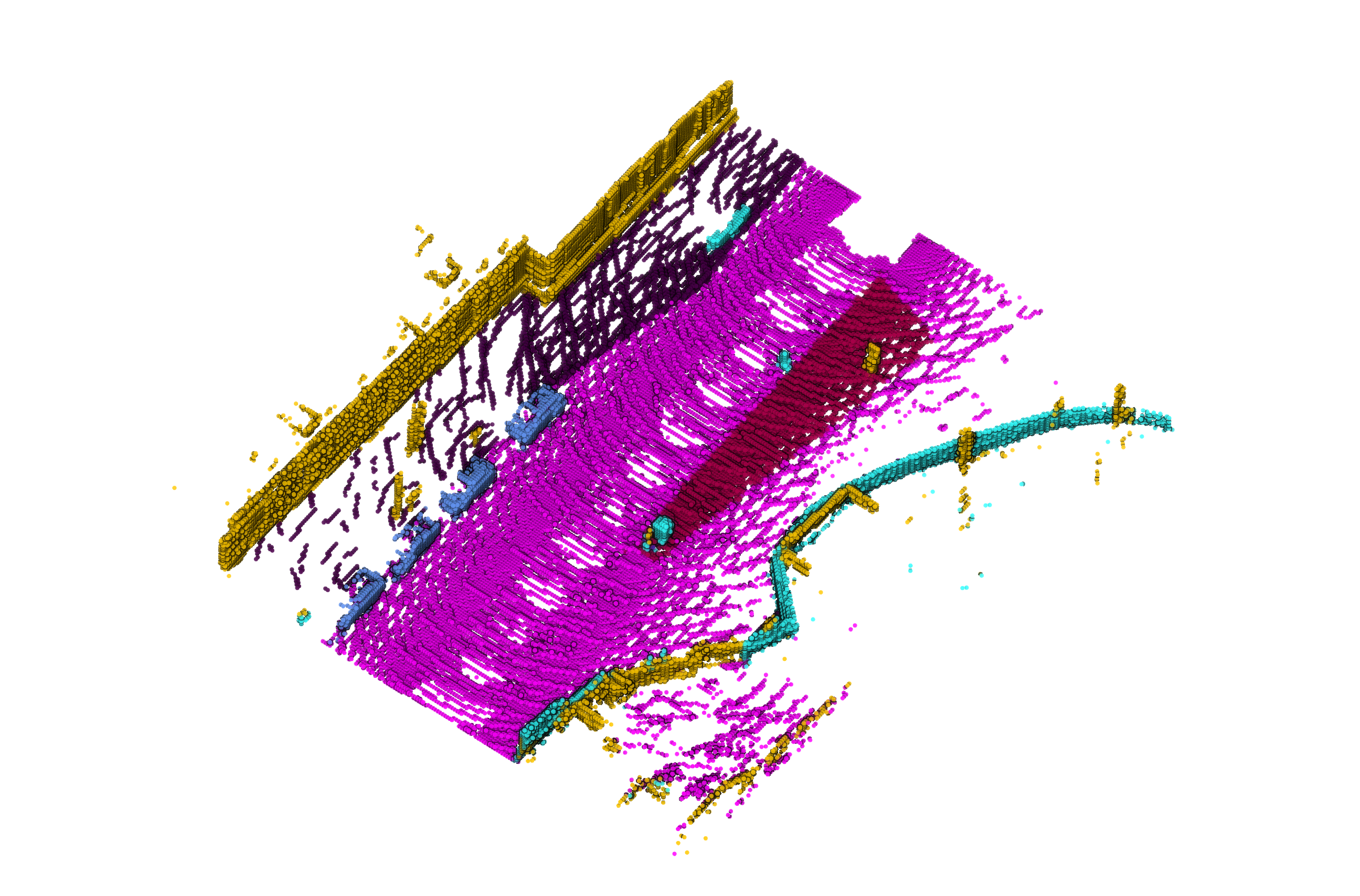} &
\qualsqDmark{\qualDellOne\qualDellTwo}{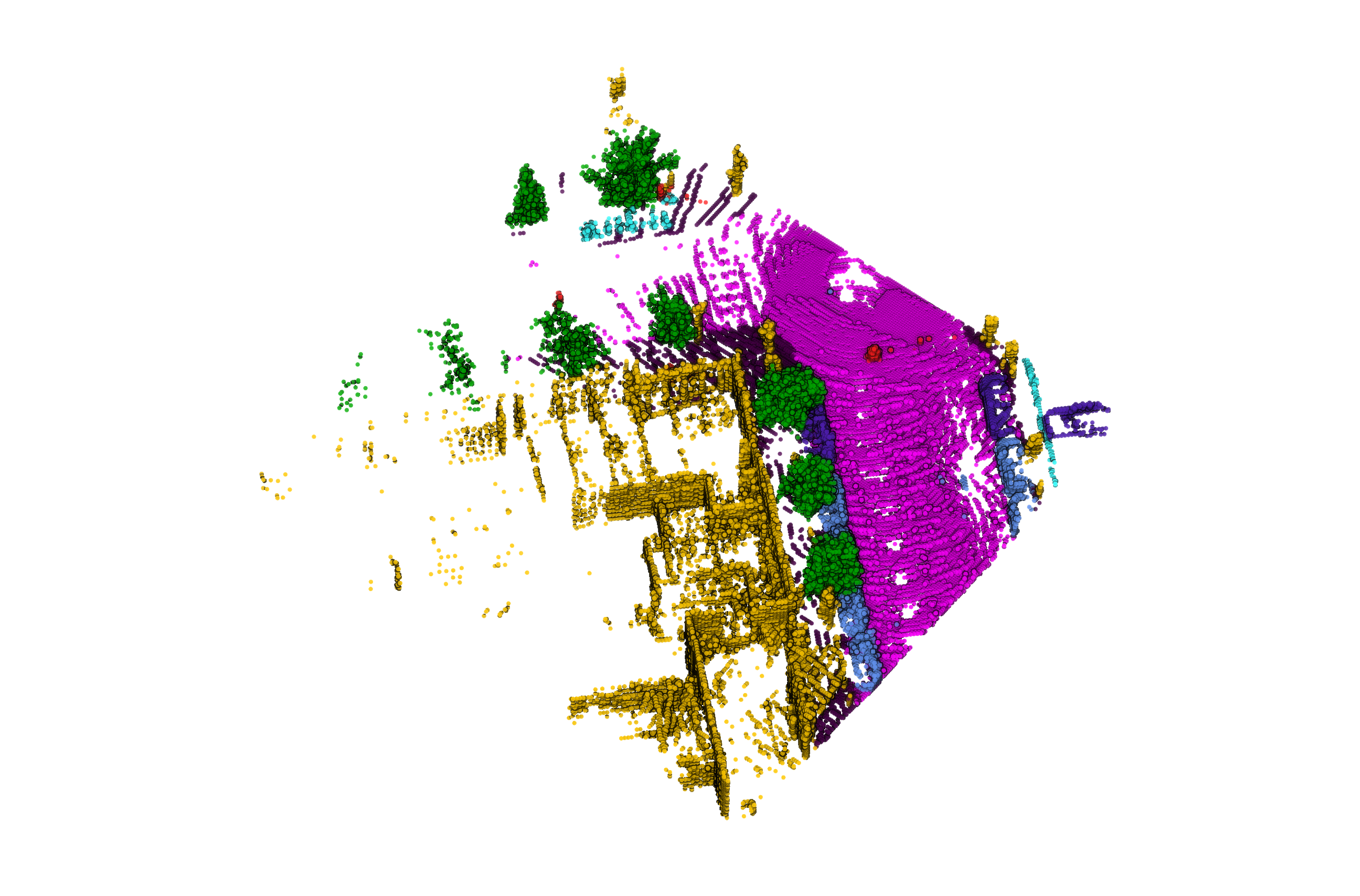} \\
\end{tabular}
\caption{
\textbf{Qualitative SSC results across datasets and backbones.}
Each column is one validation scene. For each backbone, rows show the prior-free \emph{baseline} and the same backbone equipped with our input priors (\emph{+ our priors}), flanked by the input point cloud (\emph{input}) and the ground truth (\emph{GT}).
\textbf{(A)} SemanticKITTI (64-beam) and \textbf{(B)} the sparser SSCBench-nuScenes (32-beam), two scenes each (columns), for the two main backbones SemCity-AE and LMSCNet-SS (row pairs).
Best viewed zoomed in.
}
\label{fig:qualitative}
\end{figure*}

%% file: tables/segmenter-agnostic-deployment.tex
\providecommand{\PENDING}{\textcolor{red}{--}}
\begin{table}[t]

\caption{\textbf{Segmenter-agnostic training.} We study the effect of training with GT semantic labels rather than with the (pseudo-)labels of a segmenter. At inference, only the segmenter labels are used. We measure on SemanticKITTI the performance for the full shipped recipe, when the semantic prior is combined with the dense visibility prior (not with the visibility oracle).
$\Delta$~measures the difference of performance between the model that trains on the GT and the model that trains on the segmenter labels. $^\dagger$retrained.
}
\label{tab:segmenter_agnostic_deployment}

\centering
\setlength{\tabcolsep}{4pt}
\begin{tabular}{l|l|cc|cc}
\toprule
    \multicolumn{2}{c|}{1-frame input} &  \multicolumn{4}{c}{SSC output~~} \\[1pt]

     \multicolumn{1}{c|}{sem.\,prior} & \multicolumn{1}{c|}{sem.\,prior} &  \multicolumn{2}{c|}{geometry} & \multicolumn{2}{c}{semantics}\\

\multicolumn{1}{c|}{@\,training} & \multicolumn{1}{c|}{@\,inference} & IoU\,\% & $\Delta$ & mIoU\,\% & $\Delta$ \\
\midrule
\rowcolor{\colortitle}
\multicolumn{6}{l}{\textbf{LMSCNet-SS}~\cite{lmscnet}$^\dagger$}\\
MinkUNet   & MinkUNet   & 57.2 & - & 24.0 & -  \\ %
GT         & MinkUNet   & 57.4 & \textcolor{MyGreen}{+0.2} & 23.8 & \textcolor{MyRed}{$-$0.2} \\ %
\cline{1-2}
&&&&&\\[-1.9ex]
WI  & WI  & 57.9 & - & 23.8 & -  \\ %
GT         & WI  & 57.4 & \textcolor{MyRed}{$-$0.5} & 24.0 & \textcolor{MyGreen}{+0.2} \\ %
\cline{1-2}
&&&&&\\[-1.9ex]
WI  & WI-TTA     & 57.9 & - & 24.1 & -  \\ %
\rowcolor{\colormain}
GT         & WI-TTA     & 57.5 & \textcolor{MyRed}{$-$0.4} & 24.6 & \textcolor{MyGreen}{+0.5} \\ %
\cline{1-2}
&&&&&\\[-1.9ex]
\rowcolor{\colorspecial}
GT         & GT         & 58.2 & - & 30.3 & -  \\ %
\midrule
\rowcolor{\colortitle}
\multicolumn{6}{@{}l}{\textbf{SemCity-AE}~\cite{semcity}$^\dagger$}\\
MinkUNet   & MinkUNet   & 56.8 & - & 25.5 & -  \\ %
GT         & MinkUNet   & 56.6 & \textcolor{MyRed}{$-$0.2} & 26.1 & \textcolor{MyGreen}{+0.6} \\ %
\cline{1-2}
&&&&&\\[-1.9ex]
WI  & WI  & 57.0 & - & 25.5 & -  \\ %
GT         & WI  & 56.7 & \textcolor{MyRed}{$-$0.3} & 26.1 & \textcolor{MyGreen}{+0.6} \\ %
\cline{1-2}
&&&&&\\[-1.9ex]
WI  & WI-TTA     & 57.0 & - & 25.9 & -  \\ %
\rowcolor{\colormain}
GT         & WI-TTA     & 56.8 & \textcolor{MyRed}{$-$0.2} & 26.9 & \textcolor{MyGreen}{+1.0} \\ %
\cline{1-2}
&&&&&\\[-1.9ex]
\rowcolor{\colorspecial}
GT         & GT         & 57.1 & - & 34.4 & -  \\ %
\bottomrule
\end{tabular}

\end{table}

%% file: sections/006_conclusion.tex
\section{Conclusion}
\label{sec:conclusion}

We revisited lidar semantic scene completion through two input priors: off-the-shelf semantic pseudo-labels and ray-cast visibility information. These priors are supplied purely at input level and kept decoupled from the rest of the SSC network. %

We summarize below the main takeaways %
of our study, based on experiments with four SSC methods, four semantic segmenters, and two datasets relying on different lidar sensors.
\setlist{nolistsep}
\begin{itemize}[itemsep=3pt]

    \item %
    \textbf{Our prior augmentation is effective.} It systematically improves both completion and semantics, lifting %
    older lightweight models to the level of recent state-of-the-art systems, among fully-reproducible single-frame SSC methods. %

    \item %
    \textbf{Our prior augmentation is simple.} It basically applies to any SSC method at no architectural cost beyond widening the input layer, and basically any off-the-shelf semantic and visibility cues can be used as priors. Moreover, our approach is segmenter-agnostic not only because any segmenter yields gains, but also in that the prior-augmented model can be trained just once, with ground-truth point labels, and later used with any segmenter.
    
    \item \textbf{The quality of prior-augmented SSC grows with input prior accuracy}. Outsourcing semantic labels thus offers better SSC for free whenever segmenters improve. Besides, the visibility prior seems to benefit from lidar density, while the semantic prior does not directly depend on it.

    \item \textbf{Our prior augmentations benefit more the lightweight models.} 
    Stronger backbones gain less, but still substantially, unless they already compute comparable semantics internally, in which case the prior adds little.

    \item \textbf{Semantics is the dominant bottleneck.}
    It is indeed the area where there is the most room for improvement, with gains up to 12 mIoU pts, while completion does not improve more than 3 IoU pts.
    Moreover, the oracle input semantics establishes a performance upper bound far above the prior-augmented baselines, which opens perspectives for even further improvements.

\item \textbf{Semantic cues have a small effect on geometric completion} in the networks we studied, despite architectural connections in their design: even ground-truth input semantics change the IoU by at most 1.1 pts. 
Conversely, visibility cues do raise the mIoU, by up to 8.4 pts with the visibility oracle, although this partly reflects that the mIoU also accounts for geometric errors. 
This nuances one of the common motivations to address both tasks jointly in SSC, and highlights a topic to consider when designing a new SSC architecture.%

\end{itemize}    
We release our priors and integration code to make these input-level gains easy to adopt across SSC backbones. Our code is available at https://github.com/astra-vision/SSC-Priors. 

Besides exploring the oracle headrooms and tightening geometry and semantics coupling, other perspectives include using multiple frames (\textit{e.g.}, from the past) and learning the visibility prior to reduce the free-space marking noise.

%% file: sections/007_appendix.tex
\appendix

\section{Conference paper extension}
\label{app:conf_paper_extension}
\textcolor{purple}{
This article is an extended version of our conference paper %
\emph{``Exploring Easy Boosts for Lidar Semantic Scene Completion''} \cite{martyniuk2026exploring}. 
The conference version studies the two input-level priors (semantic pseudo-labels and ray-cast visibility), with experiments only on SemanticKITTI~\cite{semantickitti}}. %

This journal version has been largely rewritten and %
adds the following:
\begin{enumerate}[itemsep=3pt]
    \item a systematic study of the design space of the visibility prior
    (\autoref{subsec:vis}); 
    \item experiments on another dataset featuring a different lidar sensor, namely SSCBench-nuScenes, which confirm the results obtained on SemanticKITTI (Sec.~\ref{sec:nuscenes});
    \item an architectural account of the uneven gains across different SSC methods %
    (\autoref{subsec:quantitative});
    \item a deeper analysis of the combination of the semantic and visibility priors (\autoref{subsec:quantitative});
    \item an improved treatment and discussion on the prior oracle headroom for both semantics and visibility (\autoref{subsec:quantitative}, \autoref{sec:agnostic});
    \item a discussion of the lesser impact of the visibility prior on SSCBench-nuScenes, which we find attributable to fewer and noisier visibility cues (Sec.~\ref{sec:nuscenes}); %
    \item a segmenter-agnostic training analysis showing that the completion network need not be retrained when the semantic segmenter is changed (\autoref{sec:agnostic});
    \item experiments showing that, on the contrary, training with GT visibility degrades the performance when testing with actual, noisy visibility cues (\autoref{app:vis_mismatch});
    \item an inference time study (\autoref{app:profiling});
    \item a substantially expanded related-work discussion (\autoref{sec:related}).
\end{enumerate}

\section{Inference time profiling}
\label{app:profiling}

\input{tables/inference_visibility_priors_abs}

\paragraph*{Methodology.}
We profile each variant on 50 consecutive lidar frames of SemanticKITTI
sequence 08. 
Each method is run three times per frame (150 measurements per method), preceded by a single warm-up sweep whose timings are discarded. 
The order in which methods are invoked is randomized per frame under a fixed seed, so position-dependent effects do not bias any one method. 
Each method is split into two independently timed stages: \emph{ray sampling}, which generates the free-space markers along each sensor ray; and \emph{voxelization}, which clips markers and returns to the $256\times256\times32$ grid at 0.2\,m, computes per-voxel flat indices, accumulates per-class vote counts, resolves each voxel by majority vote under the returns-override-free-space rule, and applies any method-specific refinement (marking as free voxels found as empty). 
End-to-end latency is timed separately by wrapping the full call; averaged over all 150 measurements it matches the sum of the per-stage means to within 1.8\% ($\le 0.07$\,ms) for every method, confirming that the two stages account for essentially all the runtime.
We report the independently measured end-to-end value; the per-stage means in \autoref{tab:visibility_inference_time_gpu} indicate each stage's contribution.
All measurements run on an NVIDIA A100 GPU (PCIe, 40\,GB) under \texttt{torch.inference\_mode()}. 
Frames are loaded and moved to the GPU once before timing, so disk I/O is excluded, and CUDA is synchronized before and after each timed region.

\paragraph*{Results.}
\autoref{tab:visibility_inference_time_gpu} decomposes each variant's end-to-end latency into its ray-sampling and voxelization stages. 
Voxelization dominates every variant; ray sampling takes at most 1.2\,ms for all sparse variants and becomes a substantial share only for the dense variants. 
Voxelization alone, with no markers to place, costs 3.2\,ms/frame. 
Single-marker variants add 0.4\,ms of sampling, while voxelization stays at 3.0\,ms (within the run-to-run variation of the marker-free pass), reaching 3.4\,ms. 
Across the $n$-random family both stages grow with $n$, sampling from 0.4\,ms at $n{=}1$ to 1.2\,ms at $n{=}100$ and voxelization from 3.0 to 11.2\,ms. 
The dense prior places fewer markers than 100-random, as its number of steps is bounded by the range of each return, so it voxelizes faster (8.9 vs.\ 11.2\,ms), although its markers, spaced one voxel apart, mark more distinct voxels. 
Its sampling is slower (5.8\,ms), because it builds the candidate steps of the longest ray for every ray before it discards the steps beyond each return, which brings it to 14.7\,ms end-to-end, 4.6$\times$ the marker-free pass. 
Adding the dilation step costs a further 0.5\,ms of voxelization, at 15.2\,ms.

Every variant is therefore cheap relative to the rest of the pipeline: the completion networks take from 33\,ms/frame (SSA-SC) to 665\,ms/frame (JS3C-Net) (\autoref{tab:runtimefwd}), and the segmenter that supplies the semantic prior takes 155\,ms/frame for WaffleIron and 1871\,ms/frame with TTA (\autoref{tab:runtimesemseg}). 
With both priors, the dense visibility prior accounts for about 7\% of the runtime at most (SSA-SC with WaffleIron), and for less than 1\% with WaffleIron-TTA. 
The dense prior is the default: it gives the best or near-best accuracy on both networks (\autoref{tab:lidar_vis_ablation}), it has no sample-count hyperparameter, and it costs only 2.3\,ms/frame more than 100-random.

\section{Per-class results}
\label{app:perclass}
\input{tables/table_semkitti_per_class}
\input{tables/table_12_class_per_class}
\autoref{tab:perclass_semkitti} and \autoref{tab:perclass_nuscenes12} report per-class IoU on SemanticKITTI and SSCBench-nuScenes, respectively, for the prior combinations of \autoref{tab:semkitti_impact} and \autoref{tab:nuscenes12}; the SemanticKITTI table adds the oracle visibility rows without semantics and with WaffleIron.
On both datasets, the semantic prior's gains concentrate on the classes the lightweight baselines predict worst unaided (e.g., \emph{truck} on SemanticKITTI rises from 3.7 to 30.9 IoU for LMSCNet-SS and from 6.9 to 46.0 for SemCity-AE), consistent with the dataset-level results of \autoref{tab:semkitti_impact} and \autoref{tab:nuscenes12}.

\section{Visibility oracle training}
\label{app:vis_mismatch}

\input{tables/visibility-agnostic-deployment}

\Tableref{tab:visibility_agnostic_deployment} repeats the study of \autoref{sec:agnostic} on the visibility axis, with the semantic prior held at WaffleIron. 
Unlike the semantic prior, the visibility prior does not survive the swap. 
A network trained on the oracle prior and given the dense prior at inference loses $21.7$ IoU pts on LMSCNet-SS and $11.9$ on SemCity-AE, whereas the equivalent swap on the semantic axis costs at most $0.2$ mIoU pts, and improves mIoU in most of the studied cases (\autoref{tab:segmenter_agnostic_deployment}).
Both networks lose occupancy recall: $0.66\rightarrow0.38$ on LMSCNet-SS, and $0.76\rightarrow0.53$ on SemCity-AE. 
Occupancy precision does not degrade, and increases from $0.82$ to $0.87$ and from $0.70$ to $0.75$, respectively. 
Our interpretation is that a network trained on ground-truth free space does not predict occupancy in the wrong place, it just predicts too little occupancy. 
The oracle prior is therefore a ceiling, and not a training-time substitute for the deployed prior.

%% file: tables/inference_visibility_priors_abs.tex
\begin{table}[tb]
    \caption{\textbf{Inference time (ms/frame) of a forward pass of SSC models} on the SemanticKITTI validation set, excluding I/Os and voxelization, measured on an NVIDIA A100 GPU (PCIe, 40 GB) after warm-up, averaged over 50 frames $\times$ 3 runs.}
    \label{tab:runtimefwd}
\centering
\begin{tabular}{l|r|c|r}
\toprule
& \multicolumn{1}{c|}{vanilla} & $\Delta$ & \multicolumn{1}{c}{prior aug.\!\!\!\!}  \\
\midrule
SSA-SC~\cite{ssasc}       &  33 $\pm\,\,$ 4  & +4 &  37 $\pm\,\,$ 4 \\
LMSCNet-SS~\cite{lmscnet} &  42 $\pm\,\,$ 1  & +5 &  47 $\pm\,\,$ 1 \\
SemCity-AE~\cite{semcity} & 162 $\pm\,\,$ 1  & +1 & 163 $\pm\,\,$ 1 \\
JS3C-Net~\cite{js3cnet}   & 665 $\pm\,$88    & +2 & 667 $\pm\,$88 \\
\bottomrule
\end{tabular}

\begin{tabular}{p{7cm}}\\\end{tabular}

    \caption{\textbf{Inference time (ms/frame) of semantic segmenters} on the SemanticKITTI validation set, excluding I/Os, measured on an NVIDIA A100 GPU (PCIe, 40 GB) after warm-up and averaged over 50 frames $\times$ 3 runs. $\pm$: standard deviation.}
    \label{tab:runtimesemseg}
\centering
\begin{tabular}{l|r}
\toprule
Semantic segmenter & \multicolumn{1}{c}{full infer.\!\!\!} \\
\midrule
SalsaNext~\cite{salsanext}            &   9 $\pm\,\,$ 1 \\
WaffleIron~\cite{waffleiron}          & 155 $\pm\,\,$ 6 \\
MinkUNet~\cite{minkunet}              & 164 $\pm\,\,$ 4 \\
WaffleIron\,+\,TTA~\cite{waffleiron}  & 1871 $\pm\,$57 \\
\bottomrule
\end{tabular}

\begin{tabular}{p{6cm}}\\\end{tabular}

    \caption{\textbf{Inference time (ms/frame) of visibility priors} on the SemanticKITTI validation set, excluding I/Os, measured on an NVIDIA A100 GPU (PCIe, 40\,GB) after warm-up and averaged over 50 frames $\times$ 3 runs: ray sampling (sampl.), voxelization (voxel.), end-to-end (total).}
    \label{tab:visibility_inference_time_gpu}

   \setlength{\tabcolsep}{3pt}
    \centering
    {
    \begin{tabular}{c|l|ccc}
    \toprule
    & visibility prior & sampl. & voxel. & total \\
    \midrule
    (1) & none                  & 0 & 3.2 & 3.2 \\
    (2) & 1 pt before the hit   & 0.4 & 3.0 & 3.4 \\
    (3) & 1 pt at random        & 0.4 & 3.0 & 3.4 \\
    (4) & 10 pts at random      & 0.4 & 3.4 & 3.9 \\
    (5) & 25 pts at random      & 0.5 & 4.7 & 5.2 \\
    (6) & 50 pts at random      & 0.7 & 6.9 & 7.6 \\
    (7) & 100 pts at random     & 1.2 & 11.2 & 12.4 \\
\rowcolor{\colormain}
    (8) & dense ($\delta\,{=}$\,voxel size)
                                & 5.8 & 8.9 & 14.7 \\
    (9) & dense with dilation   & 5.8 & 9.4 & 15.2 \\
    \bottomrule
    \end{tabular}
    }

\end{table}

%% file: tables/table_semkitti_per_class.tex
\providecommand{\myangle}{90}
\providecommand{\PENDING}{\textcolor{red}{--}}
\providecommand{\clsq}[1]{\textcolor{#1}{$\blacksquare$}}

\definecolor{skCar}{RGB}{100,150,245}
\definecolor{skBicycle}{RGB}{100,230,245}
\definecolor{skMotorcycle}{RGB}{30,60,150}
\definecolor{skTruck}{RGB}{80,30,180}
\definecolor{skOtherVehicle}{RGB}{0,0,255}
\definecolor{skPerson}{RGB}{255,30,30}
\definecolor{skBicyclist}{RGB}{255,40,200}
\definecolor{skMotorcyclist}{RGB}{150,30,90}
\definecolor{skRoad}{RGB}{255,0,255}
\definecolor{skParking}{RGB}{255,150,255}
\definecolor{skSidewalk}{RGB}{75,0,75}
\definecolor{skOtherGround}{RGB}{175,0,75}
\definecolor{skBuilding}{RGB}{255,200,0}
\definecolor{skFence}{RGB}{255,120,50}
\definecolor{skVegetation}{RGB}{0,175,0}
\definecolor{skTrunk}{RGB}{135,60,0}
\definecolor{skTerrain}{RGB}{150,240,80}
\definecolor{skPole}{RGB}{255,240,150}
\definecolor{skTrafficSign}{RGB}{255,0,0}

\begin{table*}[p]%
  \caption{\textbf{Per-class IoU on SemanticKITTI} (validation set, 19 classes).
  Rows, notation and colors are as in \Cref{tab:semkitti_impact}.
  The percentage below each class name is the share of that class in the occupied voxels.
  All metrics are reported in \%.
  }
  \centering
  \scriptsize
  \resizebox{\linewidth}{!}{%
  \setlength{\tabcolsep}{2.7pt}
  \begin{tabular}{@{}cc||cc||*{19}{c}@{}}
    \toprule
    \textbf{Visibility} & \textbf{Semantics} & \textbf{IoU} & \textbf{mIoU}
      & \rotatebox{\myangle}{\clsq{skCar} car}
      & \rotatebox{\myangle}{\clsq{skBicycle} bicycle}
      & \rotatebox{\myangle}{\clsq{skMotorcycle} motorcycle}
      & \rotatebox{\myangle}{\clsq{skTruck} truck}
      & \rotatebox{\myangle}{\clsq{skOtherVehicle} oth.\ vehicle}
      & \rotatebox{\myangle}{\clsq{skPerson} person}
      & \rotatebox{\myangle}{\clsq{skBicyclist} bicyclist}
      & \rotatebox{\myangle}{\clsq{skMotorcyclist} motorcyclist}
      & \rotatebox{\myangle}{\clsq{skRoad} road}
      & \rotatebox{\myangle}{\clsq{skParking} parking}
      & \rotatebox{\myangle}{\clsq{skSidewalk} sidewalk}
      & \rotatebox{\myangle}{\clsq{skOtherGround} oth.\ ground}
      & \rotatebox{\myangle}{\clsq{skBuilding} building}
      & \rotatebox{\myangle}{\clsq{skFence} fence}
      & \rotatebox{\myangle}{\clsq{skVegetation} vegetation}
      & \rotatebox{\myangle}{\clsq{skTrunk} trunk}
      & \rotatebox{\myangle}{\clsq{skTerrain} terrain}
      & \rotatebox{\myangle}{\clsq{skPole} pole}
      & \rotatebox{\myangle}{\clsq{skTrafficSign} traffic-sign} \\
    & & & & \tiny 4.2\% & \tiny 0.1\% & \tiny \textless{}0.1\% & \tiny 0.1\% & \tiny 0.2\% & \tiny 0.3\% & \tiny 0.4\% & \tiny \textless{}0.1\% & \tiny 11.7\% & \tiny 0.9\% & \tiny 7.9\% & \tiny 0.1\% & \tiny 13.4\% & \tiny 1.1\% & \tiny 42.1\% & \tiny 1.0\% & \tiny 16.1\% & \tiny 0.3\% & \tiny 0.1\% \\
    \midrule

    \rowcolor{\colortitle}
    \multicolumn{23}{@{}l}{\textbf{LMSCNet-SS}} \\
        -          & -               & 55.6 & 16.5 & 38.2 & 0.0 & 0.0 & 3.7 & 0.0 & 0.0 & 0.0 & 0.0 & 64.8 & 10.3 & 35.4 & 0.0 & 36.3 & 9.7 & 38.7 & 12.8 & 42.7 & 20.6 & 1.2 \\
      dense        & -             & 57.5 & 18.1 & 41.3 & 0.0 & 0.0 & 7.5 & 0.2 & 0.0 & 0.0 & 0.0 & 65.7 & 16.8 & 35.7 & 1.4 & 37.5 & 11.2 & 41.9 & 15.4 & 45.2 & 20.6 & 2.7 \\
    \cline{1-2}
    &&&&&&&&&&&&&&&&&&&&&&\\[-1.9ex]
      -            & WI            & 55.6 & 21.6 & 41.8 & 0.0 & 0.0 & 28.3 & 2.8 & 0.0 & 0.0 & 0.0 & 68.1 & 30.5 & 41.8 & 2.4 & 37.4 & 18.7 & 40.1 & 23.2 & 45.0 & 26.2 & 4.8 \\
      dense        & WI            & 57.9 & 23.8 & 43.5 & 0.0 & 3.0 & 40.2 & 11.3 & 0.1 & 0.0 & 0.0 & 70.3 & 32.1 & 42.7 & 3.0 & 40.4 & 18.8 & 43.0 & 24.4 & 44.5 & 26.8 & 7.5 \\
    \cline{1-2}
    &&&&&&&&&&&&&&&&&&&&&&\\[-1.9ex]
      -            & WI-TTA        & 55.6 & 21.9 & 41.8 & 0.0 & 0.0 & 30.9 & 3.0 & 0.0 & 0.0 & 0.0 & 68.1 & 30.4 & 42.2 & 2.0 & 37.6 & 19.1 & 40.3 & 23.4 & 45.5 & 26.5 & 5.0 \\
    \rowcolor{\colormain}
      dense        & WI-TTA        & 57.9 & 24.1 & 43.5 & 0.0 & 3.0 & 44.4 & 12.0 & 0.1 & 0.0 & 0.0 & 70.2 & 32.0 & 43.2 & 2.3 & 40.4 & 19.2 & 43.2 & 24.5 & 44.9 & 27.0 & 7.5 \\
    \cline{1-2}
    &&&&&&&&&&&&&&&&&&&&&&\\[-1.9ex]
    \rowcolor{\colorspecial}
      oracle       & -             & 63.6 & 19.6 & 50.9 & 0.0 & 0.0 & 4.2 & 0.8 & 0.0 & 0.7 & 0.0 & 69.2 & 18.4 & 36.4 & 0.7 & 41.2 & 11.7 & 45.0 & 18.8 & 46.8 & 26.5 & 1.6 \\
    \rowcolor{\colorspecial}
      oracle       & WI            & 63.2 & 26.5 & 52.9 & 1.7 & 3.5 & 37.2 & 16.0 & 2.1 & 5.5 & 0.0 & 76.0 & 34.0 & 45.8 & 2.4 & 43.2 & 20.1 & 46.2 & 27.0 & 50.3 & 31.5 & 8.7 \\
    \rowcolor{\colorspecial}
      dense        & oracle        & 58.2 & 30.3 & 44.9 & 2.1 & 4.5 & 40.7 & 29.2 & 4.0 & 0.0 & 0.0 & 73.6 & 53.1 & 49.6 & 23.3 & 44.2 & 33.6 & 45.8 & 33.1 & 56.4 & 30.8 & 7.3 \\
    \rowcolor{\colorspecial}
      oracle       & oracle        & 63.2 & 34.8 & 54.0 & 7.3 & 5.6 & 47.9 & 37.8 & 9.2 & 5.3 & 0.0 & 79.8 & 57.9 & 53.4 & 26.2 & 47.2 & 34.6 & 48.7 & 36.4 & 61.7 & 35.6 & 12.9 \\

    \midrule
    \rowcolor{\colortitle}
    \multicolumn{23}{@{}l}{\textbf{SemCity-AE}} \\
        -          & -               & 53.9 & 16.4 & 34.3 & 0.3 & 1.5 & 6.9 & 6.9 & 2.9 & 0.7 & 0.0 & 60.4 & 11.9 & 30.8 & 0.2 & 34.4 & 8.3 & 36.1 & 16.8 & 39.5 & 17.3 & 2.8 \\
      dense        & -             & 56.9 & 17.5 & 36.6 & 1.0 & 3.4 & 8.6 & 6.4 & 2.1 & 1.0 & 0.0 & 60.9 & 14.3 & 30.9 & 0.4 & 36.2 & 8.8 & 39.7 & 18.9 & 39.8 & 19.5 & 3.0 \\
    \cline{1-2}
    &&&&&&&&&&&&&&&&&&&&&&\\[-1.9ex]
      -            & WI            & 54.4 & 25.5 & 39.3 & 12.7 & 13.4 & 42.0 & 19.0 & 8.5 & 4.2 & 0.0 & 68.2 & 29.8 & 41.6 & 3.3 & 37.8 & 15.9 & 38.8 & 26.5 & 45.0 & 26.9 & 11.8 \\
      dense        & WI            & 57.0 & 25.5 & 40.0 & 11.7 & 10.6 & 43.3 & 18.5 & 9.2 & 4.0 & 0.0 & 67.9 & 28.6 & 41.0 & 2.7 & 38.5 & 16.7 & 42.7 & 26.6 & 45.4 & 26.0 & 11.0 \\
    \cline{1-2}
    &&&&&&&&&&&&&&&&&&&&&&\\[-1.9ex]
      -            & WI-TTA        & 54.4 & 26.0 & 39.4 & 13.6 & 14.3 & 46.0 & 20.6 & 8.7 & 4.3 & 0.0 & 68.2 & 30.4 & 42.1 & 3.2 & 37.9 & 16.1 & 38.9 & 26.7 & 45.2 & 27.3 & 12.1 \\
    \rowcolor{\colormain}
      dense        & WI-TTA        & 57.0 & 25.9 & 40.1 & 12.7 & 11.2 & 46.0 & 20.0 & 9.3 & 4.0 & 0.0 & 67.9 & 29.2 & 41.4 & 2.1 & 38.6 & 16.9 & 42.8 & 26.8 & 45.5 & 26.3 & 11.2 \\
    \cline{1-2}
    &&&&&&&&&&&&&&&&&&&&&&\\[-1.9ex]
    \rowcolor{\colorspecial}
      oracle       & -             & 60.6 & 20.0 & 43.6 & 1.4 & 3.9 & 11.7 & 7.6 & 7.3 & 12.9 & 0.0 & 64.7 & 15.7 & 31.7 & 0.3 & 38.1 & 9.1 & 41.7 & 20.7 & 40.7 & 24.7 & 3.6 \\
    \rowcolor{\colorspecial}
      oracle       & WI            & 61.0 & 28.3 & 48.6 & 12.3 & 7.9 & 50.4 & 17.3 & 17.4 & 11.4 & 0.0 & 73.5 & 30.7 & 42.5 & 2.5 & 40.0 & 18.7 & 44.7 & 30.0 & 47.4 & 31.3 & 12.0 \\
    \rowcolor{\colorspecial}
      dense        & oracle        & 57.1 & 34.4 & 43.2 & 21.6 & 20.8 & 58.5 & 38.0 & 10.3 & 4.5 & 0.4 & 71.1 & 52.7 & 46.8 & 20.6 & 41.6 & 33.3 & 45.2 & 35.9 & 55.1 & 33.4 & 21.1 \\
    \rowcolor{\colorspecial}
      oracle       & oracle        & 61.2 & 37.3 & 51.2 & 23.8 & 10.7 & 52.0 & 42.2 & 20.2 & 14.9 & 0.2 & 77.0 & 56.8 & 50.2 & 21.4 & 43.5 & 35.9 & 47.5 & 39.4 & 58.8 & 39.1 & 24.8 \\
    \bottomrule
  \end{tabular}%
  }
  \label{tab:perclass_semkitti}
\end{table*}

%% file: tables/table_12_class_per_class.tex
\providecommand{\myangle}{90}
\providecommand{\PENDING}{\textcolor{red}{--}}
\providecommand{\clsq}[1]{\textcolor{#1}{$\blacksquare$}}      %

\definecolor{nuCar}{RGB}{100,150,245}
\definecolor{nuBicycle}{RGB}{100,230,245}
\definecolor{nuMotorcycle}{RGB}{30,60,150}
\definecolor{nuPerson}{RGB}{255,30,30}
\definecolor{nuRoad}{RGB}{255,0,255}
\definecolor{nuSidewalk}{RGB}{75,0,75}
\definecolor{nuOtherGround}{RGB}{175,0,75}
\definecolor{nuBuilding}{RGB}{255,200,0}
\definecolor{nuVegetation}{RGB}{0,175,0}
\definecolor{nuOtherObject}{RGB}{50,255,255}
\definecolor{nuTruck}{RGB}{80,30,180}
\definecolor{nuOtherVehicle}{RGB}{0,0,255}

\begin{table*}[p]%
  \caption{\textbf{Per-class IoU on SSCBench-nuScenes} (validation set, 12-class scheme).
    Rows, notation and colors are as in \Cref{tab:perclass_semkitti}.
    \emph{Truck} and \emph{other-vehicle} are separate classes, as in Table~II of~\cite{sscbench}.
    All metrics are reported in \%.
}
  \centering
  \scriptsize
  \setlength{\tabcolsep}{2.7pt}
  \begin{tabular}{@{}cc||cc||*{12}{c}@{}}
    \toprule
    \textbf{Visibility} & \textbf{Semantics} & \textbf{IoU} & \textbf{mIoU}
      & \rotatebox{\myangle}{\clsq{nuCar} car}
      & \rotatebox{\myangle}{\clsq{nuBicycle} bicycle}
      & \rotatebox{\myangle}{\clsq{nuMotorcycle} motorcycle}
      & \rotatebox{\myangle}{\clsq{nuTruck} truck}
      & \rotatebox{\myangle}{\clsq{nuOtherVehicle} oth.\ vehicle}
      & \rotatebox{\myangle}{\clsq{nuPerson} person}
      & \rotatebox{\myangle}{\clsq{nuRoad} road}
      & \rotatebox{\myangle}{\clsq{nuSidewalk} sidewalk}
      & \rotatebox{\myangle}{\clsq{nuOtherGround} oth.\ ground}
      & \rotatebox{\myangle}{\clsq{nuBuilding} building}
      & \rotatebox{\myangle}{\clsq{nuVegetation} vegetation}
      & \rotatebox{\myangle}{\clsq{nuOtherObject} oth.\ object} \\
    & & & & \tiny 2.0\% & \tiny  \textless{}0.1\% & \tiny  \textless{}0.1\% & \tiny  0.6\% & \tiny  0.7\% & \tiny  0.2\% & \tiny  38.0\% & \tiny  8.5\% & \tiny  0.7\% &  \tiny 21.4\% &  \tiny 27.3\% &  \tiny 0.5\% \\

    \midrule
    \rowcolor{\colortitle}
    \multicolumn{16}{@{}l}{\textbf{LMSCNet-SS}}\\
      -          & -         & 37.6    & 15.0 & 24.0 & 0.0 & 0.0 & 8.8 & 17.0 & 10.9 & 43.4 & 15.2 & 8.2 & 29.6 & 17.3 & 5.4  \\
      dense         & -        & 38.3 & 16.6 & 25.5 & 0.0 & 0.0 & 15.0 & 18.8 & 11.3 & 44.2 & 15.0 & 12.1 & 30.0 & 17.9 & 9.1 \\
    \cline{1-2}
    &&&&&&&&&&&&&&&\\[-1.9ex]
      -          & WI         & 38.6 & 26.1 & 29.2 & 5.8 & 17.6 & 33.5 & 34.0 & 25.1 & 46.3 & 26.0 & 18.0 & 33.1 & 20.0 & 24.9 \\
      dense      & WI       & 39.2 & 26.0 & 28.7 & 4.2 & 14.8 & 33.4 & 34.3 & 25.4 & 47.0 & 25.9 & 18.8 & 33.3 & 21.1 & 25.4 \\
    \cline{1-2}
    &&&&&&&&&&&&&&&\\[-1.9ex]
      -          & WI-TTA   & 38.6 & 26.2 & 29.3 & 5.5 & 17.7 & 33.5 & 34.1 & 25.2 & 46.3 & 25.9 & 18.1 & 33.2 & 20.0 & 25.0 \\
    \rowcolor{\colormain}
      dense      & WI-TTA   & 39.1 & 26.1 & 28.7 & 4.0 & 14.7 & 33.5 & 34.5 & 25.5 & 47.0 & 25.9 & 18.8 & 33.4 & 21.1 & 25.5 \\
    \cline{1-2}
    &&&&&&&&&&&&&&&\\[-1.9ex]
    \rowcolor{\colorspecial}
      oracle & - & 51.3 & 21.7 & 27.7 & 0.0 & 0.0 & 20.5 & 21.0 & 13.4 & 61.8 & 23.7 & 21.2 & 35.5 & 23.8 & 11.3 \\
    \rowcolor{\colorspecial}
      oracle & WI & 51.6 & 31.7 & 33.5 & 5.8 & 17.2 & 40.0 & 40.2 & 27.5 & 65.4 & 35.5 & 23.7 & 38.6 & 26.2 & 26.9 \\
    \rowcolor{\colorspecial}
      oracle & WI-TTA & 51.6 & 31.8 & 33.6 & 5.6 & 17.3 & 40.0 & 40.4 & 27.6 & 65.5 & 35.4 & 23.7 & 38.8 & 26.3 & 27.1 \\
    \midrule
    \rowcolor{\colortitle}
    \multicolumn{16}{@{}l}{\textbf{SemCity-AE}}\\
      -             & -              & 37.6 & 15.6 & 23.8 & 1.2 & 5.6 & 11.7 & 12.9 & 5.8 & 42.3 & 17.6 & 12.5 & 28.6 & 17.9 & 7.6 \\
      dense         & -        & 38.8 & 16.1 & 25.0 & 1.2 & 5.2 & 12.3 & 13.0 & 6.1 & 43.1 & 17.3 & 13.0 & 29.8 & 19.2 & 7.5 \\
    \cline{1-2}
    &&&&&&&&&&&&&&&\\[-1.9ex]
      -          & WI         & 38.6 & 27.3 & 32.0 & 10.3 & 26.3 & 35.7 & 34.4 & 27.9 & 46.7 & 26.7 & 17.5 & 32.8 & 20.8 & 17.1 \\
      dense      & WI       & 39.6 & 27.4 & 32.0 & 10.4 & 24.6 & 35.4 & 33.8 & 27.8 & 47.0 & 26.7 & 16.9 & 34.1 & 22.4 & 18.0 \\
    \cline{1-2}
    &&&&&&&&&&&&&&&\\[-1.9ex]
      -          & WI-TTA   & 38.6 & 27.8 & 32.1 & 12.6 & 27.2 & 36.2 & 34.9 & 28.2 & 46.7 & 26.7 & 17.6 & 32.9 & 20.8 & 17.2 \\
    \rowcolor{\colormain}
      dense      & WI-TTA   & 39.6 & 27.8 & 32.1 & 12.0 & 25.9 & 35.8 & 34.3 & 28.1 & 47.1 & 26.7 & 17.0 & 34.2 & 22.4 & 18.1 \\
    \cline{1-2}
    &&&&&&&&&&&&&&&\\[-1.9ex]
    \rowcolor{\colorspecial}
      oracle & - & 49.2 & 19.1 & 27.4 & 1.2 & 3.4 & 13.0 & 18.2 & 6.9 & 57.9 & 21.7 & 17.4 & 33.1 & 21.9 & 7.6 \\
    \rowcolor{\colorspecial}
      oracle & WI & 50.0 & 33.4 & 36.1 & 13.4 & 28.9 & 40.5 & 40.1 & 33.1 & 63.5 & 34.9 & 21.6 & 38.5 & 26.3 & 22.9 \\
    \rowcolor{\colorspecial}
      oracle & WI-TTA & 50.0 & 33.7 & 36.2 & 16.2 & 30.0 & 40.8 & 40.5 & 33.3 & 63.5 & 34.9 & 21.7 & 38.6 & 26.4 & 23.0 \\
    \bottomrule
  \end{tabular}
  \label{tab:perclass_nuscenes12}
\end{table*}

%% file: tables/visibility-agnostic-deployment.tex
\begin{table}[t]

\caption{\textbf{Visibility prior: train/test mismatch.} 
Counterpart of \Cref{tab:segmenter_agnostic_deployment} on the visibility axis: the semantic prior is held fixed at WaffleIron while the visibility prior is varied. 
For each network the \emph{matched} setting (trained and inferred with the dense prior) is compared against the network trained on the oracle prior and switched to the dense prior at inference only.
$\Delta$IoU (train-oracle minus matched) is the cost of not retraining. 
The oracle/oracle row is the matched oracle setting (oracle visibility, WI semantics), as in \Cref{tab:perclass_semkitti}.
All metrics are reported in \%.}
\label{tab:visibility_agnostic_deployment}

\centering
\setlength{\tabcolsep}{4pt}
\begin{tabular}{llccc}
\toprule
\textbf{Train vis.} & \textbf{Infer vis.} & \textbf{IoU} & \textbf{mIoU} & \textbf{$\Delta$IoU} \\
\midrule
\rowcolor{\colortitle}
\multicolumn{5}{@{}l}{\textbf{LMSCNet-SS}}\\
\rowcolor{\colormain}
dense  & dense  & 57.9 & 23.8 & --  \\ %
oracle & dense  & 36.2 & 15.3 & \textcolor{MyRed}{-21.7} \\ %
\rowcolor{\colorspecial}
oracle & oracle & 63.2 & 26.5 & --  \\ %
\midrule
\rowcolor{\colortitle}
\multicolumn{5}{@{}l}{\textbf{SemCity-AE}}\\
\rowcolor{\colormain}
dense  & dense  & 57.0 & 25.5 & --  \\ %
oracle & dense  & 45.1 & 20.3 & \textcolor{MyRed}{-11.9} \\ %
\rowcolor{\colorspecial}
oracle & oracle & 61.0 & 28.3 & --  \\ %
\bottomrule
\end{tabular}

\end{table}